**Explanation in Human-AI Systems:**
**A Literature Meta-Review**
**Synopsis of Key Ideas and Publications**
**and**
**Bibliography for Explainable AI**

Prepared by Task Area 2

Shane T. Mueller

<u>Michigan Technical University</u>

Robert R. Hoffman, William Clancey, Abigail Emrey

<u>Institute for Human and Machine Cognition</u>

Gary Klein

<u>MacroCognition, LLC</u>

DARPA XAI Program

February 2019

# Explanation in Human-AI Systems

## Executive Summary

This is an integrative review that address the question, "What makes for a good explanation?" with reference to AI systems. Pertinent literatures are vast. Thus, this review is necessarily selective. That said, most of the key concepts and issues are exressed in this Report. The Report encapsulates the history of computer science efforts to create systems that explain and instruct (intelligent tutoring systems and expert systems). The Report expresses the explainability issues and challenges in modern AI, and presents capsule views of the leading psychological theories of explanation. Certain articles stand out by virtue of their particular relevance to XAI, and their methods, results, and key points are highlighted.

It is recommended that AI/XAI researchers be encouraged to include in their research reports fuller details on their empirical or experimental methods, in the fashion of experimental psychology research reports: details on Participants, Instructions, Procedures, Tasks, Dependent Variables (operational definitions of the measures and metrics), Independent Variables (conditions), and Control Conditions.

In the papers reviewed in this Report one can find methodological guidance for the evaluation of XAI systems. But the Report highlights some noteworthy considerations: The differences between global and local explanations, the need to evaluate the performance of the human-machine work system (and not just the performance of the AI or the performance of the users), the need to recognize that experiment procedures tacitly impose on the user the burden of self-explanation.

Corrective/contrastive user tasks support self-explanation or explanation-as-exploration. Tasks that involve human-AI interactivity and co-adaptation, such as bug or oddball detection, hold promise for XAI evaluation since they too conform to the notions of "explanation-as-exploration" and explanation as a co-adaptive dialog process. Tasks that involve predicting the AI's determinations, combined with post-experimental interviews, hold promise for the study of mental models in the XAI context.

## Preface

This Report is an expansion of a previous Report on the DARPA XAI Program, which was titled "Literature Review and Integration of Key Ideas for Explainable AI," and was dated February 2018. This new version integrates nearly 200 additional references that have been discovered. This Report includes a new section titled "Review of Human Evaluation of XAI Systems." This section focuses on reports—many of them recent—on projects in which human-machine AI or XAI systems underwent some sort of empirical evaluation. This new section is particularly relevant to the empirical and experimental activities in the DARPA XAI Program.

## Acknowledgements

Contributions to this Report were made by Sara Tan and Brittany Nelson of the Michigan Technological University, and Jared Van Dam of the Institute for Human and Machine Cognition.

This material is approved for public release. Distribution is unlimited. This material is based on research sponsored by the Air Force Research Lab (AFRL) under agreement number FA8650-17-2-7711. The U.S. Government is authorized to reproduce and distribute reprints for Governmental purposes notwithstanding any copyright notation thereon.

## Disclaimer

Outline

## 1. Purpose, Scope, and Organization

The purpose of this document is to distill from existing scientific literatures the resent key ideas that pertain to the DARPA XAI Program.

## Importance of the Topic

For decision makers who rely upon analytics and data science, explainability is a real issue. If the computational system relies on a simple decision model such as logistic regression, they can understand it and convince executives who have to sign off on a system because it seems reasonable and fair. They can justify the analytical results to shareholders, regulators, etc. But for "Deep Nets" and "Machine Learning" systems, they can no longer do this. There is a need to find ways to explain the system to the decision maker so that they know that their decisions are going to be reasonable, and simply invoking a neurological metaphor might not be sufficient. The goals of explanation involve persuasion, but that comes only as a consequence of understanding the hot the AI works, the mistakes the system can make, and the safety measures surrounding it.

> *... current efforts face unprecedented difficulties: contemporary models are more complex and less interpretable than ever; [AI systems are] used for a wider array of tasks, and are more pervasive in everyday life than in the past; and [AI is] increasingly allowed to make (and take) more autonomous decisions (and actions). Justifying these decisions will only become more crucial, and there is little doubt that this field will continue to rise in prominence and produce exciting and much needed work in the future (Biran and Cotton, 2017, p. 4).*

This quotation brings into relief the importance of XAI. Governments and the general public are expressing concern about the emerging "black box society." A proposed regulation before the European Union (Goodman and Flaxman, 2016) prohibits "automatic processing" unless user's rights are safeguarded. Users have a "right to an explanation" concerning algorithm-created decisions that are based on personal information. Future laws may restrict AI, which represents a challenge to industry.

The importance of explanation, and especially explanation in AI, has been emphasized in numerous popular press outlets over the past decades, with considerable discussion of the explainability of Deep Nets and Machine Learning systems in both the technical literature and the recent popular press (Alang, 2017; Bornstein, 2016; Champlin, Bell, and Schocken, 2017; Clancey, 1986a; Cooper, 2004; Core, et al., 2006; Harford, 2014; Hawkins, 2017; Kim, 2018; Kuang, 2017; Marcus, 2017; Monroe, 2018; Pavlus, 2017; Nott, 2017; Pinker, 2017; Schwiep, 2017; Sheh and Monteath, 2018; Voosen, 2017; Wang, et al., 2019; Weinberger, 2017).

Reporting and opinion pieces in the past several years have discussed social justice, equity, and fairness issues that are implicated by "inscrutable" AI (Adler, et al., 2018; Amodei, et al., 2016; Belotti and Edwards, 2001; Bostrom andYudkowsky, 2014; Dwork, et al., 2012; Fallon and Blaha, 2018; Hajian, et al., 2015; Hayes and Shah, 2017; Joseph, et al., 2016a,b; Kroll, et al., 2016; Lum and Isaac, 2016; Otte, 2013; Sweeney, 2013; Tate, et al., 2016; Varsheny and Alemzadeh, 2017; Wachter, Mittelstadt, and Russell, 2017).

One of the clearest statements about explainability issues was provided by Ed Felton (Felton, 2017). He identified four social issues: confidentiality, complexity, unreasonableness, and injustice. For example, sometimes an algorithm is confidential, or a trade secret, or it would be a security risk to reveal it. This barrier to explanation is known to create inequity in automated decision processes including loans, hiring, insurance, prison sentencing/release, and because the algorithms are legally secret, it is difficult for outsiders to identify the biases. Alternately, sometimes algorithms are well understood but are highly complex, so that a clear understanding by a layperson is not possible. This is an area where XAI approaches might be helpful, as they may be able to deliberately create alternative algorithms that are easier to explain. A third challenge described by Felton is unreasonableness—algorithms that use rationally justifiable information to make decisions that are nevertheless not reasonable or are discriminatory or unfair. Finally, he identified injustice as a challenge: we may understand the ways an algorithm is working, but want an explanation for how they are consistent with a legal or moral code.

## Scope of This Review

A thorough analysis of the subject of explanation would have to cover literatures spanning the entire history of Western philosophy from Aristotle onward. While this Report does call out key concepts mined from the diverse literatures and disciplines, the focus is on explanation in the context of AI systems. An explanation facility for intelligent systems can play a role in situations where the system provides information and explanations in order to help the user make decisions and take actions. The focus of this Report is on contexts in which the AI makes a determination or reaches conclusions that then have to be explained to the user. AI approaches that pertain to Explainable AI include rule-based systems (e.g., Mycin), which also make determinations or reach conclusions based on predicate calculus. However, the specific focus of the XAI Program is Deep Net (DN) and Machine Learning (ML) systems.

Some of the articles cited in this Report could be sufficiently integrated by identifying their key ideas, methods, or findings. Many articles were read in detail, by at least one of the TA-2 Team's researchers. The goal was to create a compendium rather than an annotated bibliography. In other words, this Report does not exhaustively summarize each individual publication. Rather, it synthesizes across publications in order to highlight key concepts. That said, certain articles do stand out by virtue of their particular relevance to XAI. The key points of those articles are highlighted across the sections of this Report.

## Organization of This Review

We look at the pertinent literatures from three primary perspectives: key concepts, research, and history. We start in Section 2 by looking at the pertinent literatures from the perspective of the traditional disciplines (i.e., computer science, philosophy, psychology, human factors). Next, we express the key research findings and ideas on topics that are pertinent to explanation, such as causal reasoning, abduction, and concept formation (Section 3). The research that pertains specifically to XAI is encapsulated in Section 4). In Section 5 presents an historical perspective on approaches to explanation in AI—the trends and the methods as they developed over time. The key ideas and research findings (Sections 2 through 5) are distilled in Section 6, in the form of a glossary of ideas that seem most pertinent to XAI. Next, in Section 7 we point to the reports of research that specifically addresses the topic of explanation in AI applications. Section 8 is new (compared to the February 2018 release of this XAI literature review). This Section focuses on reports in which AI explanation systems were evaluated in research with human participants.

## 2. Disciplinary Perspectives

Various academic disciplines have addressed the issue of explanation. In this Section we describe these perspectives.

### Philosophy of Science

There has been considerable interest in philosophy in the role of explanation, especially in science. The discussions have involved conceiving of explanation:

(1) From a logic point of view, as a type of deductive inference (hypothetical-deductive reasoning),

(2) From a mechanistic point of view, as a form of causal reasoning about mechanisms,

(3) From a statistical point of view, as a form of statistical inference, and

(4) From a relativist/pragmatist view.

Additional discussions focused on the relation of explanation to prediction, explanation in the physical sciences (e.g., the conundrums of quantum physics), teleological questions in the analysis of biological sciences, and the matter of reductionism (i.e., could the principles of chemistry be explained in terms of theories from physics). Common arguments include discussions of modality, causality, and contrastive reasoning modes. There is a consensus that explanations relate events to theories, that is, explanations explain particular events or phenomena by reference to theories (see Van Fraassen, 1977).

Key references in philosophy of science are Feigl, et al. (1962), Hospers (1946), Kaplan (1964, Chapter IX), Putnam (1978, Lectures II and IV), and Strevens (2004). Useful summaries are presented in Wikipedia [https://en.wikipedia.org/wiki/Explanation] and in the Stanford Encyclopedia of Philosophy [https://plato.stanford.edu].

Johnson and Johnson (1992) pointed out that accounts of scientific explanation are inadequate for the analysis of explanation, broadly. That said, some of the ideas from philosophy are pertinent to XAI, and those are called out in this Report (see Section 4). Overton (2013) and Miller (2017) provide excellent reviews of theories of explanation in the Philosophy of Science.

Other relevant scholarship in this domain:

(Brodbeck, 1962; Dunbar, 1995; Eberle, Kaplan, and Montague, 1961; Feigl and Maxwell, 1962; Ginet, 2008; Giordano and Schwind, 2004; Eberle, Kaplan, and Montague, 1961; Hankinson, 2001; Harman, 1965; Hempel and Oppenheim, 1948; Hospers, 1946; Hume, 1748; James, 2003, 2003; Kaplan, 1964; Kim, Khanna and Koyejo, 2016; Kment, 2014; Langley, 1987; Lulia Nastasia and Rakow, 2010; Machamer, Darden, and Craver, 2000; Marr and Poggio, 1976; Maxwell, 1974; Maze, 1954; Meissner, 1960; Rozeboom, 1956; Rozenblit and Keil, 2002; Scriven, 1958; Simon, 1992; Smith, 2014; Strevens, 2004; Thagard, 1989; Thorpe, 1974; Van Der Linden, Leiserowitz, Rosenthal, and Maibach, 2017; Wilkenfeld, 2014, 2016)

**Psychology of Science**

Research on the psychology of science has investigated scientific reasoning and scientific explanation empirically. Although the work references explanation and explanatory coherence, its concern is broader—with the cognition of scientists and their reasoning strategies, such as hypothesis formation and exploration. Scholars such as Trout (2002) and Reiss (2012) have discussed scientific explanation in philosophy, and Salmon (2006) provided a comprehensive retrospective review of scholarship on scientific explanation, citing Hempel and Oppenheim (1948) as having seminal "epoch-making significance." Subsequent research in psychology and cognitive science on scientific reasoning includes research reports by Klahr and Dunbar, (1988) Dunbar (1996), Langley, et al. (1987), and Grotzner (2003). Within economic disciplines, the "Explanation Paradox" (Reiss, 2012) has fueled debated on the explanatory role of economic models, focusing on whether a model must be accurate in order to be explanatory.

## Psychology of Explanation

There is a large body of scholarship and research on the psychology of explanation. Much of the research focuses on the role of mental models, including the role of mental models in the understanding of computer systems (e.g., Besnard, Greathead, and Baxter, 2004; Borgman, 1986). Although much of the research is based in educational psychology (learning, and explanation to school children), useful hypotheses and taxonomies of explanation have been created. For example, Vasilyeva, Wilkenfeld, and Lombrozo, (2016) demonstrated how the learner's goals determine what counts as a good explanation.

Key theoretical, review, and empirical articles include:

(Byrne, 1992; Chin-Parker and Bradner, 2010; Craik, 1967; Ebel, 1974; Fernbach, Sloman, Louis, and Shube, 2012; Geldard, 1939; George, 1953; Giffin, Wilkenfeld and Lombrozo, 2017; Gopnik, 2000; Kass and Leake, 1987; Keil, 2006; Keil, Rozenblit, and Mills, 2004; Koehler, 1991; Kuhn, 2001; Leddo, Abelson, and Gross, 1984; Lombrozo, 2006, 2009a, 2011, 2016; Lombrozo and Carey, 2006; Lombrozo and Gwynne, 2014a; Lombrozo, Sloman, Strevens, Trout, and Skolnick Weisberg, 2008; Malle, 1999,2004, 2011; Manicas and Secord, 1983; McClure, 2002; McClure and Hilton, 1997, 1998; McClure, Sutton, and Hilton, 2003; Malle, 1999, 2004, 2011; Rey, 2012; Royce, 1963; Sloman, 1994; Swartout, 1987; Vasilyeva, Wilkenfeld, and Lombrozo, 2016; Vasilyeva and Lombrozo, 2015)

## Social Psychology

One focus of explanation has been on social interaction, which often links to how an individual creates an explanation of another person's beliefs or behaviors. This includes contexts such as learning and teaching (Lombrozo, 2016); lay explanations of social behavior (e.g., Antaki, 1989; Axelrod, 2015), argumentation (Antaki and Leudar, 1992), and social attribution (Hilton, 2017; Slugoski, Lalljee, Lamb, and Ginsburg, 1993).

**Psycholinguistics**

Language has always been an necessary part of explanatory reasoning. To a large extent, first generation explainable AI systems of the 1970 and 1980s had the goal of verbalizing internal goal and strategy states using production rule systems and other logics (see Clancey, 1986). Lingual descriptions of complex systems remain a topic of interest (see Hendricks et al., 2016). In fact, some have argued that explanation is necessarily lingual and conversational (Hinton, 1990; Walton, 2004, 2007, 2011). Among the most important concepts—especially in relation to automated language production—is the notion of "common ground" (Clark et al., 1991), because the language used for explanation needs to refer to the understanding that is shared by the explainer and explainee. Also important are Grice's (1975) conversational maxims (quality, quantity, relation, and manner), which dictate norms and expectations for communication. For example, the maxims suggest that choosing what should not be explained is critical; if something that is understood is explained, it might incorrectly signal to the explainee that they might not understand the concept. There are also connections between linguistic theory and social contexts (Antaki 1989; Antaki and Leudar, 1992; Slugoski, Lalljee, Lamb, and Ginsburg, 1993). Graesser and colleagues have contributed substantially to work on intelligent tutoring systems, with language capabilities that are closely linked to findings from basic psycholinguistic research (Graesser and Franklin, 1990; Graesser, Gordon, and Brainerd, 1992; Graesser, Lang, and Roberts, 1991). Walton (2004, 2007, 2011) has discussed dialectical (dialog-based) theories of explanation.

Additional pertinent references are:

(Blokpoel et al., 2012; Cardona-Rivera, Price, Winer, and Young, 2016; Core, et al., 2006; Gkatzia, Lemon, and Rieser, 2016; Grice, 1975; Rips, Brem, and Bailenson, 1999; Searle, 1985; Slugoski )

**Team Science**

The literature on coordination (i.e., how individuals work together to solve common problems) is relevant in several ways. Importantly, Ford (1990) and Brézillon (1994) were among the first to argue that explanation in intelligent systems involves cooperative problem solving—a coordination problem. The literature on coordination is broad, and much of it is not relevant to the XAI problem, but schemes put forward by Klein, Feltovich, Bradshaw, and Woods (2005) and Johnson et al. (2014) are especially relevant to human coordination with AI systems.

Klein et al. and Johnson et al. build on Clark's (1996) linguistic concept of common ground as a framework for understanding team coordination. The central process is ensuring that the team members' actions are interpredictable. Interpredictability assumes that all the team members sign on to a "basic compact" to facilitate coordination and to prevent its breakdown. That usually requires the team members to align multiple goals, and to establish and sustain common ground—the knowledge, beliefs and assumptions shared among the team members. These may include: the roles and functions of each participant, the skills and competencies of each participant, the routines that the team is capable of executing, the goals of the participants, their degree of commitment to the team goals, and the stance of each participant (e.g., perception of time pressure, level of fatigue, competing priorities).

To support common ground, teams engage in the following activities: structuring the preparation to calibrate knowledge and establish routines, sustaining common ground using clarifications and reminders, balancing the task work and the team work, updating each other about changes that may not have been noticed by everyone, monitoring one another to detect signs of common ground breakdown, detecting anomalies that might signal a potential loss of common ground, and repairing common ground once it has been lost. Typically, one or more team members will relax their own shorter-term goals in order to permit more global and long-term team goals to be addressed.

Klein, et al. and Johnson, et al. laid out some "rules for team players," which extends the common ground concepts by applying them to all of the agents (both human and machine) in sociotechnical work systems:

1.  An intelligent agent needs to explicitly sign on to a Basic Compact and work to build and sustain common ground.

2.  An intelligent agent needs to model the other agents' intentions and actions; if another participant is struggling, the intelligent agent will need to adapt.

3.  An intelligent agent needs to be predictable. Predictability is essential to coordination.

4.  The intelligent agent needs to be directable.

5.  An intelligent agent needs to signal its status and intentions to its teammates.

6.  An intelligent agent needs to observe and interpret the signals of the status and intentions of its teammates.

7.  An intelligent agent needs to be able to engage in goal negotiations.

8.  An intelligent agent is not autonomous. It must engage in collaborative sensemaking, problem solving, replanning, and task execution.

9.   An intelligent agent must participate in the team's activities that manage attention.

10. An intelligent agent must participate in efforts to control coordination costs.

For XAI we would highlight the assertion that machines—if they are to act as team players— must make themselves observable and understandable.

Additional pertinent papers include: (Brézillon and Pomerol, 1997, 1997; Galinsky, Maddux, Gilin, and White, 2008; Karsenty and Brezillon, 1995; Khemlani and Johnson-Laird, 2010; Linegang, Stoner, et al., 2006; Llinas et al., 1998).

**Other Human Factors, Cognitive Systems Engineering and Decision Aids**

Some human factors research has examined the role of explanation in decision support systems. Research includes empirical and theoretical studies of how humans use technology, especially in relation to explanations. A recent discussion by Abdul, et al. (2018) proposed a research agenda on XAI from the human factors perspective; a number of researchers who have discussed human-centered frameworks for explanation. A number of general discussions on developing systems from a human factors perspective are relevant (e.g., Conant and Ashby, 1970; Bellotti, et al., 2002; Landauer, 1988; Lazar, Feng, and Hochheiser, 2017; Rasmussen, 1983).

Within human factors community, human interaction with AI systems (and automation broadly) is often examined from only one of a few perspectives. These include system transparency (Lyons, 2013; Mercado, Rupp, Chen, Barnes, Barber, and Procci, 2016) and monolithic trust states (Hoffman, 2013; Hoffman, et al., 2013; Balfe, Sharples, and Wilson, 2018; Fallon and Blaha, 2018).

Other relevant research on human-automation teaming, communication, and explanation cover a range of topics (Ballas, 2007; Bisantz, Finger, Seong, and Llinas, 1999; Blokpoel et al., 2012; Cahour and Forzy, 2009; Clarke and Smyth, 1993; Davies, 2011; Dunbar, 1995; Frank et al., 2016; Groce et al., 2014; Hoffman, 2013, 2017b; Johnson and Johnson, 1993; Kilgore and Voshell, 2014, 2014; Klein and Hoffman, 2008; Klein, Moon, and Hoffman, 2006a; 2006b; Lyons, 2013; McBride, Rogers, and Fisk, 2014; Miller, 1988; Muir, 1994; Muramatsu and Pratt, 2001; Pirolli and Card, 1999, 2005, Prietula et al., 2000, 2000; Roth, Butterworth, and Loftus, 1985; Sebok and Wickens, 2017; Strauch, 2017; Tullio et al., 2007; Vermeulen, Luyten, van den Hoven, and Coninx, 2013; Wallace and Freuder, 2001; Wille, 1997; Wright, Chen, Barnes, and Hancock, 2016; Yates, Veinott, and Patalano, 2003).

**Conclusions Concerning the Disciplinary Perspectives**

Goodman and Flaxman (2016) ranked information processing methods in terms of their human interpretability, ranging from simple linear models, to non-parametric methods, to ensemble methods, to neural nets and deep learning. The latter type of method is described as "the biggest challenge" (p. 6). The challenge of explainable AI entrains concepts of representation, modeling, language understanding, and learning. Broad success of XAI is tantamount to the full realization of the aspirations of AI.

> *While explainable AI is only now gaining widespread visibility, the ML literature and that of allied fields contain a long, continuous history of work on explanation and can provide a pool of ideas for researchers currently tackling the task of explanation (Biran and Cotton, 2017, p. 4).*

Looking across the disciplinary literatures, a number of taxonomies have been put forward, to classify types of explanations, qualities of good explanations, functions of explanation, and goals of explanation. Listing all of these in a discipline-by-discipline structure would be unhelpful, primarily because there is essential overlap. It is possible to avoid needless repetition by synthesizing the various schemes and taxonomies. The purpose of this Report is to synthesize that material, and then carry that synthesis forward, to formulate challenges and possibilities for XAI.

## 3. Findings From Research on Pertinent Topics

**Previous Reviews of the Literature**

The state of explanation in artificial intelligence systems has been reviewed extensively. It is generally agreed that "Enhancing the explanatory power of intelligent systems can result in systems that are easier to use, and result in improvements in decision-making and problem-solving performance" (Nakatsu, 2004, p. 575).

(Alonso, Castiello, and Mencar, 2018; Biran and Cotton, 2017; Brézillon, 1994; Brézillon and Pomerol, 1997; Buchanan and Shortliffe, 1984; Chakraborty et al., 2017; Clancey, 1986; Core et al., 2006; Craven and Shavlik, 1996; Doran, Shulz, and Besold, 2017; Doshi-Velez and Kim, 2017ab; Doyle, Tsymbal, and Cunningham, 2003; Ellis, 1989; Fox, Long, and Magazzeni, 2017; Gilpin, et al., 2018; Hoffman and Klein, 2017; Hoffman, Mueller, and Klein, 2017, 2018; Kass and Finin, 1988; Lacave and Díez, 2002; Lipton, 2016; Miller, 2017; Miller, Howe, and Sonenberg, 2017; Molnar, 2018; Moore and Swartout, 1988; Moulin, Irandoust, Bélanger, and Desbordes, 2002; Nakatsu, 2004; Sørmo, Cassens, and Aamodt, 2005; Swartout, 1987; Swartout and Moore, 1993; Weston et al., 2015; Wick and Thompson, 1992).

A number of researchers have proposed prescriptions on how to make explanations, or taxonomies of explanation, or descriptions of properties of explanation, either as the main thesis of a paper or as preliminary or synthesized discussion (Byrne, 1991; Felton, 2017; Kass and Leake, 1987; Kulesza et al., 2015; Rosenthal et al., 2016; Sørmo, Cassens, and Aamodt, 2005; Swartout and Moore, 1983; Van Der Linden, 2002; Wallis and Shortliffe, 1981; Wick and Thompson, 1992; Zhang et al., 2016).

**Fairness, Transparency, Safety, Accountability, and Ethics**

One aspect of explainable systems is related to ethical issues (Bostrom and Yudkowsky, 2014).

These are generally related to how modern AI systems impact society and technical-social systems. Felton (2017) identified four social and ethical aspects of transparent algorithms, and research has focused on many of the social and ethical aspects of algorithms. These include:

- Transparency (Bansal, Farhadi, and Parikh, 2014; Fallon and Blaha, 2018; Hayes and Shah, 2017; Wachter, Mittelstadt, and Russel, 2017),

- Accountability (Adler, et al., 2018; Belotti and Edwards, 2001; Kroll et al, 2016),

- Safety and privacy (Amodei, Olah, Steinhardt, Christiano, Schulman, and Mané, 2016, Hajian, et al., 2015; Lum and Isaac, 2016; Otte, 2013; Verheny and Alemzadeh, 2017),

- Fairness and bias (Dwork, Hardt, Pitassi, Reingold, and Zemel, 2012; Hajian, et al., 2015, Joseph et al., 2016ab; Sweeney, 2013), and licensure (Tate, et al., 2016).

**Trust**

Numerous studies have linked trust to explanation in the context of AI systems use. Hoffman, Mueller and Litman (2018) provide a review on this topic. Various forms on intelligent systems are trusted more if their recommendations are explained. (See also Antifakos et al., 2005; Dourish, 1995; Suermondt and Cooper, 1992; Tullio et al., 2007).

**Causal Reasoning and Abductive Inference**

In the fields of AI and philosophy of science, problem diagnosis, causal explanation, scientific discovery, and explanation are regarded as highly related or even overlapping (e.g., Darden, Machamer, and Crave, 2000; Overton, 2013; Shank, 1986). The literatures in psychology, education and philosophy make it clear that explanation, as a cognitive process, is closely related to, and perhaps even the same as causal reasoning, since explanations often refer to causation or causal mechanisms, and causal analyses are believed to have explanatory value (Cohen, 2008; Halpern and Pearl, 2005a,b; Krull and Anderson, 1997; Pearl, 1988; Pearl and Mackenzie, 2018; Trout, 2002). Explanations relate the event being explained to principles and invoke causal relations and mechanisms (Lombrozo, 2012). Researchers generally regard explanation as the process of generating a mechanistic or causal analysis of why something happened or how something works.

Relevant research on causal reasoning relevant to explanation includes:

Ahn (1998); Ahn et al. (1995); Ahn and Kalish (2000); Chater and Oaksford, (2006); Cheng (1997, 2000); Einhorn and Hogarth (1986); Eiter and Lukasiewicz (2002, 2006); Feltovich, Hoffman, Woods, and Roesler (2004); Graesser, Baggett, and Williams (1996); Halpern and Pearl (2005a,b); Heckerman and Shachter (1994); Hesslow (1988). Hilton (1990, 2005); Hilton, McClure, and Sutton (2010); Klein, Rasmussen, Lin, Hoffman, and Case(2014); Krull and Anderson (1997); Lombrozo (2007, 2010, 2010); McGill (1991); Rasmussen, Nixon, and Warner (1990); Samurçay and Hoc (1996); Selfridge, Daniell, and Simmons, D. (1985); Woodward, (2005); Zwaan, Langston, and Graesser (1995); Zwaan, Magliano, and Graesser (1995).

The scholarship in psychology, education and philosophy makes it clear that explanation, as a cognitive process, is closely related to abductive inference that is, "inference to the best explanation" (Harman, 1965; Overton, 2012, 2013). The literatures in psychology, education and philosophy make it clear that explanation, as a cognitive process, is also closely related to retrospection since explanations are often focused on understanding past events (Douven, 2011; Lombrozo, 2012; Shank, 1998; Wilkenfeld and Lombrozo, 2015; Lombrozo and Gwynne, 2014a).

**Causal and Mechanistic Reasoning about Events and Concepts**

Some research has specifically investigated reasoning about events, entities, and categories. Although correlation is a good signal for causal reasoning, when asked to provide explanations for events people prefer mechanistic and causal accounts rather than ones based purely on correlations (Ahn, 1998; Anh and Kalish, 2000; Ahn, Kalish, Medin, and Gelman, 1995; Bechtel and Abrahamsen, 2005; Croskerry, 2009; De Kleer and Brown, 2014; Glennan, 2002; Hegarty, 2004; Hoffman, Klein, and Miller, 2011; Klein, et al., 2014; Szpunar, Sprenng, and Schachter, 2014). Similarly, Murphy and Medin (1985) showed that a conceptual coherence is often governed by an underlying theory—a causal mechanism of understanding; a point echoed by Lombrozo (2009b), in which she argued that explanations and concept formation are closely related.

**Analogy**

Explanations often take the form of analogical comparisons. The psychology of analogical reasoning has long been studied in cognitive science, and the literature of reports and books is extensive. A traditional focus for research has been on the problem of how analogies are spontaneously noticed. With regard to theory, analogy is generally regarded in terms of the mapping of parts, concepts, and relations (cf. Falkenhainer, 1980; Falkenhainer, Forbus and Gentner, 1993; Gentner and Stevens, 1983). The nature and structure of analogy has long been considered in the philosophy of science, including the relations of analogy to scientific explanation and reasoning, and the relation of analogy to models and scientific metaphors. Indeed, the very concept of a "neural net" can be understood as an analogy or metaphor.

In computer science and philosophy of science as well as psychology, there has been considerable work on analogical reasoning. Computational systems for coherent structure mapping have been developed and evaluated. Schemes for assessing the quality of analogs have been proposed and empirically evaluated. Ways in which people engage (or fail to engage) in analogy-based transfer during problem solving have been studied in the psychology laboratory. Key references are Clausner (1994), Eliot (1986), Gentner, Holyoak and Kokinov (2001), Gentner and Schumacher (1986), Hoffman (1989, 1998), Hoffman, Eskridge and Shelly (2009), Spiro, et al. (19898), and Thagard (1989). While analogs are often used in explanations, and may have some use in the creation of global explanations for how AI systems operate, we focus here on the qualities and functions of explanations, broadly.

Additional relevant research includes:

(Clement, 1988; Eliot, 1986; Falkenhainer, Forbus, and Gentner, 1989; Forbus, Gentner, and Law, 1995; Gentner, Holyoak, and Kokinov, 2001; Hoffman, 1995; Spiro, Feltovich, Coulson, and Anderson, 1988)

## Understanding Explanations

Although research on explanation is typically focused on the person (or system) producing the explanation, a critical aspect of this is whether the offered explanation has an impact on the individual who is the recipient or beneficiary of the explanation: Does the explainee understand the system, concepts, or knowledge? Research and scholarship in this area ranges from work on understanding in general (e.g., Rosenberg, 1981; Keil, 2006; Smith, 2014), to experimental philosophy examining the distinction between knowing and understanding and what it means to understand (Wilkenfeld, 2016; Wilkenfeld et al. 2016ab), to applied research on education and learning examining factors that influence understanding (e.g., Chi et al., 1994); Feltovich et al., 2001)

Additional research related to how explanations are understood includes:
(Ahn, Kalish, Medin, and Gelman (1995); Chi, Leeuw, Chiu, and LaVancher(1994); Feltovich, Coulson, and Spiro (2001); Keil (2006); Smith (2014); Rosenberg (1981); Wilkenfeld (2016); Wilkenfeld, Plunkett, and Lombrozo, (2016a,b).

## Failures and Limitations in Understanding

An important area of research concerns the ways in which people fail to learn in general, and either fail to understand or are actually misled by explanations. This research has similarities to the heuristics-and-biases literature in the field of judgment and decision making, insofar as it focuses on the limitations and difficulties of understanding (see Fernbach et al., 2013; Tworek and Cimpian, 2016). One branch of this research is centered on teaching and learning (see Bjork and Bjork, 2014; Bjork, Dunlosky, and Kornell, 2013), with special focus on how learners are often miscalibrated about their own judgments of learning and their expectations for what makes learning effective (i.e., metacognition). Feltovich et al. (2001) and Mueller and Tan (2017) focused on how beliefs (including incorrect beliefs) are mutually-supportive thus resist change, sometimes via so-called "knowledge shields." These are arguments that learners make that enable them to preserve their reductive understandings. Other research has focused more directly on explanation, especially depth and detail that can be distracting (Goetz, and Sadoski, 1995), or the "illusion of explanatory depth" in which people feel they sufficiently understand something, and overestimate how well they understand (e.g., Rozenblit and Keil, 2002; Van Der Linden et al., 2017).

**Comprehension of Complex Systems**

Especially in the field of instructional design, the explanation and understanding of complex systems (e.g., electronic systems; biological systems) has been a focus of research. The ways in which people tend to form simplistic or reductive mental models has been revealed, and this includes a rostering of the knowledge shields. A focus for instructional design has ben to develop methods to get people to recognize when their understandings are reductive, and recognize when they are employing a knowledge shield that prevents them from developing richer mental models.

(Hilton, 1986, 1996; Prietula, Feltovich, and Marchak, 2000; Schaffernicht and Groesser, 2011; Tullio, Dey, Chalecki, and Fogarty, 2007)

**Counterfactual and Contrastive Reasoning**

Counterfactual reasoning is a central notion in some theories of explanation. Explanations should provide information about when, why and how effects or outcomes might change or be different (e.g., Woodward, 2003). Both causal understanding and the understanding of concepts often involve asking hypothetical counterfactual questions. Questions of the counterfactual form

"What if?" and "What would happen if?" and questions in the contrastive form "You says it is X but why is it not Y?" represent the attempt of learners to explore the range of variation of categories and concepts, and the causal structure of events. It has long been argued in psychology that the development of schemas, concepts, prototypes, or categories hinges on experience with instances that are typical but also instances that are atypical. Individuals need to experience the range of variation on instances in order to develop robust categories or concepts. It has long been understood in the field of Expertise Studies that genuine expertise is achieved only after long practice and experience with rare, challenging or especially difficult cases or situations (Hoffman, et al., 2014).

Additional relevant research includes: Byrne (1997; 2017); Byrne and McEleney (2000); Kahneman and Varey (1990); Halpern and Pearl (2005a,b), Hilton, McClure and Slugoski (2005); Hitchcock (2003); Hitchcock and Woodward (2003); Hoffman, et al., (2014); Khemlani and Johnson-Laird (2010); Kim, Khanna, and Koyejo (2016); Lipton, (1990); Mandel (2003a,b), McCloy and Byrne (2000); and Thompson and Byrne (2002). Mandel et al (2011) and Mandel (2015) provides reviews of scholarship on counterfactual reasoning.

**Individual Differences and Motivation**

A small body of research has taken an individual differences approach to understanding the effectiveness of explanations. Researchers have looked for systematic factors (such as a personality trait) that differ across people and explains their ability to understand an explanation. For example, Fernbach et al. (2013) showed that the score on the Cognitive Reflection Test (CRT)—a brief assessment of mathematical problems with a hidden twist—predicts susceptibility to biases in explanatory depth.

Individual differences research relevant to explanation includes: Fernbach, Sloman, St. Louis, and Shube (2013); Gopnik (2000); Keil (2006); Klein et al. (2014); Lipton (2000); Litman and Sylvia, (2006); Lombrozo and Rutstein,(2004); Maheswaran and Chaiken (1991); Rozenblit and Keil (2002); and Singh, Molloy, and Parasuraman (1993b).

## Learning and Concept Formation

Explanation and understanding are intrinsically linked to learning—certain forms of learning can be thought of as the act of attaining an understanding; teaching frequently involves explaining. The process of explanation sometimes helps learners develop new concepts or conceptual categories, or revise their understanding of concepts they already know (Lombrozo, 2009b; Murphy and Medin, 1985). A body of research has explored these connections in some detail, examining how explanations (especially causal and mechanistic explanations) can help support faster learning and better understanding of complex concepts. A number of other researchers have discussed aspects of learning relevant to explanation.

(Andriessen, 2002; Bjork et al., 2013; Doumas and Hummel, 2005; Falkenhainer, 1990; Facione and Gittens, 2015; Lombrozo, 2016; Rehder, 2003; Selfridge, Daniell, and Simmons, 1985; Williams, Lombrozo, and Rehder, 2012; Williams and Lombrozo, 2013; Feltovich, Coulson, and Spiro, 2001).

## Mental Models

The understanding that people achieve of computer systems (and other kinds of complex systems) is widely described as involving representations called "mental models" (sometimes called "cognitive maps"). These are believed to be partial abstractions that rely on domain concepts and principles. Mental models are often described as being inductive or abductive inferences. Because the notion of a mental model is so central to many conceptions of explanation, there is considerable relevant work both on explanation and on mental models in general.

(e.g., Axelrod, 2015; Besnard, Greathead, and Baxter, 2004; Byrne, 2002; Chunpir and Ludwig, 2017; De Kleer and Brown, 2014; Erlich, 1996; Friedman, forbus, and Sherin, 2018; Friedman and Nissenbaum, 1996; Gary and Wood, 2011; Gentner and Stevens, 1983; Gigerenzer, 1993; Gigerenzer, Hoffrage, and Kleinbölting, 1991; Halasz, and Moran, 1983; Hegarty, 2004; Hegarty, Just, and Morrison, 1987; Heller, 1994; Hilton, 1986, 1996; Hilton and Erb, 1996; Jamieson, 1996; Johnson-Laird, 1980, 1983; Kieras and Bovair, 1984; Klein and Hoffman, 2008a, 2008b; Klein, Phillips, Rall, and Peluso, 2007; Kulesza et al., 2012; Lewis, 1986; Mandel, 2011; Moray, 1987b, 1990, 1996, 1998; Moultin and Kosslyn, 2009; Phillips, Ososky, Grove, and Jentsch, 2011; Rouse and Morris, 1986; St-Cyr. and Burns, 2002; Wilkison, Fisk, and Rogers, 2007; Williams, Hollan, and Stevens, 1983; Wilson and Rutherford, 1989; Young, 2014; Zachary and Eggleston, 2002; Zhang and Wickens, 1987; Zovod, Rickert, and Brown, 2002 )

Especially pertinent to XAI are reports on attempts to elicit, represent and measure or evaluate mental models (Besnard, Greathead and Baxter 2004; Borgman, 1986; deGreef and Neerincx, 1995; Dodge et al., 2018; Doyle, Radzicki, and Trees, 2008; Kim, et al., 2016; Klein and Hoffman, 2008; Muramatsu and Pratt, 2001; Schaffernicht and Groesser, 2011; Tullio et al., 2007). Unfortunately, most studies provide little or no detail concerning the exact elicitation and representation methods used (an exception being the Doyle et al. study), but generally speaking, researchers rely on one of two methods: either interviews (semi-structured or structured survey) or some sort of diagram creation task (e.g., influence diagrams, flow diagrams, etc.). Muramatsu and Pratt (2001) utilized a method that might be particularly applicable to XAI. Specifically, they presented test cases to users (web search queries) and asked users to predict the results and then explain why they thought the predicted results would obtain.

A study that is pertinent to XAI and that clearly illustrates a methodology for eliciting mental models, is Dodge et al. (2018). The researchers sought a general methodology for formulating human-understandable explanations of software agents. Their research was contextualized in the development of explanations of the AI competition bots the Starcraft competitions. The assumption was that the learners/game players would have to have confidence and appropriate trust in the explanations, which would derive from knowing that the explanations came from an expert.

The researchers sought explanations that came from a domain expert who was not also an expert in AI. The reason for this choice was that such a person would be more likely to provide explanations that would make sense to a game learner/player who is a layperson (that is, also not an expert in AI). The researchers studied the reasoning of shoutcasters, networked individuals who share their knowledge and reasoning. At Starcraft tournaments, shoutcasters provide running commentary on games, which is synched with a video of the game play. Their commentary is generally focused on the question of What is happening? and Why is it happening? and their goal is to answer these questions in anticipation of what the Their commentary is in the form of a dialog between pairs of shoutcasters. The researchers examined transcripts of 10 videos, involving the commentary of 16 shoutcasters. The researchers asserted that the participant shoutcasters were "expert explainers," though this was not empirically validated, nor were data provided on the participant's experience level. However, the shoutcasters provided data that was consistent in terms of the kinds of explanations that were provided. Specifically, they explained what was happening in game play, what might happen next, why certain things happened, and why certain other things did not happen.

Shoutcasters have full observability of the game, whereas the bot agents have limited visibility into the game (e.g., some objects might not be visible to them). Given their "God's eye view," the shoutcasters can take the perspective of the bots to understand why they are doing what they are doing, and what they might do next. For example, if it seems that a fight was brewing, the shoutcasters would seek pertinent information, for example. Shoutcasters' commentary is generally focused on the questions of What is happening? and Why is it happening? and their goal is to answer these questions for a person who is viewing the game, in anticipation of what the viewer might wonder about.

The researchers asked the following questions of the transcribed shoutcaster dialogs: What information do the shoutcasters seek? How do they seek it? What questions do they try and answer and how do they form their answers, what game concepts do they rely upon when building their explanations? This research can be taken as a model for studying human-human explaining with respect to AI agents.

Related to the use of diagramming tasks for the elicitation, representation and analysis of mental models, it is noteworthy that extensive research on Concept Mapping has proven the utility of the Concept Map generation task in allowing people to express their knowledge, and has proven the value of propositional analysis of Concept Maps in the evaluation of mental models. This includes the evaluation of mental models of complex systems. See for instance Cañas, et al., 2003; Hoffman et al., 2017; Moon, et al., 2011; Novak, 1998. Also see the discussion of Concept Mapping in Section 5 of this Report.

**Prospective Reasoning and Planning**

Explanation often serves the function of enabling people to mentally project to possible futures states, outcomes, or events. Prospective (causal) reasoning is different from counterfactual reasoning: the latter focuses on how things might or might not have been different if aspects of the past had been different. The former focuses on things that might or might not happen in the future. In addition, prospection includes abstractions in the form of durative events (a cause can continue on into the future even after an effect has manifested). Because of the focus on the future, planning may be considered a form of prospective reasoning, and planning has been a focus of research on explanation.

(Bidot, Biundo, Heinroth, Minker, Nothdurft, and Schattenerg, 2010; Chakraborti, Sreedharan, Zhang, and Kampbhampati, 2017; Fox, Long, and Magazzeni, 2017; Ghallab, Nau, and Traverso, 2004; Kim, Cacha, and Shah, 2013; Reidl and Young, 2010; Seegebarth, Müller, Schattenberg, and Biundo, 2012ab; Smith, 2012; and Sohrabi, Baier, and McIlraith, 2011).

With very few exceptions, most analyses of causal explanation are retrospective. That is, they deal with things that might or might not have happened in the past (this the "standard" form of the counterfactual). The goal of the reasoning is to explain what has already happened. (Axelrod, Mitchell, Russo, and Rennington, 1989: Klein and Crandall, 1995; Moore and Hoffman, 2011).

**Explanation as Dialog**

The notion that explanation is an "interactive construction" or dialog process—has been the subject of extensive discussion; there is a broad consensus across the pertinent disciplines spanning first generation systems (e.g., Kass and Finn, 1988; Moore and Swartout, 1988) and recent scholarship (Arioua, Buche and Croitoru, 2017; Arioua and Croitoru, 2015; Bansal, 2018; Bansal, Farhadi, and Parikh, 2014; Brezillon and Pomerol, 1997; Core, et al., 2006; Graesser and Franklin, 1990; Goguen et al., 1983; Graesser, Gordon, and Brainerd, 1992; Graesser, Lang, and Roberts, 1991; Sacks and Schegloff, 1974; Suthers, Woolf and Cornell, 1992; Walton, 2004, 1007, 2011; Weiner, 1989).

**Self-Explanation**

Psychological research has demonstrated that deliberate self-explanation improves learning and understanding. This finding holds for both self-motivated self-explanation and also self-explanation that is prompted or encouraged by an instructor. Self-explanation serves a particularly useful role in helping learners overcome the illusion of explanatory depth. Such reductive understanding can be corrected by asking for explanations.

(Berry and Broadbent, 1987' Bjork et al., 2013; Calin-Jageman and Ratner, 2005; Chi, et al., 1989, 1994; Chi and VanLehn, 1991; Fernbach, et al., 2012; Ford, Cañas, and Coffey, 1993; Jones and VanLehn, 1992; Mills and Keil, 2004; O'Reilly, Symons, and MacLatchy-Gaudet, 1998; Rittle-Johnson, 2006; Rosenblit and Keil, 2002; VanLehn, Ball, and Kowalski, 1990).

**Transfer and Generalization**

The attempt by experimentalists to demonstrate task-to-task transfer (both near and far transfer) in the academic laboratory environment has met with limited success (Baldwin and Ford, 1988; Sala and Gobet, 2017). What it is that transfers (e.g., concepts, principles, strategies) is not always entirely clear. It is often assumed that reasoning is limited when people focus on superficial features of problems, and that, in the ideal case, what transfers is higher level principles of problem "deep structure." Alternatively, it might be that what appears to a research to be a superficial concept might, in the minds of the research participants, be more complex. Additionally, "transfer" might be the wrong metaphor. Rather, the effect of learning and experience across multiple types of problems may be to expand the learner's repertoire of reasoning strategies (Lobato and Siebert, 2002). Nevertheless, it is generally acknowledged that understanding and training need to promote the transfer of concepts and principles, and good explanations should facilitate that (Lombrozo and Gwynne, 2014b; Rittle-Johnson, 2006; Williams and Lombrozo, 2010; Williams, Lombrozo, and Rehder, 2013).

## 4. Key Papers and Their Contributions that are Specifically Pertinent to XAI

Looking across the disciplines surveyed in the previous Section, this Section summarizes those papers that are especially pertinent to XAI because they introduced the key concepts and they entrain the important challenges with a particular focus on the ways in which people understand complex computational systems. Section 8 of this Report provides details on a second set of key papers, those reporting some form of empirical evaluation of explanations.

**Table 4.1. Key Papers and Their Contributions.**

| Berry and Broadbent (1987) |
| --- |
| Explored the effect of explanations on user's performance at a simulated decision task, deciding about which tests to perform for potential river pollutants, and thereby determine which factory was responsible for the pollution. The software system (based on probability calculations) provided its own recommendations and these were accompanied by human-generated explanations. In one condition, a global explanation was provided to explain each computer-generated recommendation and in another condition the participants could ask about the "why" of the recommendations. and receive some further (local) explanation. Participants who were required to self explain (out loud) performed significantly better than participants who were required to verbalize but who had not received any form of explanation. Participants who received the additional "why" (local) explanations or the explanation/verbalization combination maintained a superior performance level on subsequent unaided trials. |

| Biran and Cotton (2017) |
| --- |
| Provide an up-to-date review of explanation in AI, machine learning, neural network systems, and related fields.<br>Distinguishes between machine-interpretable models and human-interpretable models.<br>Distinguish between explanation and justification.<br>Review approaches to active exploration and challenges to explaining various kinds of AI systems. |

**Blokpoel, et al. (2012)**

Experimental demonstration of how human-human cooperative communication depends on a mutual perspective-taking capacity.

**Boy (1991), Greenbaum and Kyng (1991)**

Concept of "cooperative design," in which users play a role in the creation of the AI systems.

**Brézillon (1994), Brezillon and Pomerol (1997), Karsenty and Brézillon (1995)**

Discusses context and goals in explanation.

Explaining is conceived as cooperative problem solving.

**Chakraborty, et al. (2017)**

Argue that interpretability is multidimensional, and identify some of the dimensions, including: interpretability, explainability, accountability, fairness, transparency, functionality. They also consider the need to rationalize, justify, and understand systems, and users' confidence in their results.

**Clancey (1981)**

Argued that the user needs a mental model of the rule base of expert systems. Proposed that the key weakness of explainability in expert systems could only be solved if the system has a human-interpretable model of the procedural rule base, distinguishing the domain model (e.g., taxonomic and causal relations), reasoning strategy, and justification for inferences. Mycin's rules were written in a formal, object-attribute-value LISP notation that is not human-interpretable. However, Mycin's explanation system could generate English forms that are readily understood, that is, a human-interpretable form or representation of the rule base. GUIDON and Mycin's explanation systems interpreted the rule bases directly. For examples, see Figure 8.5 in Clancey (1987) and the Appendix in Daliwhal and Benbasat, 1996.

**Clarke and Smyth (1993)**

Demonstration of human-computer cooperation in problem solving. The AI system relied on a goal hierarchy, a shared set of definitions of domain concepts, and a mechanism to declare and share goals and sub-goals, and a mechanism to generate and negotiate alternative solutions to problems. Although not focused on explanation per se, the work illustrates how the computational instantiation of the a model of the user can lead to improved performance of the human-machine system by improving the user's understanding of problems.

**Darlington (2013)**

Good and succinct review of explanation systems, types of explanations and how various intelligent systems (including expert systems) were architected so as to provide explanations. Argues that symbolic systems are better suited to the task of providing explanations than non-symbolic systems.

**Dhaliwal and Benbasat (1996)**

Proposed an framework for evaluating knowledge-based systems that provided explanations. Based on cognitive theories, the argued that explanations need to be evaluated in terms of the extent to which they facilitate learning and decision making by relying on a model of explanation strategies. They argued that explanation capabilities need to take into account the user's level of expertise. They asserted that evaluation must address three high-level questions: (1) To what extent are the explanations generated by expert systems relied upon in decision making, (2) What factors influence the use of the explanations? and (3) In what ways do explanations empower users? They also listed a number of specific scalar measures, such as satisfaction, confidence, usefulness, trust, and a number of performance measures, such as decision efficiency and accuracy.

**Degani (2003)**

Useful and interesting examples of what controls and interfaces of electro-mechanical and software systems reveal about their operations.

| Dodge, et al. (2018; see also Kim, et al., 2016). |
|---|
| The researchers focused on the elicitation of mental models of experts. The research was contextualized in the development of explanations of the AI competition bots in Starcraft competitions. At Starcraft tournaments, shoutcasters provide running commentary on games, which is synched with a video of the game play. The researchers systematically studied audio-video transcripts of shoutcast tournaments. The research is a good example of a methodology for eliciting mental models. |

| Doshi-Velez and Kim (2017) |
|---|
| Distinguish explanation from formal interpretability. |
| Describe some desiderata of XAI: fairness, privacy, reliability, robustness, causality, usability, and trust. |
| Propose three different types of taxonomies of evaluation approaches for interpretability: application grounded, human-grounded, and functionally grounded. |
| Hypothesize some task-related and method-related latent dimensions of interpretability so that work across different systems/applications can be compared to help create a more cumulative science. |

| Feltovich, Coulson, and Spiro (2001) |
|---|
| Discuss the "dimensions of difficulty" in understanding complex systems, dimensions that lead to reductive explanations and knowledge shields. |

| Ford, et al. (1993) |
|---|
| Discuss explanation in expert systems as a participatory or "co-constructive" process. |

| Goguen, et al. (1983); see also Clancey (1984b), Weiner (1980). See also Johnson and Lewis, (2018) |
|---|
| Across the late 1970s-1980s there was an emphasis on "dialogue management" and "discourse processes" in the design of intelligent tutoring systems. The explanation system requires "subject knowledge" and "teaching knowledge." Most Intelligent Tutoring Systems (ITS) |

research emphasized that the system needed a method/model for how to interact with the learner. Explanations produce by the expert system should correspond to the ways in which humans communicate explanations.

Explanation is described as the justification of assertions about reasons for decisions or choices, examples, alternatives that are eliminated, or counterfactuals.

Used a tree representation in which internal nodes are embedded and ordered, and mapped onto task goal structure.

**Goodman and Flaxman (2016)**

European Union regulations express the "right to explanation."

The importance of XAI success is brought into sharp relief.

**Hoffman (2017)**

Presents a taxonomy that distinguishes a variety of ways in which human can trust, mistrust, and distrust computational systems. Regardss trusting as a process that emerges in exploration-during-use.

**Kass and Finin (1988)**

Review explanation systems from the 1970-80s. Reviews philosophy of science literature to define and distinguish between explanation, explaining, justification, and understanding.

Propose a way of characterizing GOOD explanations, which includes knowledge about the user's goals and knowledge.

Argue for the need for user models in XAI systems.

Demonstrate how user models can be developed and used.

"Individualized user models are not only important, it is practical to obtain them. A method for acquiring a model of the user's beliefs implicitly by "eavesdropping" on the interaction between user and system is presented, along with examples of how this information can be used to tailor an explanation" (p.1)

| Keil, 2006 |
| --- |
| Review of studies on the benefits of explanation, in research with laypersons, college students, and children as participants. 1. The focus is on self-assessment of one's own gradually accumulating knowledge, especially with respect to the illusion of explanatory depth. |
| Some studies focus on self-rating, but others focus on participant rating of explanations given by Expert explanations are evaluated in some experiments via a final self-rating of current knowledge which occurs after the participant has read the explanation. The effectiveness of explanations is based on these ratings, which demonstrate the degree to which a lay-person could gain understanding from the explanation. |

| Klein and Hoffman (2008) |
| --- |
| Discuss a wide variety of methods and measures for studying mental models. |

| Kulesza, Burnett, et al., 2011 |
| --- |
| Created an explanation system that relied on a domain-specific ontology and a generative grammar to restrict the set of possible user queries, and thence compile appropriate explanations. Their restricted set of nine queries (of the why and why not forms) are resonant to the notion that there is a limited set of "triggers" that motivate users to seek explanations. See Hoffman, et al. (2018). |

| Kulesza, et al (2015). See also Kulesza, et al, 2010; Kulesza, Stumpf, et al, 2011, 2012, 2013 |
| --- |
| Advocate the notion that explanation is a process that continues in the use context. |
| Machine learning is enhanced by allowing the user to offer corrections. Corrections could be implemented by adjusting the Bayesian weights or using the user's reclassification as a new data point for the classifier. |
| Provide a taxonomy of the goodness features of explanations. |
| Their system generates explanations to answer to "why" questions (to satisfy a soundness criterion). |
| Offer a number of guidelines for "explanatory debugging": The ML system must make it |

apparent how it can be useful; changes in how the ML system operates must be made immediately apparent to the user; changes must be reversible if they lead to worse performance. Conducted human subjects studies that illustrate how user mental models can be evaluated and how explanations impact mental models. Mental models were evaluated by giving instances and having participants judge how the machine would classify them and then express their reasons for their judgment. The reasons were scored to reveal features of the users' mental models.

## Lacave and Diez (2002)

The distinction between Structure (domain), Strategy (reasoning), and Support (evidence) as elements of AI system architectures dates to the earliest days of expert systems (see Clancey, 1987). In the same vein, Lacave and Diez (2002) distinguished explanations of the AI's model of the domain from explanations of the AI system's reasoning, and from explanations that reference the evidence for particular determinations.

Provide account of explanation in Bayesian reasoning systems

Propose a set of properties by which to understand and taxonomize explanations.

## Lim and Dey (2009, 2010); and Lim, Dey, and Avrahami (2009).

Identified the types of information end users want applications to provide when explaining their current context, to increase both trust in and understanding of the system. this includes information about the features a classifier uses, information about how the classifier works, and information about how certain the classifier is its classifications.

Describe a toolkit for applications to generate explanations for popular machine learning systems. Emphasis is on providing answers to "why-not" questions as well as to "why" questions.

## Lipton (2016)

Discusses and questions the goal of interpretable models and model explanations. Distinguishes between interpretable, and post-hoc model explanations. Discusses interpretability in terms of transparency, simulatability, and decomposability. Describes some types of post-hoc explanations, including text, visualization, local explanations, and explanation by example.

| **Lombrozo (2006); Lombrozo et al. (2008); Lombrozo (2016)** |
|---|
| Describe research and psychological theory on explanation with a focus on cognitive development.<br><br>Incorporate perspectives from development, philosophy, and cognitive science.<br><br>Argue that explanation involves causal reasoning and abduction. |

| **Miller (2017); see also Miller, Howe, and Sonenberg (2017)** |
|---|
| Lament that XAI systems are built for developers, not users.<br><br>Discuss lack of assessment of systems via studies of human performance.<br><br>Propose methods for assessing XAI.<br><br>Provides a comprehensive review of the philosophical, cognitive, and social foundations in regards to 'everyday' explanation.<br><br>Argues that users are pragmatic in their evaluations of XAI. The most important factor of XAI is not necessarily the validity of the causal cue but rather simplicity, generality, and coherence. |

| **Moore and Swartout (1988)** |
|---|
| Review XAI literature and suggests that previous approaches had been limited. The limitations fall under five categories: narrow, inflexible, insensitive, unresponsive, or inextensible. Limitations are caused from two main system weaknesses: limited knowledge base, lack of a general model of explanation, and inadequate text planning and generation strategies.<br><br>Propose a new reactive approach framework to XAI which takes advantage of user feedback and mirrors more natural language strategies. The Explainable Expert System (EES) provides a conversational explanation to the user based on the user's feedback and its knowledge of the user model. |

| **Mueller and Klein (2011)** |
|---|
| Present a methodology for creating instructional materials explaining AI systems to users. |

**Overton (2013); Miller (2017)**

Overton provides an excellent review of theories of explanation in the Philosophy of Science. Miller presents an integration of philosophical considerations and an excellent capsule view of Overton's general model of (scientific) explanation.

**Ribiero, et al. (2016a, b)**

Describe a system for creating interpretable models for classifiers (both text and image) and presents users with explanations of particular examples (key features used by a classifier). Explanations present selected features with the highest positive weights with the goal of "giving an intuition" as to why the classifier would think that a class is present. For image classification, the selected features are used to generate cropped versions of the images that illustrate the identified class or classes (e.g., parts of the image that have the highest weights for the class "dog").

Present an experiment in which human users are tasked with (1) choosing which of two classifiers performs better, based on a comparison of the explanations that are provided about how the classifiers work, (2) and identify classifier anomalies based on the provided explanations, and (3) explain the spurious correlations that led to the misclassifications.

**Ritter and Feurzig (1988)**

Describe an intelligent tutoring system for teaching tactical radar operation. The trainee interacts with a simulated fighter pilot and the tutor system sends alerts to the trainee based on its ability to detect trends (e.g., the potential loss of a radar track, immanent loss of a speed advantage). A simulated "expert" explains its decisions and actions. An event replay with expert narratives provides explanations of where the trainee made errors.

**Schaffernicht and Groesser (2010); see also Besnard, Greathead and Baxter (2004) and Doyle, et al. (2008)**

Review methods for coding and comparing user mental models. Although their review focuses on dynamical systems in which the coded propositions refer to feedback loops (both closed and interacting), analysis in terms of entities and feedback loops can be understood generally as the analysis of variables and functions expressed as directed graphs. Mental models are compared in terms of the number of shared concepts, number of shared relations, and number of shared propositions.

Mathematical schemes for the comparison process are presented.

**Schank (1989)**

Presents many examples of how people explain various everyday things and events. Presents models of the explanation processes (ninety "explanation patterns").

Discusses SWALE, a program that generated explanations by: (1) detecting anomalies, (2) searching for an explanation pattern that best matches the goals and needs of the recipient of the explanation, then (3) creating a new explanation from archived instances of the selected explanation pattern.

**Sheh and Monteath (2018)**

Argue why the requirements for explanations depend on both the application and the nature of the machine learning technique employed by the AI.

Present a taxonomy of explanation types in terms of what they are useful for (e.g., "rationalization" explanations are useful for fault detection tasks).

**Sørmo (2005)**

Presents a comprehensive review of philosophical and cognitive accounts on explanation in intelligent systems.

Reviews five important goals for XAI systems: transparency, justification, relevance, conceptualization, and learning.

Evaluates the different Case Based Reasoning (CBR) methods of explanation according to the five identified goals.

Argues for the value of CBR type explanations.

Provides detailed discussion of past attempts to taxonomize explanations.

## Southwick (1991)

A good review of the development of explanation capabilities in first and second generation systems (knowledge-based systems; see Section 5 of this Report). He pointed to the psychological literature on human discourse and explanation (i.e., abductive inference) to highlight issues that he predicted would become important, such as user-centered analysis and user model-based customization of explanations.

He distinguished between explanations about the content and structure of a knowledge base, explanations of the strategy used by the problem solver, and justifications of the system in terms of its underlying causal model.

"While reasoning trace explanations rely only on the formulation of knowledge that comprises the knowledge base, the other two explanation types require additional knowledge. A system that can give strategic explanations must be able to reason about its own activity, which may require knowledge about the ordering of problem solving tasks, for example. Deep explanations obviously require a great deal of information in the form of a causal model of the domain" (p. 3).

"An explanation system that has access to a deep model can provide explanations that are intuitively more satisfying, since they relate to the deeper concepts that underlie the domain model" (p. 6). (See also Clancey, 1983; Boy, 1991.)

## Swartout (1981)

Pioneering attempt to enable expert systems to generate human-understandable (local) explanations, by modeling the procedural rule hierarchy as a set of abstract goals and reasoning strategies and using that to generate an "audit trail" of expert reasoning.

| **Swartout and Moore (1993)** |
|---|
| Retrospective account of explanation in expert systems from 70s-80s. |
| Review key differences between first and second generation XAI systems. |
| Include desiderata for explanations in intelligent systems. |
| Distinguished between justification and explanation. |
| Distinguished between first-generation systems that provided direct explanations of rules/concepts, and systems that place actions in context and engage in an interactive dialog with the user |

| **Tullio, et al. (2007)** |
|---|
| Illustrate a method for eliciting user mental models (structured interviews), scoring the protocols for data analysis (frequency counts), and tracking changes in mental models across practice. |

| **Wick and Thompson (1992), see also Clancey (19864a, 1986a)** |
|---|
| Presented a taxonomy of explanation that combines goals, audience, and focus (global-local). |
| Argued that the existing explanation systems provide descriptions of the system's problem solving that are justifications appropriate for knowledge engineers rather than explanations for users. |
| Argued that explanation must be a collaborative problem solving process, a process utilizing procedural rules ("decoupled reasoning graphs"), which is distinct from the reasoning process. |
| Proposed the inclusion of a knowledge-based explanation system separate from the knowledge-based problem solving system. "Explanation is not an add-on to the expert system's reasoning, but a problem solving activity in its own right" (p. 37). Their primary goal, like that of Mycin, was to create a system that can tell explanatory stories that organize information in a coherent flow from data to conclusion. |

| **Walton (2004, 2007, 2011)** |
|---|
| Argues that explanation systems in AI need to be founded on the dialectical concept, that is, explanation is a process of dialog to achieve shared understanding. <br><br> Advocates for the use of speech act theory to develop formal explainable systems. |

## 5. Explanation in Artificial Intelligence: An Historical Perspective

*Much of the contemporary moment's enthusiasm for and commercial interest in artificial intelligence, specifically machine learning, are prefigured in the experience of the artificial intelligence community concerned with expert systems in the 1970s and 1980s. ... artificial intelligence communities today have much to learn from the way that earlier communities grappled with the issues... (Brock, 2018, p. 3).*

## Introduction

A number of computer science reports and review articles examine explainability in AI, and propose what are believed to be good or useful explanation types and formats. It is generally held that explanations must present easy-to-understand "coherent stories" in order to ensure good use of the AI or good performance of the Human+AI work system.

(Biran and Cotton, 2017; Brézillon, 1994; Brézillon and Pomerol, 1997; Buchanan and Shortliffe, 1984a; Clancey, 1986a; Core et al., 2006; Craven and Shavlik, 1996; Doshi-Velez and Kim, 2017; Doyle, Tsymbal, and Cunningham, 2003; Ellis, 1989; Fox, Long, and Magazzeni, 2017; Goguen, Weiner, and Linde, 1983; Hoffman and Klein, 2017; Hoffman, Mueller, and Klein, 2017; Kass and Finin, 1988; Lacave and Díez, 2002; Miller, 2017; Miller, Howe, and Sonenberg, 2017; Miller et al., 2017; Moore and Swartout, 1988; Moulin, Irandoust, Bélanger, and Desbordes, 2002; Mueller and Klein, 2011; Sørmo, Cassens, and Aamodt, 2005; Swartout and Moore, 1993; Weston et al., 2015; Wick and Thompson, 1992).

There were three distinct generations of what might be collectively or broadly called Explantion Systems. The original wave of expert systems that emerged in the 1970s involved representations of knowledge using rules and networks of relationship—sometimes probabilistically. The systems included a logic to ask the user questions (e.g., the result of blood tests), and would make a recommendations, predictions, or diagnoses based on the data and the rules.

One use of expert systems that was discovered very early is that in addition to being with being a decision aid, expert systems might help novices understand the knowledge, and with small variations could serve as tutors, digital mentors, or digital assistants.

**First Generation Systems: Expert Systems**

The first and second generations of Explainable AI systems were called expert systems. Applications included decision aiding and diagnosis. These began as systems that expressed their internal states, and grew to be systems that were more interactive, context-aware, flexible, and had some ability to offer explanations and justifications that the users wanted and would accept. Research showed that explanations of how computer systems work can convince users to adopt the system recommendations, but also that users will not necessarily be satisfied with their decisions (see Bilgic and Mooney, 2005). Research on how expert systems could explain their analyses and decisions also clarified the distinction between explanation of how a system works that makes sense to users versus the justification of a system architecture or process that one AI researcher would present to another (see for instance, Chandrasekaran, Tanner and Josephson, 1989). These are referred to, respectively, as global explanation versus local justification.

In this First Generation era, there was substantial research attempting to show that expert systems that provided and explanations had greater impact on reasoning (cf. Dhaliwal and Benbasat, 1996; Eining and Door, 1991; Murphy, 1990), with the effect depending on the skill level of the users (Lamberti and Wallace, 1990). But many of the first generation expert systems failed to have the hoped-for benefits. In work on medical expert systems, researchers recognized that physicians would ignore advice from an expert system unless a rationale (a justification) for why the advice is being given were provided. Initial explanation systems attempted to provide this rationale by describing the underlying goals and steps used to make the diagnosis. This approach, which Swartout and Moore (1988) called the "Recap as explanation myth," was also inadequate, and this led to a rethinking of the approaches, goals and purposes of XAI systems. This first-generation began in the late 1970s (e.g., Shortliffe, 1976), and continued for about a decade.

 (Buchanan and Shortliffe, 1984a, 1984b; Chandrasekaran and Swartout, 1991; Chandrasekaran, Tanner, and Josephson, 1989; Clancey, 1983, 1984a, 1986a, 1992, 1993; Druzdzel, 1990; Druzdzel and Henrion, 1990; Ford et al., 1992; Gautier and Gruber, 1993; Goguen et al., 1983; Greer, Falk, Greer, and Bentham, 1994; Gruber, 1991, 1991; Hasling, Clancey, and Rennels, 1984; Hayes-Roth, Waterman, and Lenat, 1984; Henrion and Druzdzel, 1990a, 1990b; Kass and Finin, 1988; Langlotz and Shortliffe, 1989; McKeown and Swartout, 1987; Merrill, 1987; Moore and Swartout, 1988, 1990, 1991, Neches, Swartout, and Moore, 1985a, 1985b; Scott, Clancey, Davis, and Shortliffe, 1977; Scott, Clancey, Davis, and Shortliffe, 1984; Shortliffe, 1976; Silverman, 1992; Southwick, 1988, 1988, Swartout, 1977, 1981, 1983, 1985; Swartout and Moore, 1993; Swartout, Paris, and Moore, 1991; Wallis and Shortliffe, 1981, 1984; Weiner, 1980; Wick and Slagle, 1989; Wick and Thompson, 1992; Wood, 1986).

The inadequacies of the first generation of expert systems gave birth to the first generation of explanation systems, including Mycin and its related systems (see Clancey, 1986a, for a detailed history), the Digitalis Therapy Advisor (Swartout, 1977); BLAH (Wiener, 1980), and other special-purpose explanation systems. Although they employed a variety of methods, these systems often would work by creating expressions of the logical and probabilistic rules used to make a diagnosis or answer a question. In general, because knowledge and expertise was framed in terms of rules, those rules could have natural-language descriptions that could be derived from their logical form, or written as 'canned' text by experts. A simple explanation would express the rules used to make a decision. Such explanations were typically written in readable English, but often they were simple translations from underlying if-then production rules in LISP to text descriptions.

Such systems would typically operate in question-answering mode, where the data about the current case would be entered based on queries of the system. Explanations could be generated both during data entry and as a post-diagnostic analysis. As an example of explanation during data entry, consider the operation of XPLAIN (a successor to the Digitalis Advisor) described by Swartout (1985). This system might ask the user to enter the results of a particular blood test (What is the level of Serum Calcium?). Upon seeing the question, the user could ask "Why?" In response, the system would give a justification for why the information is needed, which might be helpful ("My goal is to begin therapy. One step in doing that is to check sensitivities. I am now trying to check sensitivity due to calcium"). This illustrates how these explaining systems explained what they were doing—providing an explanation, but also playing the role of a tutor for less experienced users.

In fact, the clarity provided by these systems was sometimes discussed as an asset for system developers and knowledge engineers, and not just users (e.g., Wick and Thompson, 1992). Consequently, many of these systems, which were developed to fill a need of end-users, may have ended up being of greater value to developers. (Perhaps also related to the fact that they were mostly research vehicles that were not put into practical use.)

Swartout and Moore (1993) described the first generation systems as ones that created explanations by paraphrasing the rules that were used to come to a decision. As a whole, the first generation expert systems and explanation systems were focused on expressing machine-internal states, goals, and plans. This was sometimes fairly straightforward—because the rules themselves were formalizations (from verbal protocols) of the rules used by experts themselves. Thus, translating back to natural language could be relatively straightforward. Sometimes, however, the language descriptions would bear scant resemblance to human explanation, or natural language at all. First, they depended on the if-then format of logical and causal rules, rather than providing an explanation at a higher-level reasoning strategy (e.g., gathering basic information about the patient, casting a net for alternative explanations, trying to support a specific hypothesis). Second, some of the rules were logically required needed to make the system work and might not necessarily make sense to users or even be pertinent to a user's "why" question. Third, domain knowledge was sometimes 'compiled out' (e.g., causal relations between symptoms and diagnoses), and so it was not included in an explanation.

Several researchers (e.g. Teach and Shortliffe, 1981; Clancey, 1983; Moore and Swartout, 1988; Sørmo, et al., 2005) noted that initial attempts to develop expert systems failed at adoption because the expert users did not understand the reasoning used to make the recommendation or diagnosis. For Mycin the main issues in 1975-1980 were that: 1) there were no patient databases at the time, so data had to be entered manually; 2) there were no computer terminals in the clinics or on the wards; 3) the consultation dialogue took 20 minutes, which was too long. These systems had the main goal of being accurate and replacing or augmenting the experts such as physicians, but this goal may have been unrealistic, and was perhaps misguided. Moore and Swartout (1988) noted that of 15 desired abilities of such medical diagnostic systems, 'never make an incorrect diagnosis' was ranked 14th most important, but 'explain their ...decisions to physician users' was the most important. These systems were designed to solve the problem of diagnosis, but this is not the problem that physicians needed help with (cf. Yates, Veinott, and Patalono, 2003). The main justification for Mycin design was that MDs were prescribing antibiotics without making a diagnosis first, particularly without a culture. They wanted help justifying and explaining why the diagnosis was made, or at least required explanation if they were to trust the diagnosis. The idea was to force them to think through the symptoms and likely bacteria. Hence they prescribed broad-spectrum drugs that were feared would lead to resistant bacteria.

One limitation of first generation systems was that they could not justify inferences—why in a certain situation a particular interpretation (e.g., relation of symptom and disease) or action (e.g., taking an antibiotic for two weeks) was correct. Although rule-based systems could present the current goal (e.g., to determine the type of infection) and conditions of a rule in an explanation, why the rule was correct was not represented. As a clear example, consider the Mycin rule, "If the patient has an infection and lives in the San Joaquin Valley, the type of infection may be fungal." How living in the San Joaquin Valley causes an infection is not included in Mycin's model of diseases and therapies. Furthermore, the clauses in the "if" part were often ordered; for example, in the above rule, the second clause tests for evidence of a subtype of the disease category (is it fungal?) established by the first clause (there is evidence of an infection).

When Mycin was originally developed, the inability to explain the implicit design of rules and their justifications was not deemed a deficiency because producing any explanations at all in a readable form from a recorded trace of the program's reasoning was already challenging and a substantial advance in AI capability. Furthermore, such associational models were "rules of thumb," based on demographics and physiology that would be generally familiar to the users, who it was thought would simply follow the program's advice once it had been tested and certified by experts. However, when attempting to extend or refine these rule-sets in "knowledge acquisition," or to use them for teaching, the inability to explain these high-level associations became of concern.

Contemporaneous with Mycin, other researchers developed medical expert systems using different formalisms, such as probability models and semantic networks. For example, the OWL semantic network framework was used to explicitly represent a disease hierarchy and symptom-disease causality; these relations could then be presented in the "trace" of the program's reasoning. Nevertheless, it is not possible to include in the model why every heuristic association is correct (i.e., why it works in practice)—causal relations may be poorly understood (e.g., how experimental cancer treatment drugs affect the body) or involve a great deal of scientific detail (e.g., how an antibiotic affects a living cell). Both Swartout and Clancey analyzed the limitations of the original expert system explanation systems very clearly in their 1979 dissertations, and Swartout reiterated it in 1985.

Moore and Swartout (1988) went further, presenting explanation as a problem-solving process that involves text planning and natural language discourse, not merely playing back chains of reasoning that relate data and interpretations. Rather than just reading out what happened inside the program, explanation is viewed as a structured interaction that takes into account the user's objectives, background, and the context:

> *Explainers of the future must have strategies that are based on those observed in naturally-occurring text. These strategies must be flexible so that explanations can be tailored to the needs of individual users and specific dialogue settings. Explainers must have alternative strategies for producing responses so that they may provide elaboration or clarification when users are not satisfied with the first explanation given. Furthermore, the system must be able to interpret follow- up questions about misunderstood explanations or requests for elaboration in the context of the dialogue that has already occurred, and not as independent questions. (p. 3)*

As researchers began pondering the inadequacies of the systems, a 'second generation' began to emerge (Swartout and Moore, 1993).

## Second Generation: From Expert Systems to Knowledge-Based Tutors

By the mid-1980s, the limitations of the first generation of explanation systems were being confronted (e.g., Moore and Swartout's [1988] Recap-as-explanation Myth). The realization was that it was not enough to simply summarize the inner working of the system; this did not make for a sufficient explanation or justification. The text produced might be true/valid; it just isn't necessarily what the user wanted to know, or (worse) was not comprehensible.

Second generation systems frequently blurred the line between a consultation program (cf. Clancey, 1986a), a tutor (cf. Clancey, 1986b), a recommender (Sinha and Swearingen, 2002), and a data entry system (cf. Gruber, 1991), but researchers had the goal of going beyond the methods that had failed to improve understanding or adoption of the expert systems. A primary driver of second generation systems was the need to develop more abstract frameworks that would promote reusability and ease of development. NEOMYCIN did that in providing a diagnostic procedure (tasks/metarules) and associated language of "dysfunction" taxonomic and causal relations (Clancey, 1984, 1986a).

It was recognized early during the development of explainable expert systems that the same knowledge base, rules, and explanations could be used to develop intelligent tutors. The distinguishing characteristic of intelligent tutoring systems is that they inferred each student's mental model of the domain (their "knowledge base") from the student's behavior. These systems were first collectively called "Intelligent CAI," to distinguish intelligent tutors from the simple teaching machines of the 1960s (and pursued by the "AI in Education" community particularly in the 1980s and early 1990s).

For example, in GUIDON (related to English "guide"), the student's request for patient data and his stated hypotheses were used to work backwards through MYCIN's rule network (or whatever EMYCIN network was being used) to determine what rules were not being considered. The explanation was a network of inferences and sometimes included uncertainties (e.g., evidence for a student knowing a rule based on prior interactions had to be distinguished from applying it in a particular case). This is why the systems were sometimes called "knowledge-based tutors," contrasting them with the teaching machines of the 1960s. The main ideas were that the domain knowledge of the expert system (e.g., medical diagnosis rules) was separate from the tutoring knowledge base (e.g., dialogue management rules). Also, an interpretative process of "reading the domain rules" analogous to explaining the expert system's behavior from a model ("trace") of its internal processes was used to explain the student's behavior by constructing a model of how the student was reasoning.

Intelligent tutoring systems emerged in parallel with the emergence problem solvers that conducted "case-based" reasoning. In a sense these incorporate human explanation (see Doyle, Tsymbal, and Cunningham, 2003) and might be considered second generation explanation systems. On the other hand, some intelligent tutoring systems would not qualify as second generation systems. This is because they relied on "student models" that were not models of users or students at all, but just equations that tracked the kinds of topics and test items that users got correct and incorrect in order to select additional material and practice problems. The user (student) models contained nothing that was at all like knowledge or meanings, and had no process that might be said to instantiate reasoning. The situation was different for knowledge-based tutoring systems. In fact, what might be seen as the very first intelligent tutoring system, Carbonell's Scholar (Carbonell, 1970), was a Socratic tutor that carried on a meaningful dialogue with the learner about the causes of rainfall. The student models were called "process models" because they didn't simply tick off right or wrong problem records, but inferred the argument (or procedure) used to derive the student's answer.

In intelligent tutoring systems generally, the knowledge base is used by a tutorial program to solve the problem (e.g., answer a question, diagnose a circuit, make a math calculation) and derive the problem with the student's answer. The modeling process might involve deleting part of the expert derivation, or applying some variant of the tutor's knowledge, such as a faulty causal relation.

Gregor and Benbasat (1999) reviewed a myriad of empirical studies on explanations in Knowledge Based Systems. They advocate for explanations that conform to Toulmin's model of argumentation (addressing claims, grounds, warrants, backing, qualifiers, and possible rebuttals creates strong explanations). Their Appendix A lists numerous studies cited as empirical evidence for the conclusions it is synthesizing.

We have identified a number of citations from the very large body of tutoring and training research to provide some pointers to XAI-relevant research.

(Aleven and Koedinger, 2002; Anderson, Boyle, Corbett, and Lewis, 1990; Anderson, Corbett, Koedinger, and Pelletier, 1996; Bellman, 2001; Bengio, Louradour, Collobert, and Weston, 2009; Boy, 1991; Burns, Luckhardt, Parlett, and Redfield, 2014; Cao et al., 2013; Chandrasekaran and Swartout, 1991; Clancey, 1981, 1984b, 1986b, 1987, 1988; Clancey and Letainger, 1984; Corbett and Anderson, 1991, 2001; Doignon and Falmagne, 1985; Fonseca and Chi, 2011; Forbus and Feltovich, 2001; Frederiksen, White, Collins, Allan, and Eggan, 1988; Guerlain, 1995; Haddawy, Jacobson, and Kahn, 1997; Katz and Lesgold, 1992; Kehoe, Stasko, and Taylor, 2001; Lajoie, 2009; Lane et al., 2005; Lesgold et al., 1988; Loksa et al., 2016; Mark and Simpson, 1991; McGuinness and Borgida, 1995; Merrill, 1987; Miller, 1988; Moray, 1987a; Mueller and Klein, 2011; Nakatsu, 2006; Nakatsu and Benbasat, 2003; Polson and Richardson, 1988; Psotka, Massey, and Mutter, 1988; Ritter and Feurzeig, 1988; Silverman, 1992; Sleeman and Brown, 1982; Southwick, 1991; Swartout, Paris, and Moore, 1991; Tanner and Keuneke, 1991; Tintarev, 2007; Woolf, 2007; Woolf, Beck, Eliot, and Stern, 2001)

## How Explanations Were Constructed in Second Generation Systems

Chapters in Sleeman and Brown's (1982) Intelligent Tutoring Systems make it clear how second generation systems such as GUIDON constructed arguments, sometimes generating multiple explanations, and how the systems accounted for the student's recent behaviors and claims (e.g., SOPHIE by Burton and Brown, 1982). Second generation systems, which often focused on making explanations context-sensitive. Here, context included knowledge about the person, about the history, about the goals of the user, and about the domain. Some second generation systems were also self-improving. Many of the systems were proof-of-concept or research systems (with the exception of some of the tutoring systems).

Perhaps the most mature collection of papers on intelligent tutoring systems is the 1988 volume edited by Psotka, Massey, and Mutter, ITS: Lessons Learned. Some of these were aimed directly at the recap-as-explanation myth, and others helped support explanation and explainability in others ways. Some of these ways include:

## *Probabilistic, Bayesian, and Case-Based Reasoning Approaches*

Following the evolution of expert systems into intelligent systems in the 1980s, less research was conducted on explanation in AI systems. It did find a foothold in several sub-disciplines throughout the 1990s and early 2000s, including some work on case-based reasoning, and probabilistic/Bayesian approaches., e.g., determining the topics of a paper given a list of keywords.

(Chajewska and Halpern, 1997; Chan and Darwiche, 2012; Chang, Gerrish, Wang, Boyd-Graber, and Blei, 2009; Doyle et al., 2003; Druzdzel, 1990, 1996a, 1996b; Druzdzel and Henrion, 1990; Eaves Jr and Shafto, 2016; Fox, Glasspool, Grecu, Modgil, South, and Patkar, 2007; Freuder, Likitvivatanavong, and Wallace, 2001; Haddawy, et al., 1997; Henrion and Druzdzel, 1990a, 1990b; Herlocker, Konstan, and Riedl, 2000; Khan, Poupart, and Black, 2009; Kisa, Van den Broeck, Choi, and Darwiche, 2014; Kofod-Petersen, Cassens, and Aamodt, 2008; Lacave and Díez, 2002; Letham, Rudin, McCormick, Madigan, and others, 2015; Liang and Van den Broeck, 2017; Nugent, Doyle, and Cunningham, 2009; Papadimitiou, Symeonidis and Manolopoulos, 2012; Pearl, 1988; Suermondt, 1992; Shafto and Goodman, 2008; Shafto, Goodman, and Griffiths, 2014; Sinha and Swearingen, 2002; Sørmo et al., 2005; Theis, Oord, and Bethge, 2015; Wang, Ruin, Velez-Doshi, Liu, Klampfl, and MacNeille, 2010a,b; Yap, Tan, and Pang, 2008).

## *Applications with Explanation Capabilities and Context-Aware or Ubiquitous Computing*

The area of context-aware and ubiquitous computing made strides in integrating an explanation capability into human-facing software systems. Such systems may have only modest intelligence (e.g., an auto-pilot), but the human-machine interaction design elements for systems that understand the context of the environment and the user certainly overlap with explanation-based systems. This area of research has seen important for several decades, but some references that appear most relevant to XAI are: Belotti and Edwards (2001); Chalmers and MacColl (2003); Lim and Dey (2009, 2010); and Lim, Dey, and Avrahami (2009).

## *Autonomy, Agent-based Systems, and Reinforcement Learning*

Another early follow-on to expert explanation systems was agent-based simulations. This area of research continues in various forms today, including schemes for the design of agents to provide different kinds of explanations (e.g., ontological, mechanistic), agents for human-robot interaction (explicability of AI-generated task plans) and transparent agents for strategy game-play supported by deep networks.

(Biran and Cotton, 2017; Bogomolov, Magazzeni, Minopoli, and Wehrle, 2015; Bogomolov, Magazzeni, Podelski, and Wehrle, 2015; Chai, She, Fang, Ottarson, Littley, Liu, and Hanson, 2014; Chai, Fang, Liu, and She, 2016; Chakraborti, Sreedharan, Zhang, and Kambhampati, 2017; Dodge t al., 2018; Fox et al., 2017; Frank, McGuire, Moses, and Stephenson, 2016; Hayes and Shaw, 2017; Haynes, Cohen, and Ritter, 2008; Jain, Zamir, Savarese, and Saxena, 2016; Johnson, 1994; Kim, Chacha, and Shah, 2013; Lane, Core, Van Lent, Solomon, and Gomboc, 2005; Langley, Meadows, Sridharan, and Choi, 2017; Lomas et al., 2012; Metoyer et al., 2010; Mnih et al., 2015; Moulin et al., 2002; Rosenthal, Selvaraj, and Veloso, 2016; S. Singh, Lewis, Barto, and Sorg, 2010; Stubbs, Hinds, and Wettergreen, 2007; Van Lent, Fisher, and Mancuso, 2004; Zahavy, Ben-Zrihem, and Mannor, 2016; Zhang et al., 2016; Zhou, Khosla, Lapedriza, Oliva, and Torralba, 2016)

## *Movement from Explanation to Justification*

The distinction between 'explanation' and 'justification' seems sharp, although it is possible to think about a justification as a type of explanation. Nevertheless, early explanation systems focused on making the justification visible to the user, but not the explanation. Swartout's (1983) XPLAIN system was one of the first to make this distinction, especially with respect to justifying a recommendation based on background knowledge. It should be noted that justifications can also be deceptive, as they may attempt to convince the user to believe a result, rather than justify in terms of the background knowledge that fed into a decision. An argument from authority ("This system was generated based on the input of the best physicians in the country") is a justification that may support adoption but may not be genuine. In that sense, justifications may sometimes focus on establishing trust or ensuring reliance rather than informing users about the both the strengths and limitations of the system, and thereby fail to help the user develop a mental model of AI the process. (This was a key point of STEAMER; which was a tool intended to help end-users [experts] build their own knowledge bases; see Hollan, Hutchins and Weitzman, 1984).

*Improved Explanation Language*

Systems such as EES (Explainable Expert System; Neches et al., 1986ab), which was the successor to XPLAIN, were intended to move past inadequacies of the earlier systems by representing more directly the knowledge that went into the decision system in the first place. In a sense, both the expert system and the explanation system were generated from the same high-level knowledge specification. Other systems (e.g., REX; Wick and Thompson, 1992) built secondary explanation systems to 'reconstruct' an argument that was more understandable. Although natural language capabilities have been part of XAI since the early days (McKeown and Swartout, 1997; Selfridge, Daniell, and Simmons, 1985; Moore and Swartout, 1991), as natural language systems have improved, they have continued to be implemented as means for creating explanations and interactions with intelligent systems (e.g., Barzilay, McCullough, Rambow, DeCristofaro, Korelsky, and Lavoie, 1998; Chai, Fang, Liu, and She (2016); Dodson, Mattei, and Goldsmith, 2011; Henrion and Druzdzel (1990ab), Lazaridou, Peysakhovich, and Baroni, 2017; Neihaus and Young, 2014; Pace and Rosner, 2014; Papamichail, and French, 2003). Furthermore, the apparent ease with which natural language explanations can be created using recurrent neural networks (cf. Karpathy 2015) has led to the rise of visual question-answering systems (e.g., Hendricks et al, 2016), and deep networks whose explanations are embedded within the systems, for example, by interpreting feature weights as a measure of salience (e.g., Lei, Barzilay, and Jaakkola, 2016; Li, Chen, Hovy and Jurafsky, 2015; and many more).

## *User Models*

The value of user models is demonstrated in the knowledge-base tutoring systems which have achieved noteworthy success in training contexts (e.g., Lesgold, Lajoie, Bunzo, and Eggan, 1988; see also Chapter 9 in Hoffman, et al., 2014). Gott (1995) conducted a test of the effectiveness of SHERLOCK, an intelligent tutor for the debugging of electronic circuits. Senior apprentices having three years' experience received the SHERLOCK training, and on a subsequent test they outperformed a group of practitioners having had 10 or more years' experience. Apparently, with SHERLOCK apprentices were able to acquire the strategic knowledge of highly proficient technicians. Approximately 25 hours of Sherlock training was the equivalent of four years of on-the-job training. Gott attributed the dramatic effect to the "situatedness" of the training, the ecological validity of the scenarios and the "learn-by-doing" approach.

Systems theorists Conant and Ashby (1970), argued that any good regulator of a system must be a model of the system. Here, a regulator is a system that is trying to control another system (such as a thermostat). An explanation system is essentially a regulator of the human-AI system; thus it must include within itself a model of that human-AI system in order to succeed.

Kulesza, et al. (2013) demonstrated clearly how explanations influence users' mental models of software systems. Several researchers have advocated or used 'user models' in their explanation systems. In their development of the GUMAC system, Kass and Finin (1988) assumed that an explanation must be appropriate and understandable for a particular user. They developed a means of tailoring explanations based on five principles: appropriateness, economy, organization, familiarity, and processing requirements. They argued that a system that understands how these might differ across individuals would have a better chance of success. They acknowledged that user models may take a variety of forms, and can be models of individuals or can be fairly generic.

Two versions of user models are ones that consider a user type (e.g., an expert versus a novice), and ones that consider the knowledge of an individual—a model of their goals, knowledge, and history of interaction with the system. Wallis and Shortliffe (1984) implemented general user models that included self-rated expertise and amount of preferred detail, so that explanations would be tailored to these levels. Greer et al. (1994), Moore and Swartout (1990), and Sarner and Carberry, (1992) described ways of providing tailoring or individualized explanations, and the selection of rhetorical strategies, based on user models or other inferences. User models have been used to support the "tailoring" of explanations in a number of ways (cf. Carberry, 1990 for a discussion of a 1990 workshop; see also Kay, 2000; 2006, Kieras and Polson, 1985; May, Barnard, and Blandford, 1993; Maida and Deng, 1989; Weiner, 1989)

## *User History or Knowledge State*

One particular style of user modeling involves inferring the knowledge or goal state of the individual user, and tracking it in order to provide better and more tailored explanations. In contrast to first generation systems, which simply restated the reasoning rules, a number of advantages can come from tracking a particular user and his or her interaction history. First, redundant information can be avoided by knowing whether certain information has already been given. Also, particular states of knowledge can better be determined, in order to tailor new information to the user. Systems with this style of user modeling are akin to many intelligent tutoring systems, which involve a student and/or knowledge model that tries to identify the current knowledge state and thereby provide learning experiences that moves students to a new state. Scott et al. (1984) described one example of user history, and Biran and Cotton (2017) examined how personal histories might be useful in making and explaining recommendations.

## *Context and Goals*

Part of the importance of user models and history is to tailor to system to the goal(s) of the user. But this requires a specification of the goals. Several possible goals include affording knowledge engineers transparency during development of the system (Wallis and Shortliffe, 1981), and verification, duplication, and ratification (Wick and Thompson, 1992) explaining how a system reached an answer, justifying why an answer is good clarifying concepts, and teaching a user about the domain (Sørmo, 2005). A similar point was made by Brézillon (1994), who advocated the need for the understanding of context in explanation systems. Brézillon emphasized that explanation is a cooperative problem solving exercise between the user and the system, and that each needs to understands the goals of the other to succeed.

*Contrastive Explanation, Counterfactual Explanation, Comparative Reasoning, Forgone Alternatives*

During this second generation, researchers also began proposing that effective explanations describe why choices were not made, depict assessments of alternative courses of action, or infer that a specified behavior differs from typical and should be described specifically. This involves contrastive reasoning, which has been shown to be effective in human learning (see Chin-Parker and Cantelon, 2016; Williams and Lombrozo, 2010). Such functions actually appeared in first generation systems—the Digitalis Advisor (Swartout, 1977) included means of describing how recommendations changed as a result of information; and BLAH (Wiener, 1980) involved logic about alternative options in decision making. MYCIN's explanation system allowed questions of the form "Why didn't you ask about X?" and "Why didn't you conclude Y?" GUIDON's therapy module allowed the student/MD to enter a drug regimen and have it compared to the expert system's ("critiquing model").

These explanation modes became more fully explored in the second generation. One example of this is described in a series of papers by Druzdzel and Henrion (Druzdzel, 1990; Druzdzel and Henrion, 1990; Henrion and Druzdzel, 1990ab), Explanation systems were developed by the use of 'scenarios.' Scenarios involved describing the likelihood of the best option, but also pointed to less-optimal alternative option scenarios. In fact, the importance of contrast cases is so clear that later researchers argued that "all why-questions ask for contrastive explanations, even if the foils are not made explicit." (Miller, 2017). Wachter, Mitelstadt, and Russel (2017) argued that a type of contrast, counterfactual reasoning, may form a basis for evaluating the transparency or fairness of machine decisions.

*Concept Mapping*

Many of the new developments of the second generation (improved explanation, context and goal relevance, contrastive explanation) were pulled together in NUCES, a project aimed at building an expert system for diagnosis of heart function using the then-new method of first-pass radionuclide angiography (Ford, et al., 1992, 1996). The method for eliciting expert knowledge was Concept Mapping, a process in which the expert articulates their knowledge in the form of propositions (triples), which are then composed as a directed graph. The propositions were subsequently used to create the inference engine for the expert system. At that point the researchers' insight was that expert system did not need a traditional explanation capability, i.e., a means for explaining the rules that were applied in any particular case analysis. Rather, it was realized that the concept maps themselves would serve not only as an interface for the user to interact with the expert system, but at the same time provide a representation of the expert's own explanations of cases. The Figure below is a screen shot showing one of the Concept Maps and some of the digital media that were appended to nodes in the Concept Map. These resources included text in which the expert described the diagnosis for different cardiac issues, brief videos that enabled the user to "stand on the shoulders" of the expert, and arrays of exemplars showing clear cases and marginal cases.

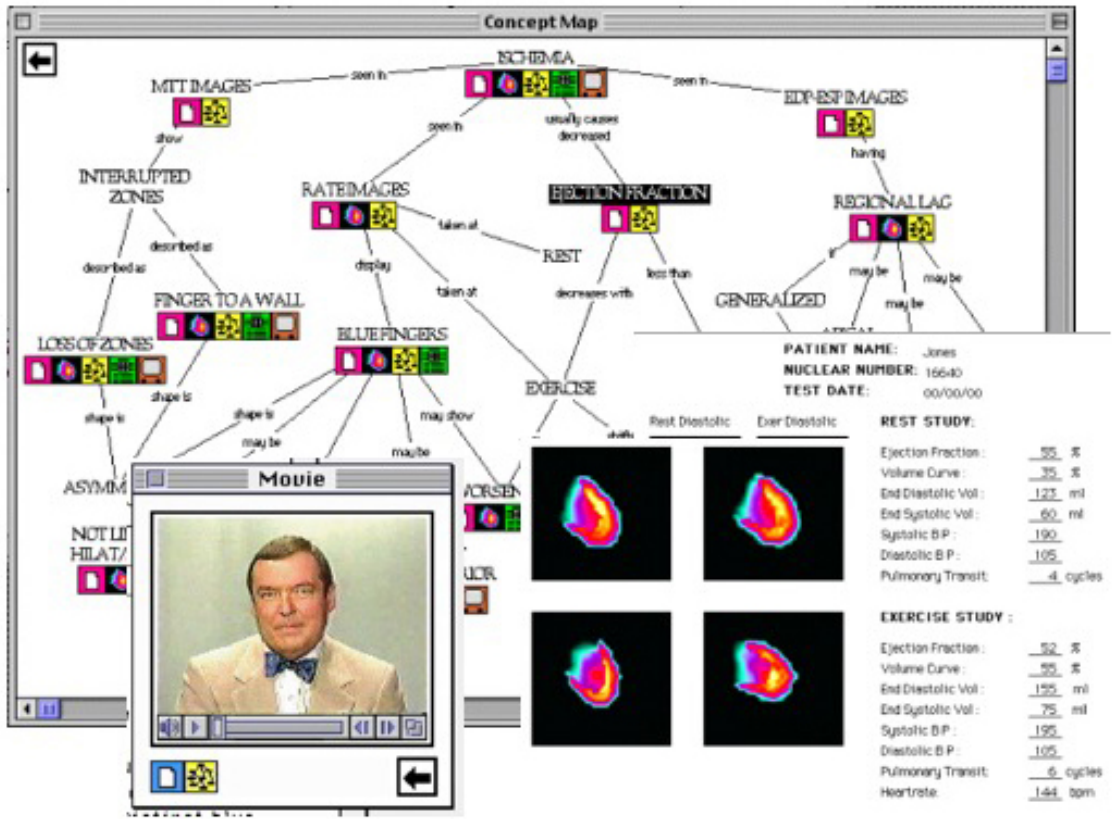

**Figure 5.1. A screen shot showing one of the Concept Maps in NUCES and some of the hyperlinked resources.**

## The Explainability Winter

The decline of expert systems involved its morphing, becoming the general class of "intelligent systems." There were roughly two decades of an 'Explainability Winter' in which little new research was produced, although some researchers translated the earlier methods to new applications. Many of these explanation systems were created for the same reasons that the first generation systems were created—to help researchers and stakeholders understand them, and to explain and justify behaviors to users and decision makers. Figure 5.2 shows a count of relevant publications we have identified regarding explanation in intelligent systems.

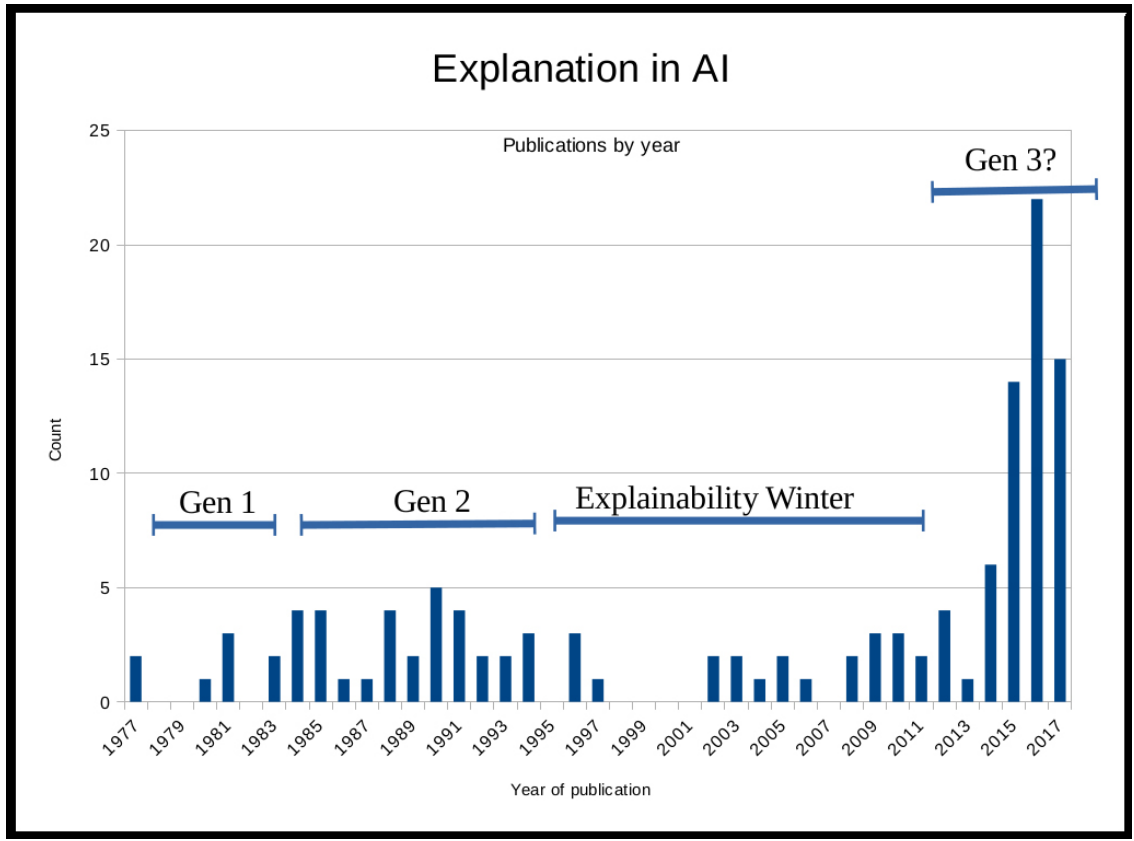

**Figure 5.2. Histogram of the number of publications per year identified in our literature review that are relevant to explanation in intelligent systems (AI, machine learning, and related fields).**

From the early 1990s through the early 2010s, research on explainable systems followed the waning fortunes of expert systems in general, and a general skepticism about the value of AI in making useful tools. Thus, as fewer expert systems found footing, it is not surprising that fewer explainable expert systems were produced.

However, during this time, other forms of AI and machine learning became popular, and researchers examined some of these domains with respect to explanation. This includes probabilistic reasoning approaches that dated back to Druzdzel and Henrion's research (Druzdzel, 1990; Druzdzel and Henrion, 1990; Henrion and Druzdzel, 1990ab), and later investigated in terms of Bayesian reasoning frameworks (Druzdzel, 1996; LaCave and Diez, 2002). In addition, several research groups have applied many of the explanation approaches of the earlier expert systems to agent-based systems (Johnson, 1994; Van Lent et al., 2004). There were also several publications devoted to case-based reasoning (Doyle et al., 2003; Sørmo et al., 2005; Kofod-Petersen et al., 2008), and explainable tutoring systems (Aleven and Kedinger, 2002; Lane et al., 2005). Although many of these research efforts investigated processes, techniques and explanation in systems other than expert systems, they often reached similar conclusions or used many of the properties of second-generation explanation systems.

During the period in which second generation intelligent systems were being developed, there was nothing like a winter in the area of cognitive systems engineering for software systems design. de Greef and Neerincx (1995) among others, advocated for an approach to software engineering that regarded the human-computer as a work system, designed around human knowledge and reasoning capacities. They asserted that this was to be achieved by designing interfaces that provide "cognitive support," that is, the interface is laid out in such a way as to convey knowledge that the user is lacking. In an experiment, they found that users with little expertise could perform better using a statistics program if the interface was designed on the basis of a model of expert reasoning (in the form of a state-transition diagram). This is similar in spirit to the use of Concept Maps as an explanatory interface (Ford, et al., 1996), discussed above.

Since the mid 2010s, a new generation—which we will call the Third Generation—has emerged.

## Third Generation Systems

In the last few years as machine learning and deep net technologies has been expanding in scope and application, the need for explanation has arisen reawakened. This has been driven by two factors: (1) users and external stakeholders are unclear how decisions and classifications are made (and so may not be able to defend their use), and (2) developers themselves often do not understand how their systems work. Large numbers of research papers have focused on visualizing, comparing, and understanding the operation of deep learning networks, with an audience of other researchers (e.g., Goyal et al., 2016abc; Karpathy et al., 2015; Kim, et al., 2016; Moeyersoms, et. al., 2016; Sadeghi, et al., 2015; Selvaraju, et al., 2017; Yosinski, et al., 2015; Zahavy et al., 2016; Zeiler and Fergus, 2014). However, a number of researchers have focused more specifically on explanations that might be presented to non-developer users of the systems (Hendricks et al., 2016; Kulesza et al, 2012; 2015; Lomas et al., 2012; Park et al., 2016; Ribiero et al., 2016; Rosenthal et al., 2016)

This third generation is typified by many of the same approaches as the first generation systems—attempts to make clear the inner workings of the system. As was true for the first generation efforts, this itself remains a substantial challenge. First generation systems embedded expert knowledge in rules often elicited directly from experts, and attempted to craft language descriptions from expert's assessments. These rules were often transformed into natural language expressions, and so much of the hard work was embedded within the knowledge representation. In third generation systems, this is substantially more difficult, and so we might expect challenges even achieving the types of explanations that were deemed inadequate in the early 1980s. Similarly, critiques of the first-generation system regarding improper levels of detail, nonsensical language, and the like may be challenges for third-generation systems.

In contrast to the first generation systems, computer technology in data visualization, animation, video, and the like have advanced substantially, so that many new ideas have been proposed as potential methods for generating explanations. Although first generation systems had natural dialogs and interactivity in question-answering systems, today's systems can support this in ways those systems could not. An example is the enablement of argumentative explanatory human-machine dialogs to deal with inconsistencies in knowledge bases in the process of software-assisted knowledge acquisition for expert systems (cf Arioua, Buche and Croitoru, 2017).

*Explainable Classifiers and Deep Learning Networks*

Researchers began to realize that the reasons for their systems' performance were difficult to identify. Consequently, a substantial number of researchers have tied to use systematic methods to explain system workings to one another or themselves. This might cover almost all current research on deep networks, but we have identified a set of more relevant sources that have used specific methods or approaches in service of explanation.

There are some techniques that can generally be applied to any classifier, or have been used on traditional classifiers such as support vector machines or decision trees (Akata, 2013; Akata, et al., 2015; Bilgic and Mooney, 2005; Biran and McKeown, 2014, 2017; Craven and Shavlik, 1996; Lakkaraju, Kamar, Caruana, and Leskovec, 2017; Lou, Caruana, and Gehrke, 2012; Martens, Huysmans, Setiono, Vanthienen, and Baesens, 2008; Martens and Provost, 2014; Montavon, Lapuschkin, Binder, Samek, and Müller-(2017). Možina, Demšar, Bratko, and Zupan, 2005; Robnik-Šikonja and Kononenko, 2008; Robnik-Šikonja, Likas, Constantinopoulos, Kononenko, and Štrumbelj, 2011).

There is also a large and still growing body of research developing explainable deep learning networks that rely on specific aspects of those systems. LeCun, Bengio and Hilton (2015) provide a review.

(Amos, Xu, and Kolter, 2017; Bach et al., 2015; Baehrens, et al., 2010; Clos, Wiratung, and Massie, 2017; Donahue, et al., 2015; Doshi-Velez and Kim, 2017; Elenberg, Dimakis, Feldman, and Karbasi, 2017; Erhan, Bengio, Courville, and Vincent, 2009; Féraud and Clérot, 2002; Fukui et al., 2016; Gan, Wang, Yang, Yeung, and Hauptmann, 2015; Goyal, et al., 2016a,b,c; Hendricks et al., 2016; Hinton, et al., 2015; Jain et al., 2016; Krizhevsky, et al., 2015; Kumar, Wong, and Taylor, 2017; Landecker, Thomure, Bettencourt, Mitchell, Kenyon, and Brumby, 2013; Lei, Barzilay, and Jaakkola, 2016; Li, Chen, Hovy, and Jurafsky, 2015; Li, Yosinski, Clune, Lipson, and Hopcroft, 2015; Liu ,et al., 2017; Mahendran and Vedaldi, 2015;

Moeyersoms, et al., 2016; Nguyen, Yosinski, and Clune, 2015; Papernot et al., 2017; Rajani and Mooney, 2017; Reed, Akata, Lee, and Schiele, 2016; Sadeghi, Kumar Divvala, and Farhadi, 2015; Samek et al., 2017; Schank, 2013; Schmidhuber, 2015; Shwartz-Ziv and Tishby, 2017; Simonyan, Vedaldi, and Zisserman, 2013; Springenberg, Dosovitskiy, Brox, and Riedmiller, 2014; Sturm, Lapuschkinb, Samek, and Müller, 2016; Sutskever, and Hinton, 2012; Tapaswi et al., 2016; Thrun, 1995; Usunier, Synnaeve, Lin, and Chintala, 2016; Xu and Saenko, 2016 Yosinski, et al., 2015; Zeiler and Fergus, 2014; Zeiler, Taylor and Fergus, 2011; Zhou et al., 2016).

## *Computational Models of Explanation*

Attempts to generate computational models of human explanation date to the 1980s (i.e., Goguen, Weiner and Linde, 1982; Ritter and Feurzeig, 1988; Friedman et al., 2018).

## *Other Modern AI that Explains or is Explainable*

Other modern approaches have sometimes gone beyond attempting to explain the system to other researchers, and have embedded logic (often as part of a question-answering system) to explain the system to users. The distinction between these systems and the approaches described above are sometimes fuzzy, but modern explainable systems tend to work toward interacting with a user rather than a developer or researcher. We also list here modern explainable systems that are difficult to classify.

(Assad, Carmichael, Kay, and Kummerfield, 2007; Brinton, 2017; Clos, Wiratunga, and Massie, 2017; Codella, et al., 2018; Craven and Shavlik, 1996; Fukui et al., 2016; Hu, Andreas, Rohrbach, Darrell, and Saenko, 2017; Jha, Raman, Pinto, Sahai, and Francis, 2017; Kim, 2018; Kim, Shah, and Doshi-Velez, 2015; Kim Lin Collins, Taylor, and Amer, 2016; Koh and Liang, 2017; Kononenko, et al., 2013; Kulesza, Burnett, Wong, and Stumpf, 2015; Park et al., 2016; Rajani and Mooney, 2017; Ribeiro, Singh, and Guestrin, 2016b, 2016a; Ross, Hughes, and Doshi-Velez, 2017; Selvaraju et al., 2016; Sheh, 2017; Symeonidis, Nanopoulos, and Manolopoulos, 2009; Tapaswi et al., 2016; Williams et al., 2016).

## Progressing Toward a Fourth Generation?

Some of the "explanations" produced in the first generation systems were easy to create in comparison to those in the third generation, as they were fairly direct restatements of hand-coded rules. The current generation may have a more difficult time producing the simple type of explanations produced by the first generation systems. But in some ways, third generation systems currently in development have mirrored developments in the first generation systems. We anticipate that many of these systems can anticipate similar problems faced by the first generation systems, and they might be addressed in ways that the second generation produced.

We might predict that many of the lessons learned in the first generation and implemented in the second generation will be learned again by constructing third generation systems (see Brock, 2018; Johnson and Lester, 2018). For example, Sheh and Monteath (2018) point out that explanations in the form of models of the how the AI works "have been largely forgotten in the XAI literature" in favor of explanations that present attributes used in a classification (e.g., saliency maps). To reiterate some of the lessons with a look toward the future, we might consider:

### _Global versus Local Explanation_

An important goal of explanation is to help a user develop a robust and predictive mental model of a system. Justifications of why an AI system made a particular determination for a particular case (local explanation) does not provide the same information as global explanations regarding how the system works generally. Global explanations may serve to help a user understand the individual justifications for particular decisions (P. Lipton, 1990; Z. Lipton, 2016; Ribiero, et al., 2016a,b; Wick and Thompson, 1992).

*Linguistic Realism*

Researchers have recently been demonstrating success at developing language abilities via recurrent neural networks (RNNs) to create chatbots and question-answering systems. One success is the explanation system described by Hendricks et al. (2016), but for many third generation explanation systems, explanation takes the form of expressions in natural language. First generation systems had problems when the language mapped onto rules that were foreign to humans; by analogy, we might expect that the mapping of language to the true decision processes will continue to challenge today's systems.

*User Models and Goals*

One striking parallel between fist-generation and second generation systems is that they generally ignored the goals, knowledge, abilities, and skills of the user. This means that they may not even understand or take into account previous information and explanations, and may not understand what an individual knows so to try to provide the best explanation at that point. Furthermore, the goals of users, systems, and explanations remained ill-specified. It may be that most users want explanations to help them understand the system in the future, but most explanations are providing justification for why they are doing something currently. Other goals that have been seen in the past include learning about unfamiliar information, gaining trust in the system in order to support its adoption, using it for debugging and validation checking, etc. Goals and purposes should be clarified so that, even if an explanation only serves one of these goals, it is tested in the appropriate context.

*Contrastive Explanation*

Interestingly, research papers aiming to explain deep nets to other scientists have often taken contrastive approaches. These include: (1) occlusion (e.g. Zeiler and Fergus, 2014) which show how classifications differ as regions are removed from an image, (2) ablation (e.g., Sadeghi et al., 2015) in which layers or functions are removed from the system to compare, and (3) alternate training methods (e.g., Goyal et al., 2016) in which the same system is trained on different data sets to understand how it is using information; providing counter-examples (e.g., Shafto and Goodman, 2008; Shafto, Goodman, and Griffiths, 2014; Nguyen et al., 2015). These types of reasoning schemes may help third generation systems go beyond simple descriptions of the internal workings, toward crafting genuine explanations that are informative and that resonate with users of the systems.

## _Explanation versus Justification_

As has been pointed out above, explanations and justifications each have particular uses or contributions, but it is important not to confuse or equate the two. Especially important for XAI is the possibility that an "explanation" of a system architecture that justifies it for a computer scientist might be mistaken as a genuine explanation, that is, one that helps a user develop a good mental model. In the XAI context a considerable number of formalisms have been mentioned as being explanations. This includes diagrams of algorithms, diagrams showing chains of reasoning modules, decision trees, feature lists or matrices, parameters interpreted as conditional probabilities, bar charts, queries mapped onto ontology graphs, proof paths, and-or graphs, and others. At the same time, discussions of XAI have recognized that explanations also take the form of examples, category exemplars, distinguishing features, counter-cases to knock down misconceptions. It is recognized that the human must be able to explore the operation of the XAI. It is recognized that explanation in XAI may involve a human-machine dialog in which the XAI and the user are co-adaptive. It will be important for XAI to be mindful of the blurry line between describing things in the world (classifications of objects or events) versus explaining how the AI determined its classifications.

*Evaluation: Measurement of Explainable AI*

Evaluating the performance of an explanation in AI is multi-faceted. It can involve assessing anything from competency, to subjective trust, to adoption, to logical analysis of the information provided as an explanation. Many researchers have used specific methods for assessing the knowledge of the system, and those methods might be generalized for assessment of future systems. In addition, several researchers have identified specific taxonomic measurement frameworks or recommendations for assessment.

(Doshi-Velez and Kim, 2017; J. Doyle, Radzicki, and Trees, 2008; Freitas, 2014; Groce et al., 2014; Kulesza et al., 2011; Kulesza, Stumpf, Burnett, and Kwan, 2012; T. Miller, 2017; T. Miller et al., 2017; Pacer, Williams, Chen, Lombrozo, and Griffiths, 2013; Sauro and Dumas, 2009; Shinsel et al., 2011; Teach and Shortliffe, 1981)

Performance evaluation for AI systems is discussed in more detail on Section 8 of this Report.

**Summary**

We have reviewed the approaches taken by a number of kinds of "explanation systems," across three generations since the 1970s—spanning expert systems, intelligent tutoring systems. and a recent renaissance that could be called the third generation. We have described some of the approaches currently being explored in this third generation, and pointed out some commonalities with the limitations that emerged in the first generation.

## 6. Psychological Theories, Hypotheses, and Models

Strictly speaking, there are no comprehensive theories of explanation in psychology, in the sense of well-formed theories that make strong predictions. The literature is best described as presenting a number of hypotheses. Many if not most of these have been adopted and adapted from other disciplines, especially philosophy and computer science. What makes psychology stand out is that many of the hypotheses have been demonstrated or verified in experimentation. The various hypotheses are typically not at odds with one another, in the sense of making different predictions. Rather than presenting the psychological work as a number of alternative theories, or even conceptual process models, we instead present the literature for what it is: A roster of hypotheses, or empirical assertions about relationships between manipulations and outcomes.

With regard to this empirical base, it should be noted that most of the investigations are either: (1) studies conducted in the academic laboratory using college students as "subjects" and simplified reasoning problems, or (2) studies conducted using school children as participants to explore questions about cognitive development. These features entail a need for caution—researchers often generalize their results to "people," they often generalize from simple laboratory tasks to the understanding of complex (computational) systems, and then often generalize "years of experience" as a metric for expertise.

The hypotheses proposed in the psychology literature can be placed into these categories:

- Taxonomics of types of explanation,
- The relation of explanation to other fundamental reasoning processes,
- The relation of explanation to learning,
- The value, utility or uses of explanations,
- The features of good explanations,
- The limitations and foibles of explanatory reasoning,
- Individual differences in explanatory reasoning.

After listing the hypotheses found in the literature, we lay out several conceptual models of explanation that have appeared in the literature.

**Taxonomics**

It can be inferred from the literature that explanations can refer to essentially anything that people might experience: causal mechanisms, descriptions of the purposes of things, the purposes or goals of actions or activities, the causes of phenomena, the purposes of goals of human behavior, the features of categories, etc. (Lombrozo, 2009). Across disciplines, there are dozens of taxonomies that organize explanation according to type, purpose, property, and other organizing principles. The following categories of theories are each linked to potential taxonomic organizations of explanations.

**Relation of Explanation to Fundamental Cognitive Processes**

A number of hypothesis have been proposed relating to how explanation is influenced by or influences fundamental cognitive processes. These include:

Hypothesis: The search for explanation is an active, motivated process that guides strategies in the search for information (Krull and Anderson, 1997; Pirolli and Card, 1999).

Hypothesis: The search for explanations is a core part of the sensemaking process (Pirolli and Card, 1999; Klein, Moon and Hoffman, 2006).

Hypothesis: Explanation is the search for answers to why, how and what-if questions, and as such is a form of inference making (Graesser, Baggett and Williams, 1996).

Hypothesis: Explanation is the process of articulating the causes of phenomena or events. In other words, explanation and causal reasoning are co-implicative. Instances of explanation often are expressions of the causes of things (Ahn, et al., 1995; Gopnik, 2000; Keil, 2006; Lombrozo, 2010).

Hypothesis: Explanation can be equated with abductive inference, that is, the empirical determination of a best possible explanation (Harman, 1965; Lombrozo, 2012, 2016: Wilkenfeld and Lombrozo, 2015).

Hypothesis: Explanation of "why something is what it is" entails an explanation of "why it is not something else." In other words, explanation and counterfactual reasoning are co-implicative. Instances of explanation often are expressions of counterfactual reasoning (Byrne, 2017).

Hypothesis: Explanation of "why something is what it is" entails a prediction of "what might happen next." In other words, explanation and prospection are co-implicative. Instances of explanation are often expressions of what will happen in the future (Koehler, 1991; Lombrozo and Carey, 2006; Mitchell, Russo and Pennington, 1989).

Hypothesis: Explanation has a heuristic function, of informing the process of discovery (Lombrozo, 2011).

Hypothesis: Explanation can be a collaborative process; therefore, a theory of explanation must model the contexts of communication, common ground, mentoring, and instruction (see Clark and Brennan, 1993).

## The Relation of Explanation to Learning

Explanation is clearly linked to teaching, learning and tutoring.

Hypothesis: Explanations "recruit" prior knowledge to enable the formation of mental models (Besnard, Greathead and Baxter, 2004; Keil, 2006; Williams and Lombrozo, 2013).

Hypothesis: The process of explanation is the construction of causal chains, goal-plan-action hierarchies, and justifications (Graesser, Baggett and Williams, 1996).

Hypothesis: Explanation is the process of forming a concept or category. Explanations help learners form or refine conceptual categories, or the features of category membership, and generalize properties within or across categories (Lombrozo, 2009; Lombrozo and Gwinne, 2014; Murphy and Medin, 1985; Williams, Lombrozo and Rehder, 2010).

Hypothesis: People have strong intuitions about what constitutes a good explanation (Lombrozo, 2016).

Hypothesis: Self-explanation has a significant and positive impact on understanding. Deductions and generalizations help learners refine their knowledge (Chi and VanLehn, 1991: Chi, et al, 1989, 1994; Lombrozo, 2016; Rittle-Johnson, 2006). Having learners explain the answers of "experts" also enhances the learner's understanding (Calin-Jagerman and Ratner, 2005). The inadequacies of understanding are revealed when learners (users) are asked to provide their own explanation (Fernbach, et al., 2012).

**The Value, Utility or Uses of Explanations**

A number of hypotheses relate to the value and purposes of explanations in general.

Hypothesis: Explanations serve to justify, reinforce beliefs and enhance the perceived value of beliefs (Brem and Rips. 2000; Krull and Anderson, 1997; Preston and Epley, 2005; Santos, et al., 2009).

Hypothesis: Explanations enable people to diagnose situations, predict or anticipate the future, justify decisions or actions (Keil, 2006; Lombrozo, 2011).

Hypothesis: Explanation is the process of discovery and confirmation of hypotheses or mental models (Keil, 2006; Lombrozo, 2011).

Hypothesis: Explanation as a process is entailed by the processes of argumentation and negotiation (Andriessen, 2002; Galinsky, et al., 2008; Rips, Brem and Nailenson, 1999).

Hypothesis: Explanations enable learners to revise their beliefs, but only if the explanation enables them to construct an alternative mental model (Einhorn and Hogarth, 1986; Hilton and Erb, 1996).

Hypothesis: Explanations enable people to notice and attempt to resolve apparent inconsistencies and restore consistency between their mental model and empirical reality (Khemlani and Johnson-Laird, 2010).

Hypothesis: Confidence in one's understanding is enhanced if it is possible to imagine and explain a future possibility (Koehler, 1991; Krull and Anderson, 1997). The assumption of truth influences anticipation of the future.

**The Features of Good Explanations**

Many researchers have proposed answers to the question of what makes for a good explanation.

Hypothesis: Explanations are regarded as good if they are relevant to the learner's context or goals, in addition to expressing truths about causal relations (Hilton and Erb, 1996; Keil, 2006).

Hypothesis: Explanations that refer to causation are good if they express the cause-effect covariation and temporal contiguity (Einhorn and Hogarth, 1986).

<u>Hypothesis</u>: Explanations tend to be discounted if they refer to multiple alternative possible causes that reduce the plausibility, relevance or informativeness of the preferred (Einhorn and Hogarth, 1986; Hilton and Erb, 1996; McGill, 1991; Sloman, 1994).

<u>Hypothesis</u>: Explanations are good if they make sense in terms of the context or the enabling conditions of cause-effect relations (Einhorn and Hogarth, 1986; Hilton and Erb, 1996).

## The Limitations and Foibles of Explanatory Reasoning

Some research has examined the biases, errors, and limitations that can occur during explanation. Many of these regard the types of explanations individuals prefer or are influenced by, rather than aspects of explanations that are objectively better or worse.

<u>Hypothesis</u>: People prefer explanations that refer to single, necessary, or "focal" causes and that are both simple and yet broadly applicable (Einhorn and Hogarth, 1986; Hilton and Erb, 1996; Lombrozo, 2007, 2016; Klein et al., 2014).

<u>Hypothesis</u>: People overestimate the likelihood of events that make simple explanations more probable or plausible (Lombrozo, 2007; Lombrozo and Rutstein, 2004).

<u>Hypothesis</u>: People prefer causal explanations that refer to multiple causes when those causes are understood as a simple chain (Ahn, 1998; Klein, 2006).

<u>Hypothesis</u>: In order to adopt a complex explanation, people require disproportionately more evidence than is required for the acceptance of simple explanations (Lombrozo, 2007; Lombrozo and Rutstein, 2004).

<u>Hypothesis</u>: People will believe explanations to be good even when they contain flaws or gaps in reasoning (Lombrozo, et al., 2008).

Hypothesis: Explanations can impair learning if they do not encourage the learner to discover patterns (Williams, Lombrozo and Rehder, 2010).

Hypothesis: The way an invitation to explain is phrased (e.g. Why is x the case?) can bias peoples' assumptions about causes. Concepts or categories that are referenced in the question can influence how people ascribe causes (Giffin, Wilkenfeld, and Lombrozo, 2017).

Hypothesis: Explanation is a double-edged sword. Explanations can promote over-generalization. The understanding of broad patterns can impede the appreciation of exceptions. It is easier for people to learn patterns than to learn about exceptions.

Hypothesis: The explanations people generate for everyday events and human affairs are often biased, in that people's lack of knowledge and susceptibility to misinformation leads them to create, value, and prefer biased or incorrect explanations (Tworke and Cimpian, 2016; van der Linden, et al., 2017).

Hypothesis: People often to not consider the strength of evidence in support of a causal explanation, and do not distinguish evidence that supports truth claims from the plausibility of a proposed explanation (Kuhn, 2001).

**Individual Differences in Explanatory Reasoning**

Some research has asked whether there are systematic differences—akin to personality traits or thinking styles—that predict satisfaction with explanations across people.

Hypothesis: People differ in their satisfaction threshold.
- Some people are satisfied with simple or superficial explanations that reference fewer causes (Lombrozo and Rutstein, 2004). A consequence is the "illusion of explanatory depth," the overconfident belief that one's understanding is sufficient (Keil, 2006; Rozenblit and Keil, 2002).

- Some people are not satisfied with simple, superficial explanations and are more deliberative and reflective in their explanatory reasoning (Fernbach, et al., 2012).

- People tend to prefer complex over simple explanations if they can see and compare both forms (Klein et al., 2014)

- There are clear cultural differences in preference for simple versus complex explanations (Klein et al., 2014).

**Conceptual Models of Explanation**

We present the four psychological conceptual models of explanation that have been presented as conceptual diagrams. These are noteworthy because these kinds of conceptual models are historically been regarded in AI as first-pass architectures.

The Johnson and Johnson Model (1993). These researchers studied the collaborative explanation process, in which experts explained to novices the processes of statistical data analysis. Transcripts of explainer-learner dialogs were analyzed. A key finding was that the explainer would present additional declarative or procedural knowledge at those points in the task tree where sub-goals had been achieved. The two Johnson and Johnson models are depicted in Figure 7.1.

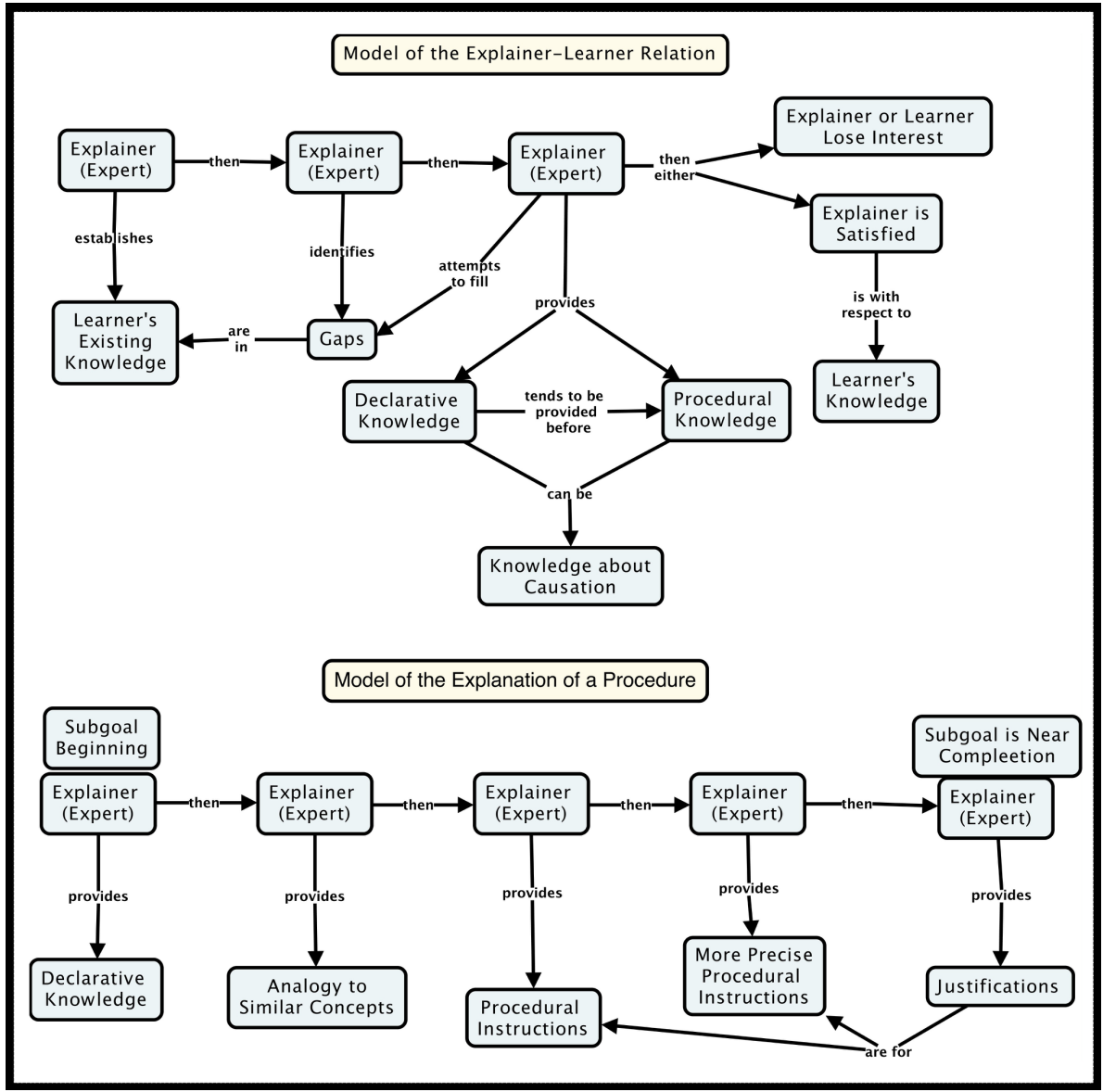

**Figure 7.1. The Johnson and Johnson Models of the collaborative explanation process.**

The Krull and Anderson Model. Anderson, Krull and Weiner (1996) and Krull and Anderson (1997) presented a model of self-explanation. This model seems patterned after models in the control theory paradigm, that is, the branchings are all "yes or no" decisions. The first step in the model, the noticing of an event, is reminiscent of the first step in C.S. Peirce's model of abduction (1891, 1903), that is, the observation of something that is interesting. That said, the model sidesteps the crucial aspects of explanation. Specifically, Problem Formulation and Problem Resolution is where the hard work of explanation occurs, and the model is not specific about what is involved in these steps. The Krull model is presented in Figure 7.2. This Figure includes some elements that are not in their published diagram but are referenced in the body of the paper. These are highlighted in color.

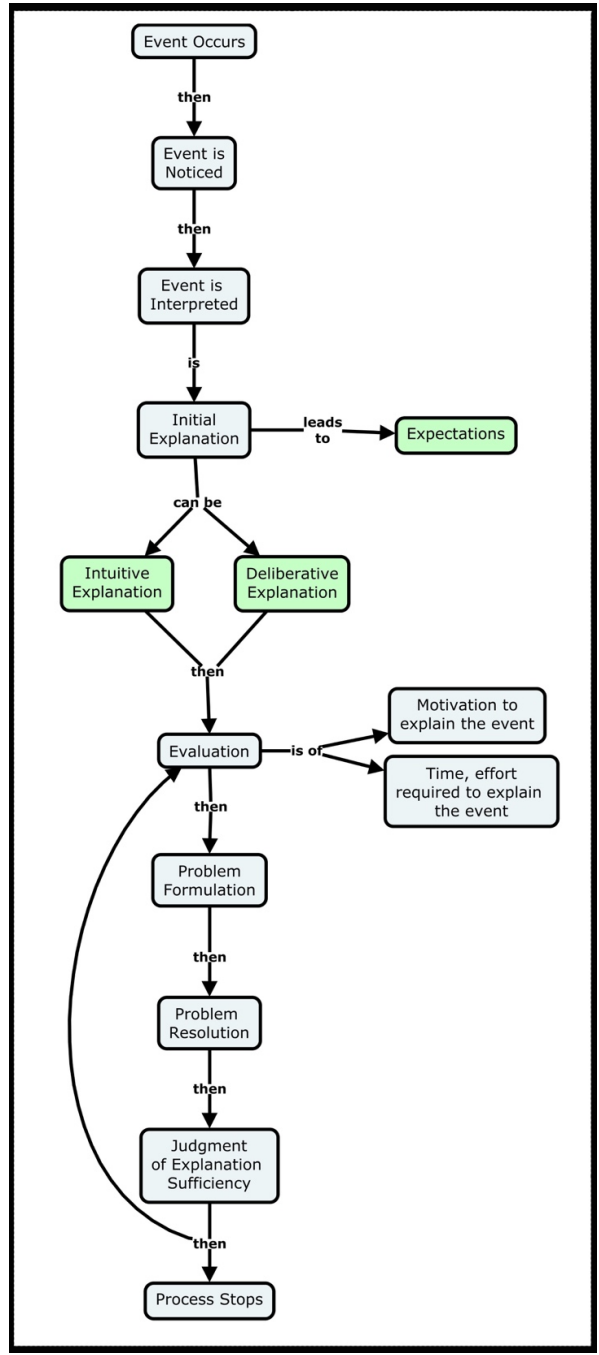

**Figure 7.2. The Krull, et al. model.**

The Klein et al. Sensemaking model. The Data/Frame model describes what happens as people try to understand complex situations, and continually work to refine and improve upon that understanding (Klein, Moon and Hoffman, 2006; Klein et al, 2007). The model is presented in Figure 7.3.

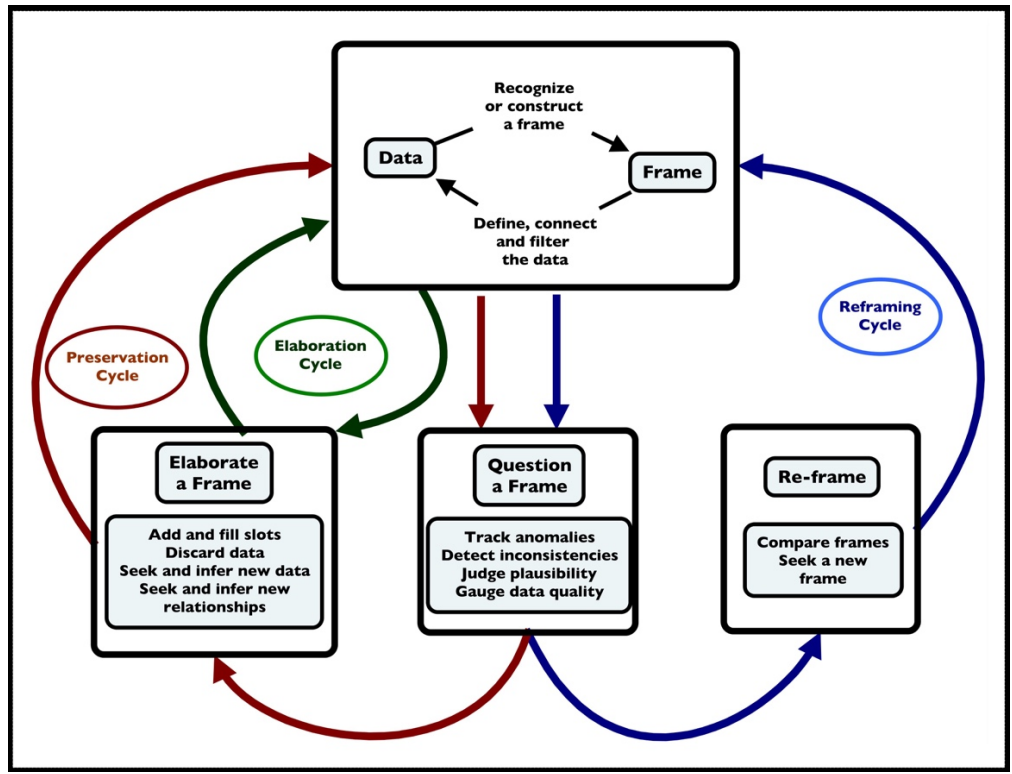

**Figure 7.3. The Data/Frame model of sensemaking.**

The Pirolli and Card Information Foraging Model. This model emerged from cognitive task analysis to reveal the reasoning of intelligence analysts (see Pirolli and Card, 2005). Like the Data-Frame model presented above, the Pirolli-Card sensemaking model can be interpreted as a model of the explanation process. Pirolli and Card generalized their model as the "information foraging" model (Pirolli and Card, 1999). A version of this model is presented in Figure 7.4. This version abstracts away from the published model, as the published model was referenced to one specific task (i.e., all-source intelligence analysis)

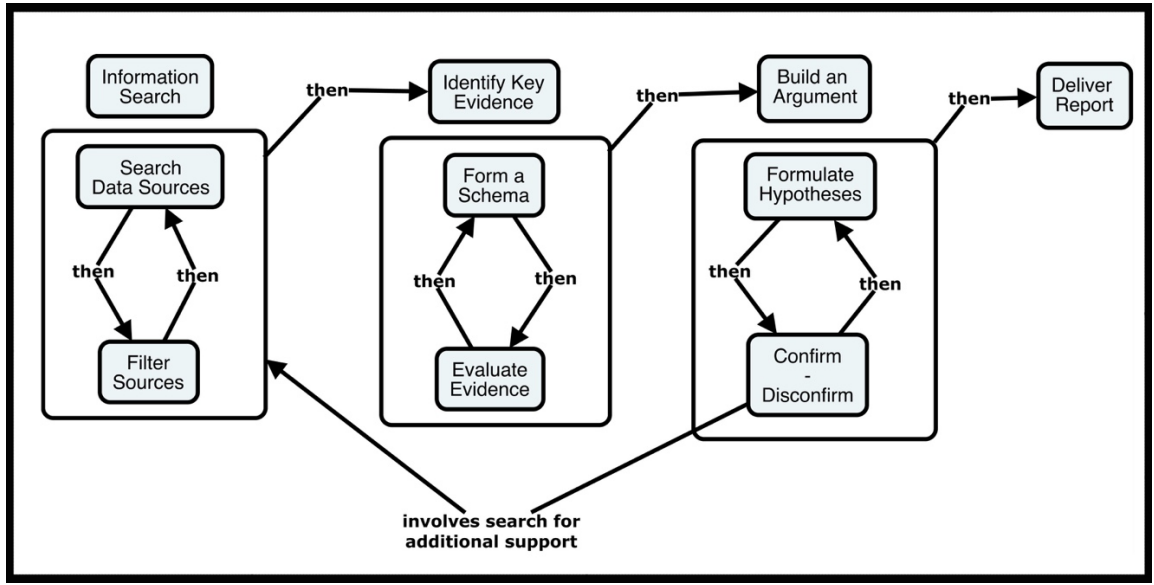

**Figure 7.4. The Pirolli-Card Sensemaking model.**

## Conclusions About the Psychological Models

- With the exception of the Data/Frame Model, all of the conceptual models assume clear-cut beginning and ending points to the explanation process.

- With the exception of the Johnson and Johnson model, which is explicitly dyadic, all of the models seem to reference self-explanation although they can be interpreted as being agnostic on this matter. That is, they might be understood as referring to team cognition or human-machine cognition.

- The models refer, variously, to "knowledge structures," "schemas," "frames" and other primary mental representations, but these are all referencing the same basic idea.

# 7. Synopsis of Key XAI Concepts

The following concepts represent the "common ground" for research and development on XAI. These concepts are ones that appear across the literatures of AI, psychology, philosophy and the other pertinent disciplines. They are concurrence points for researchers and scholars. In this section, we offer a glossary in which we provide definition and context regarding many of the terms and concepts that appear in the literature.

## The Value of Explanation

A considerable number of studies have shown that in the use of decision aids and recommender systems of various types, the provision of explanations significantly contributes to user acceptance of and reliance upon computer-provide guidance (e.g., Herlocker, et al., 2000). Users like to be provided with explanations, even if they are justifications of system operations expressed in somewhat formal terms (Biran and Cotton, 2017). On the other hand, Continued experience with an AI system, by itself, enables people to discount their initial misconceptions: "Participants were able to give lay descriptions of attributing simple machine learning concepts to the system despite their lack of technical knowledge" (Tullio, Chalecki, and Fogarty, 2007, p. 1).

## Format and Reference

Various researchers and scholars have proposed taxonomies of kinds of explanations, and principles composed from them. Many different kinds of representations are regarded as presentations of explanations. Table 6.1 lists the kinds of representations that are regarded as explanations.

Table 6.1. Format and Reference in explanations.

| Format (How it is expressed) | Reference (What it is about) |
|---|---|
| <ul><li>Visualizations (e.g., Heat maps)</li><li>Text (Statements, Narratives or Stories, Answers to queries, Human-machine dialogs)</li><li>Formal Expressions (Logical expressions, Matrices)</li><li>Conceptual Process Models (Diagrams)</li><li>Graphs, Networks</li><li>Tables</li><li>Abstractions, generalizations</li><li>Timelines</li><li>Hierarchies (Trees)</li></ul> | <ul><li>Examples (includes misclassifications, counter-examples, outliers, clear cases, close competitors)</li><li>Patterns, Classes, Ontologies</li><li>Features, Weights, Probabilities, Ranks, Parameters</li><li>Decisions, Strategies, Goals</li><li>Algorithms, Computational Processes, Proofs</li><li>Incidents, Events (includes self-explanations or stories)</li><li>Cause-effect relations</li></ul> |

It is noted that an explanation can involve mixed types (e.g., a story with diagrams; an incident account including a table, etc.).

**Interpretability and Explainability**

The notion that AI systems need an explainable model was first noted in the work on intelligent tutoring systems and expert systems (see Teach and Shortliffe, 1981; Clancey, 1983; Moore and Swartout, 1988; Sørmo, et al., 2005.) The difficulty that was then encountered, however, was that the explanations that could be generated were typically straightforward verbalizations of rules, and not justifications of the system's routines or architecture (see Moore and Swartout, 1988). A formal expression of "Why we did it this way" is a justification, not an explanation.

More recently, evidence has been adduced that explanations in terms of formal aspects of systems operations (e.g., formal graphs, etc.) are less satisfying than simple, clear explanations (e.g., that a categorization was made because of some particularly strong feature). Nevertheless, users want to see explanations of some sort.

In computer science, a system is interpretable if the input-output relationship (its decision or choice) can be formally determined to be optimal or correct, in either a logical or a statistical sense (see Ribiero, et al., 2016a,b). Measures of the accuracy of a formal model are not the same as measures of the goodness or explainability of the model from the user's perspective. On this meaning, an interpretation does not necessarily carry any explanatory value with reference to understanding on the part of humans other than those who are familiar with formal computational systems. That is, justification can be provided for systems that are not human-interpretable. For example, a linear model may be simple in a formal sense, but if it involves hundreds of factors and weights, it would be minimally explainable to a human. It is in this sense that a formal interpretation is a justification, but a justification that works for a computer scientist does not necessarily work as a justification for others.

Explanatory value must be considered with respect to the individual who is regarded as the beneficiary of the explanation.

**Explanation versus Justification**

Many of the early expert systems provided explanations that were simple verbalization of rules. This was found to be inadequate, and justifications (an argument about why a particular decision was made) began to be developed, and researchers argued that justifications led to better system adoption. However, the distinction is not always clear, because the original 'explanations' offered should not have been called explanations, but rather 'knowledge verbalization,' and both are types of candidate explanations.

In more recent approaches, the distinction between these types of explanations (and others) has remained. However, over-reliance on justification may also be problematic, as it focuses on local actions rather than global operations, and can focus on persuasion rather than knowledge-level explanation.

## Candidate Explanations

Material that is offered as an explanation, no matter its medium, format, or reference, is only an explanation if it results in good effect, that is, it has explanatory value for particular individuals. Technically, the property of "being an explanation" is not a property of text, statements, narratives, diagrams, or other forms of material. It is an interaction of: (1) the offered explanation, (2) the learner's knowledge and beliefs, (3) the context or situation and its immediate demands, and (4) the learner's goals or purposes in that context. This explains why it is possible that purely descriptive statements, not primarily intended to serve as explanations, can nevertheless have explanatory value.

## Mental Models

Mental models are a convenient term for a constellation of well-developed knowledge about a system. A good mental model will help a user interpret, predict, and mentally simulate the operation of a system, as well as to understand the system's limits and boundary conditions.

A considerable amount of research has been conducted in which user's mental models are elicited and evaluated for content and use (see Kass and Finn, 1988; Moray, 1987; Tullio, et al., 2007). As users experience work using the AI system and develop their mental model, they imagine themselves doing the work as if they were the system, or were inside the system. Instructions, explanations and user experiences shape the user's mental model, which in turn determines subsequent performance, subsequent needs for further or deeper explanation, and shapes both trust and reliance. Information about a user's mental models of computational systems is typically determined by the use of structured interviews, or a combination of interview-like probe-questioning about what participants are thinking as they perform some other task (e.g., web-based querying, or "eavesdropping" on user-system interactions (see for example Kass and Finn, 1988; Kim, Khanna and Koyejo, 2016; Muramatsu and Pratt, 2001).

**Anticipation and Prospection**

Just as good scientific explanations are ones that enable the scientist to make predictions, a successful explanation of an AI system should enable the user to correctly predict what the AI system will do for given cases. This can include cases that the AI gets right and also cases it gets wrong (e.g., failures, anomalies).

**Global and Local Explanations**

During initial instruction and practice, users are inclined to need instructions that describe "how the system works." On the other hand, during actual use, users are inclined to need explanations of why the system did what it did on particular cases. Local explanation is defined as meaningful information about the calculational or logical processes involved in the processing of a particular case. The role of global and local explanations in the overall explanation process are described in P. Lipton (1990), Z. Lipton (2016), Ribiero, et al., 2016a,b, and Wick and Thompson (1992).

**Context Dependence**

The goodness of an explanation depends on the use context and goals of the person who is the beneficiary of the explanation. Brezillon and Pomerol (1997, p. 3) noted that "System capabilities to explain its reasoning to a user is a well-studied issue" but then added, "However, few systems are able to provide explanations that are relevant to the user." The various reasons for why a person seeks an explanation (referred to as "triggers").

**Contrastive Reasoning**

A number of researchers emphasize that good explanations are contrastive: They explain the "Why," the "Why not" and the "What-if" of systems. Answering a contrastive question (explaining the difference between two cases) is not only effective in improving understanding (i.e., mental model formation), but is simpler than providing a full causal analysis (Kim, Khanna and Koyejo, 2016; Lipton, 2016; Miller, 2017).

## Correspondence

Since the 1980s, it has been recognized in AI that the explanations offered by systems (expert systems, explanation systems, intelligent assistant systems, etc.) should correspond to the explanations that appear in human-human explanatory dialogs. Thus, criteria for what makes a computer-generated explanation "good" are the criteria for what makes an explanation "good" for humans.

## Explanation Goodness and Explanation Satisfaction

An explanation may be satisfying to a user even though it does not satisfy some of the goodness criteria. A user may prefer an explanation that is clear but incomplete (a "minimum necessary" criterion) (see Miller, 2017; Miller, et al., 2017). Generally speaking, sufficient explanations are likely to fall in the sweet spot between detail and comprehensibility. Specifically, they would be explanations that invoke just a handful of causal factors and a handful of process steps. A user may prefer an explanation that is simple and perhaps even misleading but serves to help the user avoid error (an "inoculation" criterion).

Table 6.2 lists the features that are believed to characterize good and satisfying explanations. A detailed treatment of the psychometrics for XAI measurement of explanation goodness and satisfaction appears as Hoffman, Miller, Klein and Litman (2018).

**Table 6.2. Explanation "goodness" features.**

| Soundness | Plausibility, Internal consistency |
|---|---|
| Appropriate Detail | Amount of detail and focus points of the detail |
| Veridicality | Does not in any way contradict the ideal model (although there are times when inaccurate explanations work better for some users and some purposes) |
| Usefulness | Fidelity to the designer's or user's goal for system use |
| Clarity | Understandability |
| Completeness | Relative to an ideal model |
| Observability | Explains the system mechanism or process |
| Dimensions of Variation | Reveals boundary conditions |

Vasilyeva and Lombrozo (2015) demonstrated that explanations in terms of mechanisms are regarded as better than explanations that just state a covariation. Their example involves statements to the effect that: (1) people who visit a portrait gallery in a museum are more likely to make a donation to the museum versus (2) exposure to faces is known to stimulate pro-social feelings. In other words, explanations in terms of covariation does not always convince people that there is an underlying mechanism.

What counts as an explanation also depends on what knowledge the Learner/User already has, and also on the Learner/User/s goals. The dimension of usefulness can be analyzed in terms of the user (or learner's) goals. This leads to a consideration of function and context of the system (software, algorithm, tool), that is, why does a given User/Learner need an explanation? In the various pertinent literatures, this is expressed in terms of the different kinds of questions that a User/Learner might have, or different kinds of explanations provided by software agents (Haynes, et al., 2009). These "triggers" for explanation are listed in Table 6.3.

**Table 6.3. The explainee's "triggers" and goals.**

| TRIGGERS | USER/LEARNER'S GOAL | |
| --- | --- | --- |
| | NEED TO UNDERSTAND | NEED TO ACCOMPLISH |
| How do I use it? | Use | Primary task goals |
| How does it work? | Mechanism, Understandability | Feeling of Satisfaction |
| What did it just do? | Mechanism, Understandability | Feeling of Sufficiency |
| What does it achieve? | Function | Usefulness |
| What will it do next? | Observability | Trust, Reliance |
| How much effort will this take? | Usability | Primary Task Goals |
| What do I do if it gets it wrong? | Anticipation | Control |
| How do I avoid the failure modes? | Surprise | Adaptation, Work-around |
| What would it have done if x was different? | Mechanism, Understandability | Curiosity, Trust, Reliance |
| Why didn't it do z? | Anticipation, Surprise | Curiosity, Trust, Reliance |

There are trade-offs between the goals and other goodness criteria (Table 6.1). For example, the "completeness" criterion would entail the conclusion that a good explanation would have to answer all of the trigger questions. Alternatively, a User may seek an explanation that addresses only one or two of the goals. The "appropriate detail" criterion entails a concept of Minimum Necessary Information as a potential criterion to resolve this particular trade-off.

**Corrective Explanation**

User's mental models may involve reductive understandings or potentially harmful simplifications. When faced with information that challenges their understanding, users may attempt to preserve their misunderstandings or simplifications. Users need to be inoculated against explanations that are both reductive (that is, not complete) and incorrect. Explanations must enrich mental models but also correct user misunderstandings.

**Trust and Reliance**

Substantial research has focused on trust in automation in general, and the evaluation measures (mostly questionnaires and scales) can be adapted to determine whether explanations and justifications improve trust and reliance. It is clear that some of the issues that relate to human trust in automation translate to the area of human trust in AI systems and explainable AI systems. Trust is an important consideration in the discussion of automation and AI, since affective judgments and user trust often have an impact on the motivations and intentions to use the systems (Merritt, 2011).

Human factors research has shown that trust in and reliance upon automation are causally related, and not just in some simple way. While trust can lead to reliance, it can also lead to complacency, that is, the unquestioned reliance on the automation (see for example, Lyons, et al., 2017). Users can have trust in a model independently of whether they trust individual predictions of the model. In complex cognitive work systems, people always develop some admixture of justified and unjustified trust and justified and unjustified mistrust in automation. They will trust the system to do certain things under certain circumstances and mistrust the system with regard to certain functions in certain circumstances. Their admixture of trust states has a significant and direct impact on reliance. Classic papers include Moray and Inigaki (1999) and the review article by Lee and See (2004). Hoffman (2017) presents a general taxonomy of human-machine trusting relations.

**Self-Explanation**

Self-explanation is the "effort after meaning" motivated by an intrinsic need to understand. Psychological research has shown that self-explanation leads to improved learning about the operation of complex systems. Furthermore, deliberately prompting individuals to explicitly self-explain likewise facilitates learning (Chi, et al., 1994). This finding obtains even when one takes into account individual differences in the motivation or need to understand, i.e., curiosity (Gopnik, 2000; Litman and Sylvia, 2006).

## Active Exploration as a Continuous Process

While the explanation process involves the assimilation of knowledge, it also transforms the learner's knowledge and is thus an accommodative process. Thus, rather than referring to "explanations" (and assuming that the property of being an explanation is a property of statements), it might be prudent to refer to explaining, and regard explaining an active exploration process. Humans are motivated to "understand the goals, intent, contextual awareness, task limitations, [and] analytical underpinnings of the system in an attempt to verify its trustworthiness" (Lyons, et al., 2017).

One of the consensus points coming from the philosophy of science is that explanations have a heuristic function: They guide further inquiry. The delivery of an explanation is not always an end point to a process. Indeed, it must be thought of as a continuous process since the XAI system that provides explanations must enable the user to develop appropriate trust and reliance in the AI system. The user must be able to actively explore the states and behavior of the AI, especially when the system is operating close to its boundary conditions, or is making errors (see Amerishi, et al., 2015). How can XAI work in concert with the AI to empower learning-while-using?

> *Knowing why the aid might err increased trust in the decision aid and increased automation reliance, even when the trust was unwarranted… Ideally, training should include instruction on how the aid operates and experience using the aid. Both can playa role in producing an appropriate level of trust in the aid (Dzindolet et al., 2003, pp. 697,715).*

Exploratory debugging has been shown to result in better mental models and improved user performance, especially when the debugging operations can be reverses, thus allowing the user to actively explore how the AI system works (see Amrishi, et al., 2015; Kuelsza, et al., 2010, 2011, 2012; Stumpf, et al., 2007). Another approach to active exploration is "adjustable replay." In this approach, the user can instruct the AI to revert to a previous state, and then re-run a particular case while one or more features of the case are modified (removing features, changing weights, etc.). In yet another approach, the user is able to "clone" the AI system, that is, revert it to its pre-training state, and then see how it treats selected cases after being trained on a different training set, or having selected features or property ranges. Using such approaches, the user could explore the AI system's preferences, tendencies, etc. and see how, when and why the AI fails or makes an error.

Active exploration on the part of the user (or learner) may not always be enough, for two reasons. First, it is yet to be demonstrated that active exploration of specific cases (local explanation) is not always sufficient to instill a global understanding (See P. Lipton, 1990; Z. Lipton, 2016; Ribiero, et al., 2016a,b; Wick and Thompson, 1992). Second, the process aspect of explanation becomes especially salient, and especially important to XAI, when one considers the Moving Target Problem. XAI algorithms and deep net mechanisms continue to learn/adapt once they are operational and are benefitting from learning on the basis of more data. This means that explaining becomes a moving target since the competence envelope of the AI will always be changing. In other words, XAI explanations must not only explain how the AI works, but explain how the AI is able to improve and adapt. It is for this reason that exploratory/explanatory debugging has perhaps the greatest potential for XAI. (For more details, see Section 8 of this Report.)

Is it for these reasons that a consideration of explanation as a collaborative process becomes even more crucial.

## Explaining as a Collaborative or Co-adaptive Process

While explanation might be thought of as a process with clear-cut beginning and end points (the delivery of instructional material that the user simply assimilates), the understanding of a computer system requires a collaboration or co-adaptive process involving the learner/user and the system. "Explanations improve cooperation, cooperation permits the production of relevant explanations" (Brezillon and Pomerol, 1997, p. 7). " One of the observations that has emerged from the study of natural explanation is that the process of human discourse is essentially a cooperative one" (Southwick, 1991, p. 12). This is the concept of "participatory explanation," similar to the notion of "recipient design" in the conversation analysis literature, i.e., that messages must be composed so as to be sensitive to what the recipient of the message is understanding (Sacks and Schegloff, 1974). An assumption in some of the first generation of AI-explanation and intelligent tutoring systems was that it is the human who has to learn, or change, as a result of explanations offered by the machine.

Based on empirical findings, it is generally recognized that the process of explaining, and human-human communication broadly, is a co-adaptive "tuning" process, which requires that the explainer and learner have a capacity to take each other's perspective. In turn, this entails that the explainer and the learner have mental models of the mental models of each other. Although sometimes simple statements may provide a good explanation to a particular user, Swartout (1981, p. 315) argued that "A powerful knowledge representation is the secret to better explanation, not just better natural language facilities."

## Measurement and Evaluation

Speaking to the computer science/AI community, Doshi-Velez and Kim (2017) stated that: "It is essential that we as a community respect the time and effort to conduct evaluations." The context of this statement is that the computer science community has developed many benchmarks that can be relatively easily used to determine whether an algorithm is better than another, but since explanations are intended for human, they need behavioral science concepts to evaluate properly. A number of methods are available for evaluating XAI systems, and not all XAI systems have to be evaluated in research with human participants (Miller, et al., 2017). That said, evaluation in terms of human performance in actual system use is an arbiter of XAI success. Research using human participants is resource- and time-intensive.

There are significant methodological or measurement challenges:

- Explanations generated by XAI can be evaluated in terms of the goodness criteria and the results correlated to user performance.
- User comprehension of explanations can be assessed and the results correlated to qualities of their mental models and to their performance.
- User knowledge or mental models can be measured or somehow represented, and the results correlated to the change in performance attributable to the explaining process.

A variety of performance measures are conceivable, and many have been utilized in research. These present challenges for forming operational definitions, and challenges to experimental design.

- Successful explanations enable the user to efficiently and effectively use the AI in their cognitive work, for the purposes that the AI is intended to serve.
- Successful explanations enable the user to choose the best among a set of alternative formal models (e.g., algorithms).
- Successful explanations enable the user to correctly predict what the AI system will do for given cases. This can include cases that the AI gets right and also cases it gets wrong (e.g., failures, anomalies).
- Successful explanations enable the user to explain the cases that the AI gets wrong.
- Successful explanations enable the users to correctly assess whether a system determination is correct, and thereby recalibrate their trust.
- Successful explanations enable the user to explain how the AI works to other people.

- Successful explanations enable the user to judge when and how to rely on the AI even while knowing the boundary conditions of the competence of the AI.

A review of measures for XAI is presented in the DARPA XAI Report, "Metrics for Explainable AI: Challenges and Prospects" (Hoffman, et al., 2018).

# 8. Evaluation of XAI Systems:
## Performance Evaluation Using Human Participants

**Key Empirical Questions**

It was not long after the development of second generation expert systems that the notion of evaluation became salient. The first generation work has led to the conclusion that expert systems (or knowledge based systems, generally), needed to explain their reasoning. It followed that there must be means for evaluating the goodness and utility of explanations. Dhaliwal and Benbasat (1996) proposed an evaluation framework based on cognitive theories, for assessing the extent to which explanations facilitate learning and decision making by relying on a model of explanation strategies. They asserted that evaluation must address these high-level questions: (1) To what extent are the explanations generated by expert systems relied upon in decision making, (2) What factors influence the use of the explanations? and (3) In what ways do explanations empower users? They also listed a number of specific scalar measures, such as satisfaction, confidence, usefulness, trust, and a number of performance measures, such as decision efficiency and accuracy.

 Many additional researchers discussed the challenges of evaluating human-computer systems, including explainable AI systems (e.g., Abrams and Treu, 1977; Antunes, Herskovic, Ochoa, and Pino, 2012; Kulesza, et al., 2010, 2012, 2011, 2013, 2015; Nataksu, 2004; Ribiero, et al., 2016a,b; Sheh and Monteath, 2008; Treu, 1998; Wickens, 1994), and the challenges for the evaluation of explanation systems (Doshi-Velez and Kim, 2017ab; Gilpin et al., 2018; Feldmann, 2018; Miller, 2017; Miller, Howe, and Sonenberg, 2017). Much of this literature references the evaluation of the software systems or interfaces, rather than the performance of the human-machine work system. Treu (1998) reviewed measures that had been used to evaluate human-machine performance (e.g., response time, error rate, ease of use, etc.), but with respect to software systems, generally rather than explanation systems, specifically.

However, a number of communities have examined different aspects of human-machine interaction that may be relevant to evaluating explainable systems. For example, research has shown that supporting explanatory human-machine dialog can enhance the process of knowledge acquisition for expert systems (e.g., Arioua, Buche and Croitoru, 2017). As another example, research has shown that in order to obtain the best results from human-AI interaction, reliance, trust, and optimal utilization must be "calibrated" for individual operators via an interaction between performance expectations (expected system competence; Muir, 1994), user self-confidence (Muir and Moray, 1996), user training (Cahour and Forzy, 2009), and prior operator experience with AI (Dzindolet, et al., 2003). The recommendations provided hinge on the ability of AI to be probed/examined by the human user and to explain faults as they occur. Thus, the ongoing development of AI systems capable of explanation will be crucial in the development of trust between human users and AI, and appropriate reliance.

(Adams et al., 2003; Bisantz, Llinas, Seong, Finger, and Jian, 2000; Cahour and Forzy, 2009; Doshi-Velez and Kim, 2017; Dzindolet, Peterson, Pomranky, Pierce, and Beck, 2003; Dzindolet, Pierce, Pomranky, Peterson, and Beck, 2001; Hancock et al., 2011; Hoffman, Klein, and Miller, 2011; Hoffman, 2017a; Jian, Bisantz, and Drury, 1998, 2000; Johnson, 2007; Koustanai, Mas, Cavallo, and Delhomme, 2010; Lee and Moray, 1994; Lee and See, 2004; Llinas, Bisantz, Drury, Seong, and Jian, 1998; Lyons et al., 2017; Merritt, 2011; Merritt, Heimbaugh, LaChapell, and Lee, 2013; Montague, 2010; Moray and Inigaki, 1999; Muir, 1994; Muir and Moray, 1996; Parasuraman, Molloy, and Singh, 1993; Parasuraman and Riley, 1997; Pomranky, Dzindolet, and Peterson, 2001; Schaefer, 2013; Singh, Molloy, and Parasuraman, 1993a; Wang, Jamieson, and Hollands, 2009; Wickens, 1994).

## XAI Measurement Concepts

Figure 8.1 presents the conceptual model used in the DARPA XAI program to show the place and role of different classes of measures in the evaluation of the XAI system and the Human-XAI system performance.

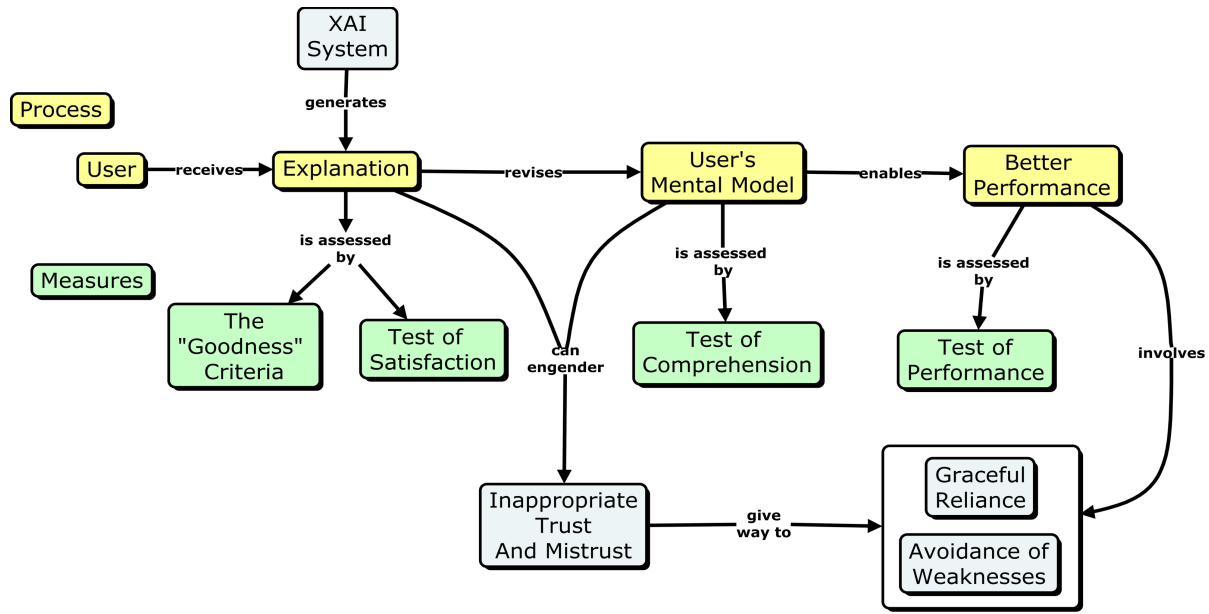

**Figure 8.1. DARPA's XAI Evaluation Framework. Explanations should induce better mental models and performance, which produce appropriate trust in the system.**

_Explanation Goodness_ refers to evaluations about properties of the explanation that are often taken as axiomatic assumptions about what constitute good explanations. For example, researchers have argued that explanations should be complete, logical, incremental, veridical, and the like. This dimension covers properties that might typically be evaluated objectively or via coding rubrics, and may not require the explanation to be placed in the context of the working system or an expert user.

_Mental Model_ refers to a user's knowledge about the system. This could be assessed in many ways (see Adams et al., 2003), but may include specific tests of knowledge, prediction about the system, generative exercises (drawing a diagram or explaining how the system works), and the like.

Performance refers to measures of whether the system (possibly including the human user) operate better or more efficiently with the explanation than without it. Performance constitutes both measures of the AI capability itself (so that if an explainable system were not as capable as its non-explainable counterpart, this could be detected), but also the effectiveness of the joint system (so that if explanations help the user work with the system or accomplish goals more effectively, this could also be measured). This might involve tasks in which the user must decide which subtasks the AI is capable or licensed to perform, triaging a large set of tasks between different AI or human performers, or overall time or financial costs associated with fielding a system. In general, performance could be multidimensional, so that an XAI-enabled drone may perform more slowly than its AI-enabled counterpart (because of the need to interact with a user), but may be more likely to succeed.

*Trust/Reliance* measures generally refer to the subjective attitudes of users toward the system. Although trust and reliance can be assessed indirectly via usage and performance patterns (using the system signals reliance on the system), in general trust measures involve human reflection about their reliance, trust in, or the trustworthiness of a system. As discussed by Adams et al. (2003), this need not be exclusively Likert-scale assessments, and for example forced-choice decisions could be used (e.g., which of two systems would you prefer to use?), or other similar assessments.

A detailed presentation on methods for measuring explanation goodness, satisfaction, mental models, trust, reliance, and performance appears in another DARPA XAI Report (Hoffman, Miller, Klein and Litman, 2018). Consonant with Hoffman, et al., Kim (2018) proposed that good explanations must not only be correct and accurate, but must enable the user to answer questions about the AI, especially counterfactual questions.

## Annotated Review of Papers Relevant to the Evaluation of XAI Systems

A number of computer science reports and review articles examine explainability in AI, and propose what are believed to be good or useful explanation types and formats. However, frequently these explanations, or the human-machine work system within which whey are embedded , are not subject to any evaluation (see Section 5, above). A number of psychological experiments involved assessing the effect of explanations on human performance using software systems, but did not involve either AI systems or computer-generated explanations (e.g., Berry & Broadbent, 1987). We have identified research reports that have attempted empirical evaluations of explanations or system performance, in the context of AI systems. To be included in the present analysis, we required (1) the system to have or simulate some non-trivial intelligent algorithm, where some aspect of the algorithm that was claimed to be either an explanation or explainable, and (2) the system had to have some sort of human-based evaluation of the explanation.

For each reference, we considered:

- Kinds of AI systems; what the AI system was supposed to do.
- Who the research participants were (e.g.,. college students, Mechanical Turk workers, etc.) and how many.
- The nature of the interface that was presented explanations to users.
- The form of the explanations that were presented (i.e., diagrams, text, whatever).
- The nature of the experiment design (any independent variables or different experimental conditions).
- Whether "explanation goodness" was evaluated, and if so, how.
- Whether user models were evaluated for their soundness (as a result of having received explanations), and if so, how.
- How the human or the human-machine performance was measured.
- Whether user trust in or reliance on the AI was evaluated and if so, how.

The last four considerations correspond to the measurement classes described in the DARPA XAI Evaluation Framework (Figure 8.1, above).

Not all of the papers that were identified as being of interest to XAI met all of the above criteria, or even involved evaluating the AI-Human system performance, but did involve some sort of evaluation. For example, Mao, et al. (1996) evaluated a knowledge-based system that provided explanations of its decisions. Their explanations could be either textual or graphical, and included domain-relevant definitions, causal descriptions, and problem-solving methods. However, because they provided no evaluation of the system, it was not included. Their purpose was to evaluate and describe a knowledge-based system that provided explanations, not to evaluate the explanations in terms of their contribution to the human-machine performance. Consequently, this Section focuses on the methods by which explainable systems have been evaluated by humans.

*Kinds of AI Systems*

Studies have used a variety of AI systems, including rule-based expert systems, case-based reasoning systems, machine learning systems, Bayesian classifiers, quantitative models, statistical models, and decision trees. The most frequent references are to expert systems and machine learning systems.

There is also a considerable variety of applications, such as gesture classification, image classification, text classification, decision-making, program debugging, music recommending, financial accounting, strategy games, command training, robotic agents, non-player character agents, and patient diagnosis. Not all XAI systems that have been evaluated have had mature use cases and domains. For many, the systems were research-based systems that explored how explanations could be added to the technology, rather than commercial systems in which explanations were added to aid users in real settings.

Because of the focus of these papers on the explanation, many offer very little details about the nature of the AI systems themselves, saying only that the system is an "intelligent agent," for example. Oftentimes, the baseline AI systems have been well documented in earlier papers, or they are common enough approaches that their basic working can be inferred.

## Nature and Number of Research Participants

Many studies relied on Mechanical Turk workers. Many relied on college students as participants. A few studies distinguished novices from experts, and at least one study appears to be against expert and novice user "profiles," rather than actual users. Sample sizes ranged from small (10 or fewer) to very large (over 600). Most studies used less than 100 participants. Sample sizes in studies using college students tended to be in the smaller range of less than 100, typically 20 or so.

## Explanation Forms and Formats

Explanation in AI systems has often been considered a process or general approach toward interacting with the user, and so it is sometimes difficult to point to the artifact produced by a system that should be considered an explanation. In the papers we reviewed, explanations took diverse forms, including text, statistical graphs, decision trees, feature histograms, color gradients, feature matrices, rule sets. inference networks or flags of "key events" imposed in a mission timeline. Textual explanations could be sets of words or text pieces denoting the features that were most important to the classifiers. Only in a few cases were explanations in the form of natural language.

In a few studies, the initial instructions provided global information about "how the AI works." (Many studies do not report details about the instructions given to the participants.) *In most cases, the explanations that were provided to participants were local justifications about why an action was taken, and the task required self-explanation on the part of the participants*. Such explanations do not explain how the AI system works directly. Rather, they explain (in some way, textual or graphical) the AI system's rationale for each particular decision or classification (e.g., the most heavily weighted features). Presumably, after sufficient interaction with these justifications, the user develops a clearer mental model of how the system works in general.

_Kinds of Study Designs or Experimental Comparisons_

Studies employed a variety of designs, including single-group, repeat measure designs and factorial designs with more than one Independent Variable. A number of studies involved a comparison of Explanation versus No Explanation conditions. Many studies presented more than one kind of explanation, attempting to evaluate the effects of learning or performance.

Kulesza et al. (2012) is perhaps one of the cleanest example of an "Explanation versus No Explanation" comparative evaluation of explanation effectiveness, in terms of both performance and mental models The user task was to use a music recommender system. Users could provide feedback about particular songs (which they described as "debugging"). Feedback could be with respect to such factors as song tempo, song popularity, artist similarity. The interactivity capability is suggestive of the notion of explanation as a human-machine dialog (see also Kulesza, et al., 2015). Participants who received richer (functional-local and structural-global) explanations during the initial instructional phase did not perform better at debugging the recommender's reasoning. Instead, the most effective participants may have learned to debug by using the system. This is suggestive of the notion of "explanation as exploration" (see also Kulesza, et al., 2015):

> _Despite the complexity inherent to intelligent agents, With-scaffolding participants quickly built sound mental models of how a music recommender operates "behind the scenes"—something the Without scaffolding participants failed to accomplish over five days. The participants' mental model transformations—from unsound to sound—was predictive of their ultimate satisfaction with the intelligent agent's output. Participants with the largest transformations were able to efficiently adjust their recommenders' reasoning, aligning it with their own reasoning better (and faster) than other participants. These same participants were also likely to perceive a greater benefit from their debugging efforts. Participants presented with structural knowledge of the agent's reasoning were significantly more likely to increase their computer self-efficacy, which is known to correlate with reduced computer anxiety and increased persistence when tackling complex computer tasks. Participants who were presented with structural knowledge showed no evidence of feeling overwhelmed_ *by this additional information and viewed interacting with the intelligent agent in*
*a positive light, while participants holding only functional mental models more*
*frequently described their debugging experience* (Kulesza et al., 2012, p. 9).

See the Appendix for further details.

## Ways in which Explanation Goodness was Manipulated

We refer to explanation "goodness" as the set of principles by which guide the development of
explanations, and through which they can be reasonably evaluated without relying on human
participants, and typically center on concepts such as correctness, completeness, incrementalism,
reversibility, and the like. Explanation goodness was not evaluated *a priori* in most studies, or
manipulated as an independent variable. (That is, no study involved a control in which
explanations were manifestly deficient.) Though the researchers assumed that (good)
explanations had to have certain qualities, such as correctness or completeness. A number of
studies focused primarily on having users or domain experts complete a questionnaire about the
features they desired in AI or decision aid systems.

## Ways in Which Mental Models were Evaluated

Few studies looked at mental models directly. Those that did utilized self-reports or post-
experiment interviews. Some studies looked at mental models only indirectly (e.g., by applying
the NASA-TLX scale for rating mental workload); this despite the clear consensus that user
mental models (that is, their understanding of the AI systems) are crucial. For example, Teach
and Shortliffe (1982) conducted a survey study in which clinicians responded to a questionnaire
on attitudes regarding clinical consultation systems, regarding the systems' acceptability, and
physicians' desirements regarding the performance capabilities of the systems. The study
provided indirect evidence about mental models and explanation. Some used specific knowledge
tests as a proxy for eliciting a more complex mental models (e.g., Kulesza et al., 2015).

The study by Tullio et al. (2007) is the richest in terms of systematic study of user's mental models and the development of mental models across experience.

## *Kinds of Measures of Task or System Performance*

Perhaps the clearest measure of whether an explanation is effective is if the performance of the human-XAI work system improves when explanations are involved. This could be a result of a better understanding of the AI system (and thus it is used properly), or of more calibrated trust (so that cases in which the system will fail are handled or checked by the user, while easy decisions are defaulted to the AI). Not all test cases for explainable systems can be easily evaluated with performance measures. Especially when the goal is to improve transparency, explanation might be meant to teach a user how a decision is made, or to provide a check to ensure the AI system is not working unfairly. An explanation of a credit score might lead a user to make changes to their financial habits, or a regulator or legislator take action that changes the decision process, but these are difficult to assess in an experiment with human participants..

All of the primary performance measures we examined are task-specific measures of correctness or efficiency. Example correctness measures are the number of loan decisions correctly judged to be non-compliant, correctness of a diagnosis of some sort, the correctness of a classifier, or the number of successful attempts to correct a deficient program. Such measures can be expressed as percent correct, but this leaves the result insensitive to context. What counts as "good performance" in one domain may be quite different from that for some other domain. What is needed in such cases is some sort of baseline where one can anchor the percentage scale by assigning ranges of values to one or more categories (e.g., 85% correct is "Satisfactory").

Response time (or time-to-learn) is also used as a primary measure. It can be expressed as a conditional, such as time to respond for correct versus incorrect judgments. Response time also cannot be understood out of context, since task durations differ widely across task domains (e.g., response time to accepting or rejecting a personalized song recommendation versus response time to find a bug in a program). Furthermore, systems with explanations will sometimes require the user to take more time to interact with and process,, and so a performance measure would have to balance time-cost and cost for erroneous outcomes in order to properly judge cost versus benefit. Performance can also be measured via trials-to-criterion, which may represent a training cost of a system. Response time can also depend heavily on user knowledge and experience. For some explainable systems, a knowledgeable user may no longer need detailed explanations, but explanations may reduce the time needed to become knowledgeable.

The study by Fails and Olsen's (2003; see also Kim et al., 2015) had as its primary measures the amount of time and the number of training iterations it took for the human-machine system to achieve 97.5% agreement with a gold standard non-interactive classifier.

Some studies used "debugging tasks" or "oddball detection" tasks, in which users' corrections were used to modify learned programs (e.g., Kulesza et al., 2010). For example, participants might have to decide which of a set of words does not belong to a category, decide which topic a document pertains to. Some studies also used some sort of "prediction task" in which participants had to predict the AI systems' determination for given test cases. These are different ways of framing performance, and are useful especially when they can be argued or demonstrated to be related to deficiencies of real-world AI systems.

### _Kinds of Measures of Trust/Reliance_

A number of studies had participants rate their satisfaction either with the explanations (soundness or completeness) or with their own performance. Many studies pertain to the trust-reliance relationship indirectly. On the other hand,, the evaluation of trust was actually the primary performance measure in a few studies.

Studies used a variety of human judgments (typically, Likert scales), such as confidence in a classification or decision, degree of trust in the AI, degree to which they relied on the AI's determinations, or their perceived efficacy at using the AI system to conduct the experiment tasks.

It should be recognized that, although XAI trust and reliance measures focus on subjective ratings, there are alternative ways in which trust can be measured (see Adams et al., 2003). These include pairwise choice methods, performance measures, and the like.

**Conclusions About Evaluation**

It is recommended that AI/XAI researchers be encouraged to include in their research reports fuller details on their empirical or experimental methods, in the fashion of experimental psychology research reports: details on Participants, Instructions, Procedures, Tasks, and Dependent Variables (operational definitions of the measures and metrics) and Independent Variables (conditions) and Control Conditions.

In the set of papers reviewed in this Section (and detailed in the Appendix) one can find considerable guidance for procedures in the evaluation of XAI systems. But there are noteworthy considerations, such as:

- Consideration of global-versus local explanations,
- Consideration of the need to evaluate the performance of the human-machine work system (and not just the performance of the AI or the performance of the users).
- Consideration that the experiment procedures tacitly impose on the user the burden of self-explanation.

Tasks that involve human-AI interactivity and co-adaptation, such as bug or oddity detection, seem to hold the promise for XAI evaluation since they conform to the notions of explanation as exploration and explanation as a co-adaptive dialog process. Tasks that involve predicting the AI's determinations, combined with post-experimental interviews, hold promise for the study of mental models in the XAI context.

# Bibliography of Research Relevant to XAI

Abdul, A., Vermeulen, J., Wang, D., Lim, B.Y., & Kankanhall (2018). Trends and trajectories for explainable, accountable and intelligible systems: An HCI research agenda. In *Proceedings of CHI 2018* [DOI:10.1145/3173574.3174156] New York: Association for Computing Machinery.

Abrams, M.D., & Treu, S. (1977). A methodology for interactive computer service measurement. *Communications of the Association for Computing Machinery, 20*, 936-944.

Adams, B. D., Bruyn, L. E., Houde, S., Angelopoulos, P., Iwasa-Madge, K., & McCann, C. (2003). Trust in automated systems. *Ministry of National Defence*

Abrams, M.D., & Treu, S. (1977). A methodology for interactive computer service measurement. *Communications of the Association for Computing Machinery, 20*, 936-944.

Adler, P., Falk, C., Friedler, S. A., Nix, T., Rybeck, G., Scheidegger, C., … Venkatasubramanian, S. (2018). Auditing black-box models for indirect influence. *Knowledge and Information Systems, 54*(1), 95–122.

Ahn, W. (1998). Why are different features central for natural kinds and artifacts?: The role of causal status in determining feature centrality. *Cognition, 69*(2), 135–178.

Ahn, W., Kalish, C.W. Medin, D., & Gelman, S. (1995). The role of covariation versus mechanism information in causal attribution. *Cognition, 54*(3), 299–352. https://doi.org/10.1016/0010-0277(94)00640-7

Ahn, W., & Kalish, C.W. (2000). The role of mechanism beliefs in causal reasoning. In R. Wilson & F. Keil (Eds.), *Explanation and Cognition* (pp. 199–225). Cambridge: MIT Press.

Ahn, W., Kalish, C.W., Medin, D.L., & Gelman, S.A. (1995). the role of covariation versus mechanism information in causal attribution. *Cognition, 54*, 299-352.

Akata, Z. (2013). Label embedding for image classification. *Proceedings of the The IEEE Conference on Computer Vision and Pattern Recognition* (CVPR), pp. 819-826. New York: IEEE. [arXiv:1503.08677v2].

Akata, Z., Perronnin, F., Harchaoui, Z, & Schmid, C. (2015). Label embedding for image classification. [arXiv:1503.08677v2 [cs.CV]]

Alang, N. (2017, August 31). Turns out algorithms are racist. *The New Republic*. Retrieved from https://newrepublic.com/article/144644/turns-algorithms-racist?utm_content=buffer7f3ea

Aleven, V.A., & Koedinger, K.R. (2002). An effective metacognitive strategy: Learning by doing and explaining with a computer-based Cognitive Tutor. *Cognitive Science*, *26*(2), 147–179.

Alonso, J.M., Castiello, C., & Mencar, C. (2018). A bibliometric analysis of the explainable artificial intelligence research field. In J. Medina, et al. (eds.), *International Conference on Information Processing and Management of Uncertainty in Knowledge-Based Systems (IPMU 2018): Information Processing and Management of Uncertainty in Knowledge-Based Systems Theory and Foundations,* pp. 3-15. New York: Springer.

Amershi, S., Chickering, M., Drucker, S.M., Lee, B., Simard, P., & Suh, J. (2015). Modeltracker: Redesigning performance analysis tools for machine learning. In *Proceedings of the 33rd Annual ACM Conference on Human Factors in Computing Systems* (pp. 337–346). ACM. Retrieved from http://dl.acm.org/citation.cfm?id=2702509

Amodei, D., Olah, C., Steinhardt, J., Christiano, P., Schulman, J., & Mané, D. (2016). Concrete problems in AI safety. *[ArXiv:1606.06565].*

Amos, B., Xu, L., & Kolter, J. Z. (2017). Input convex neural networks. In *Proceedings of the 34th International Conference on Machine Learning*. Sydney, Australia.

Anderson, C.A., Krull, D.S., & Weiner, B. (1996). Explanations: Processes and consequences. In E.T. Higgin & A.W. Kruglansky (Eds.), *Social psychology: Handbook of basic principles* (pp. 271–296). Guilford New York. Retrieved from http://www.cfs.purdue.edu/richardfeinberg/csr%20331%20consumer%20behavior%20%20spring%202011/grad/attitude%20review%20chapters%20%20stuff/mental%20explanations.pdf

Anderson, J.R., Boyle, F., Corbett, A.T., & Lewis, M.W. (1990). Cognitive modeling and intelligent tutoring. *Artificial Intelligence*, *42*(1), 7–49. https://doi.org/10.1016/0004-3702(90)90093-F

Anderson, J.R., Corbett, A. ., Koedinger, K.R., & Pelletier, R. (1996). *Cognitive tutors: Lessons learned* (No. B74F). Carnegie Mellon University.

Andriessen, J. (2002). Arguing to learn. In *K. Sawyer (Ed.) The Cambridge handbook of the learning sciences* (pp. 443–459). Cambridge: Cambridge University Press.

Antaki, C. (1989). Lay explanations of behaviour: How people represent and communicate their ordinary theories. In C. Ellis (Ed.), *Expert knowledge and explanation: The knowledge-language interface* (pp. 201–212). New York: John Wiley.

Antaki, C., & Leudar, I. (1992). Explaining conversation: Towards an argument model. *European Journal of Social Psychology, 22*, 181-194.

Antifakos, S. Kern, N., Schiele, B., & Schwaninger, A., (2005). Towards improving trust in context-aware systems by displaying system confidence. In *Proceedings of Mobile HCI 2015: The 7th International Conference On Human Computer Interaction With Mobile Devices and Services* (pp. 9-14). New York: Association for Computing Machinery.

Antunes, P., Herskovic, V., Ochoa, S. F., & Pino, J. A. (2012). Structuring dimensions for collaborative systems evaluation. *ACM Computing Surveys* (CSUR), 44 (2), Article No. 8.

Arioua, A,. & Croitoru, M. (2015). Formalizing explanatory dialogues. In *Proceedings of the International Conference on Scalable Uncertainty Management* (pp. 282-297). New York: Springer.

Arioua, A., Buche, P, & Croitoru, M. (2017). Explanatory dialogs with argumentative faculties over inconsistent knowledge bases. Submitted to the *Journal of Expert Systems with Applications*.

Arora, K.S. (1987). Expert systems. In *Encyclopedia of Artificial Intelligence* (Vol. 1, pp. 287–298).

Assad, M., Carmichael, D.J., Kay, J., & Kummerfield, B. (2007). PersonisAD: Distributed, active, scrutable model framework for context-aware services. In *Lecture Notes in Computer Science,* Vol. 4480. [DOI: 10.1007/978-3-540-72037-9_4]. New York, Springer.

Axelrod, R. (2015). *Structure of decision: The cognitive maps of political elites*. Princeton, NJ: Princeton University Press.

Bach, S., Binder, A., Montavon, G., Klauschen, F., Müller, K.-R., & Samek, W. (2015). On pixel-wise explanations for non-linear classifier decisions by layer-wise relevance propagation. *PloS One*, *10*(7), e0130140.

Baehrens, D., Schroeter, T., Harmeling, S., Kawanabe, M., Hansen, K., & Müller, K.-R. (2010). How to explain individual classification decisions. *Journal of Machine Learning Research, 11,* 1803-1831.

Baldwin, T.T., & Ford, J.K. (1988). Transfer of training: A review and directions for future research. *Personnel Psychology, 41*, 63-105.

Balfe, N., Sharples, S., & Wilson, J. R. (2018). Understanding is key: an analysis of factors pertaining to trust in a real-world automation system. *Human factors*, *60*(4), 477-495.

Ballas, J.A. (2007). Human-centered computing for tactical weather forecasting: An example of the "Moving Target Rule." In *Expertise out of context: Proceedings of the Sixth International Conference on Naturalistic Decision Making* (pp. 317–326).

Bansal, G. (2018). Explanatory dialogs: towards actionable, interactive explanations. In *Proceedings of the AAAI/ACM Conference on AI, Ethics, and Society*. New York: Association for Computing Machinery.

Bansal, A., Farhadi, A. & Parikh, D. (2014). Towards transparent systems: Semantic characterization of failure modes. In *Proceedings of the European Conference on Computer Vision ECCV 2014: Computer Vision* (pp. 366-381). New York: Springer.

Barzilay, R., McCullough, D., Rambow, O., DeCristofaro, J., Korelsky, T., & Lavoie, B. (1998). A new approach to expert system explanations. In *Proceedings of the Ninth International Workshop on Natural Language Generation* (pp. 78-87). Stroudsburg, PA: Association for Computational Linguistics.

Bechtel, W., & Abrahamsen, A. (2005). Explanation: A mechanist alternative. *Studies in the History and Philosophy of Biology and Biomedical science, 36,* 421-441.

Bellman, K. (2001). Building the right stuff: Some reflections on the CAETI program and the challenge of educational technology. In *Smart machines in education: The coming revolution in educational technology* (pp. 377–420). MIT Press.

Bellotti, V., Back, M., Edwards, W.K., Grinter, R.E., Henderson, A., & Lopes, C. (2002). Making sense of sensing systems: five questions for designers and researchers. In *Proceedings of the SIGCHI conference on Human factors in computing systems* (pp. 415–422). ACM.

Bellotti, V., & Edwards, W.K. (2001). Intelligibility and accountability: Human considerations in context-aware systems. *Human–Computer Interaction, 16*(2–4), 193–212.

Bengio, Y, Louradour, J., Collobert, R., & Weston, J. (2009). Curriculum Learning. *Proceedings of the 26th Annual International Conference on Machine Learning,* 41–48.

Belotti, V., & Edwards, K. (2001). Intelligibility and accountability: Human considerations in context-aware systems. *Human-Computer Interaction, 16,* 193-212.

Berry, D.C., & Broadbent, D.E. (1987). Explanation and verbalization in a computer-assisted search task. *Quarterly Journal of Experimental Psychology, 39A,* 585-609.

Besnard, D., Greathead, D., & Baxter, G. (2004). When mental models go wrong: co-occurrences in dynamic, critical systems. *International Journal of Human-Computer Studies, 60*(1), 117–128.

Bidot, J., Biundo, S., Heinroth, T., Minker, W., Nothdurft, F., & Schattenberg, B. (2010). Verbal plan explanations for hybrid planning. In M. Schumann, et al. (Eds.), *Proceedings of Multikonferenz Wirtschaftsinformatik* (MKWI 2010) (pp. 2309-2320). Göttingen, Germany Universitätsverlag Göttingen.

Bilgic, M., & Mooney, R. J. (2005). Explaining recommendations: Satisfaction vs. promotion. In *Beyond Personalization Workshop, IUI* (Vol. 5, p. 153). Retrieved from http://www.cs.utexas.edu/~ml/papers/submit.pdf

Biran, O., & Cotton, C. (2017). Explanation and Justification in Machine Learning: A Survey. *IJCAI-17 Workshop on Explainable Artificial Intelligence (XAI)*. Retrieved from http://home.earthlink.net/~dwaha/research/meetings/ijcai17xai/1.%20(Biran%20&%20Cotton%20XAI-17)%20Explanation%20and%20Justification%20in%20ML%20-%20A%20Survey.pdf

Biran, O., & McKeown, K. (2014). Justification narratives for individual classifications. In *Proceedings of the AutoML workshop at ICML* (Vol. 2014). Retrieved from http://www.cs.columbia.edu/~orb/papers/justification_automl_2014.pdf

Bisantz, A., Llinas, J., Seong, Y., Finger, R., & Jian, J.-Y. (2000). *Empirical Investigations of Trust-Related Systems Vulnerabilities in Aided, Adversarial Decision Making*. DTIC Document.

Bisantz, A M., Finger, R., Seong, Y., & Llinas, J. (1999). Human performance and data fusion based decision aids. In *Proceedings of the FUSION* (Vol. 99, pp. 918–925). Citeseer.

Bjork, E.L., & Bjork, R.A. (2014). Making things hard on yourself, but in a good way: creating desirable difficulties to enhance learning. In *Psychology and the real world: Essays illustrating fundamental contributions to society* (2nd ed., pp. 59–68). New York (NY, USA): Worth.

Bjork, R.A., Dunlosky, J., & Kornell, N. (2013). Self-regulated learning: Beliefs, techniques, and illusions. *Annual Review of Psychology*, *64*, 417–444.

Blokpoel, M., Van Kesteren, M., Stolk, A., Haselager, P., Toni, I., & van Rooii, I. (2012). Recipient design in human communication: Simple heuristics or perspective taking? *Frontiers in Human Neuroscience*, *6*(253), 1–22. https://doi.org/10.3389/fnhum.2012.00253

Bogomolov, S., Magazzeni, D., Podelski, A., & Wehrle, M. (2014). Planning as model checking in hybrid domains,. In *Proceedings of the twenty-eighth AAAI conference on artificial intelligence* (pp. 2228-2234). Palo Alto, CA: AAAI.

Bogomolov, S., Magazzeni, D., Minopoli, S., & Wehrle, M. (2015). PDDL+Planning with hybrid automata: Foundations of translating must behavior. In *Proceedings of the twenty-fifth AAAI conference on artificial intelligence (pp. 42-46). Palo Alto, CA: AAAI.*

Borgman, C.L. (1986). The user's mental model of an information retrieval system: an experiment on a prototype online catalog. *International Journal of Man-Machine Studies*, *24*(1), 47–64.

Bornstein, A.M. (2016, September 1). Is Artificial Intelligence Permanently Inscrutable? Retrieved August 29, 2017, from [http://nautil.us/issue/40/learning/is-artificial-intelligence-permanently-inscrutable]

Bostrom, N., & Yudkowsky, E. (2014). The ethics of artificial intelligence. In W. Ramsey and K. Frankish (Eds.), *Cambridge handbook of artificial intelligence* (pp. 316-344). Cambridge: Cambridge University Press.

Boy, G.A. (1991). *Intelligent assistant systems*. Academic Press. Retrieved from http://cds.cern.ch/record/217351

Bradshaw, J.M., Ford, K.M., Adams-Webber, J.R., & Boose, J.H. (1993). Beyond the repertory grid: new approaches to constructivist knowledge acquisition tool development. *International Journal of Intelligent Systems*, *8*(2), 287–333.

Brem, S.K., & Rips, L.J. (2000). Explanation and evidence in informal argument. *Cognitive Science*, *24*(4), 573–604.

Brézillon, P. (1994). Context needs in cooperative building of explanations. In *First European Conference on Cognitive Science in Industry* (pp. 443–450).

Brézillon, P., & Pomerol, J.-C. (1997). Joint cognitive systems, cooperative systems and decision support systems: A cooperation in context. In *Proceedings of the European Conference on Cognitive Science, Manchester* (pp. 129–139).

Brinton, C. (2017). *A framework for explanation of machine learning decisions*. [http://home.earthlink.net/~dwaha/research/meetings/ijcai17xai/2.%20(Brinton%20XAI17)%20A%20Framework%20for%20Explanation%20of%20Machine%20Learning%20Decisions.pdf]

Brock, D.C. (2018, Fall). Learning from Artificial Intelligence's previous awakenings: The history of expert systems. *The AI Magazine*, pp. 3-15,

Brodbeck, M. (1962). Explanation prediction and "imperfect" knowledge. *Scientific Explanation*, *3*, 231–273.

Buchanan, B., & Shortliffe, E. (1984a). Explanation as a topic of AI research. In *Rule-based expert systems: The MYCIN experiments of the Stanford Heuristic programming project* (pp. 331–337).

Buchanan, B., & Shortliffe, E. (1984b). Rule-based expert systems: the MYCIN experiments of the Stanford Heuristic Programming Project.

Burns, H., Luckhardt, C. A., Parlett, J. W., & Redfield, C. L. (2014). *Intelligent tutoring systems: Evolutions in design*. Psychology Press. Retrieved from https://books.google.com/books?hl=en&lr=&id=Xd6YAgAAQBAJ&oi=fnd&pg=PP1&dq=Burns,+H.,+Parlett,+J.W.,+and+Redfield,+C.L.+(1991).+Intelligent+tutoring+systems:+Evolutions+in+design.+New+York:+Psychology+Press.&ots=xTIs2rRzXo&sig=VWDqbJHcspTx6ZfVxJ34MIY6LIQ

Burton, R.R., & Brown, J.S. (1982). An investigation of computer coaching for informal learning activities. In D. Sleeman & J.S. Brown (Eds.). *Intelligent tutoring systems* (pp, 79-97). London: Academic Press.

Butts, C.T. (2004). Latent structure in multiplex relations. In *Proceedings of the Annual Conference of the North American Association for Computational Social and Organizational Science (NAACSOS)* (pp. 27–29). Citeseer.

Byrne, R. M. (1991). The Construction of Explanations. In *AI and Cognitive Science'90* (pp. 337–351). Springer.

Byrne, R. M. (1997). Cognitive processes in counterfactual thinking about what might have been. Retrieved from http://psycnet.apa.org/psycinfo/2003-02320-004

Byrne, R.M. (2002). Mental models and counterfactual thoughts about what might have been. *Trends in Cognitive Sciences*, *6*(10), 426–431.

Byrne, R.M. (2017). Counterfactual Thinking: From Logic to Morality. *Current Directions in Psychological Science*, *26*(4), 314–322.

Byrne, R. M., & McEleney, A. (2000). Counterfactual thinking about actions and failures to act. *Journal of Experimental Psychology: Learning, Memory, and Cognition*, *26*(5), 1318.

Cahour, B., & Forzy, J.-F. (2009). Does projection into use improve trust and exploration? An example with a cruise control system. *Safety Science*, *47*(9), 1260–1270.

Calin-Jageman, R., & Ratner, H. (2005). The role of encoding in the self-explanation effect. *Cognition and Instruction*, *23*(4), 523–543.

Cañas, A.J., Coffey, J.W., Carnot, M.J., Feltovich, P., Hoffman, R., Feltovich, J. & Novak, J.D. (2003). "A summary of literature pertaining to the use of concept mapping techniques and technologies for education and performance support." Report to The Chief of Naval Education and Training, prepared by the Institute for Human and Machine Cognition, Pensacola FL.

[https://ihmc.us/users/acanas/publications/conceptmaplitreview/ihmc%20literature%20review%20on%20concept%20mapping.pdf]

Cao, J., Kwan, I., Bahmani, F., Burnett, M., Fleming, S. D., et al. (2013). End-User Programmers in Trouble: Can the Idea Garden help them to help themselves? In *Visual Languages and Human-Centric Computing (VL/HCC), 2013 IEEE Symposium on* (pp. 151–158). IEEE. Retrieved from http://ieeexplore.ieee.org/abstract/document/6645260/

Carberry, S. (1990). Second international workshop on user modeling. *AI Magazine*, *11*(4), 57.

Carbonell, J. (1970). AI in CAI: An Artificial Intelligence Approach to Computer-Assisted Instruction, *IEEE Transactions on Human-Machine Systems, MMS-11*, 190-202.

Cardona-Rivera, R.E., Price, T., Winer, D., & Young, R. M. (2016). Question answering in the context of stories generated by computers. *Advances in Cognitive Systems*, *4*, 227–245.

Chai, J., Fang, R., Liu, C., & She, L. (2016). Collaborative language grounding toward situated human-robot dialogue. *AI Magazine*, *37*(4), 32–45.

Chai, J.Y., She, L., Fang, R., Ottarson, S., Littley, C., Liu, C., & Hanson, K. (2014). Collaborative effort towards common ground in situated human-robot dialogue. In *Proceedings of the 2014 ACM/IEEE International Conference on Human-robot Interaction* (pp. 33–40). ACM. Retrieved from http://dl.acm.org/citation.cfm?id=2559677

Chajewska, U., & Halpern, J.Y. (1997). Defining explanation in probabilistic systems. In *Proceedings of the Thirteenth conference on Uncertainty in artificial intelligence* (pp. 62–71). Morgan Kaufmann Publishers Inc. Retrieved from http://dl.acm.org/citation.cfm?id=2074234

Chakraborti, T., Sreedharan, S., Zhang, Y., & Kambhampati, S. (2017). Plan explanations as model reconciliation: Moving beyond *Explanation as Soliloquy*. In *Proceedings of the Twenty-Sixth International Joint Conference on Artificial Intelligence* (pp. 156-163). International Joint Conferences on Artificial Intelligence. [https://www.ijcai.org].

Chakraborty, S., Tomsett, R., Raghavendra, R., Harborne, D., Alzantot, M., Cerutti, F., … Gurram, P. (2017). Interpretability of deep learning models: A survey of results. Presented at the IEEE Smart World Congress 2017 Workshop on Distributed Analytics InfraStructure and Algorithms for Multi-Organization Federations (DAIS 2017).

Chalmers, M., & MacColl, I. (2003). Seamful and seamless design in ubiquitous computing. In *Workshop at the crossroads: The interaction of HCI and systems issues in UbiComp* (Vol. 8). New York: Springer.

[https://www.researchgate.net/profile/Matthew_Chalmers/publication/228551086].

Champlin, C., Bell, D., & Schocken, C. (2017). AI medicine comes to Africa's rural clinics. *IEEE Spectrum*, *54*(5), 42–48.

Chan, H., & Darwiche, A. (2012). On the robustness of most probable explanations. *ArXiv Preprint ArXiv:1206.6819*. Retrieved from https://arxiv.org/abs/1206.6819

Chandrasekaran, B., & Swartout, W.R. (1991). Explanations in knowledge systems: the role of explicit representation of design knowledge. *IEEE Expert*, *6*(3), 47–49.

Chandrasekaran, B., Tanner, M.C., & Josephson, J.R. (1989). Explaining control strategies in problem solving. *IEEE Expert*, *4*(1), 9–15.

Chandrashekar, G., & Sahin, F. (2014). A survey on feature selection methods. *Computers and Electrical Engineering, 40,* 16-28.

Chang, J., Gerrish, S., Wang, C., Boyd-Graber, J.L., & Blei, D.M. (2009). Reading tea leaves: How humans interpret topic models. In *Advances in neural information processing systems* (pp. 288–296). Retrieved from http://papers.nips.cc/paper/3700-reading-tea-leaves-how-humans-interpret-topic-models.pdf

Chater, N., & Oaksford, M. (2006 ). Mental mechanisms: Speculations on human causal learning and reasoning. In K. Fiedler and P. Juslin (Eds.), *Information sampling and adaptive* cognition (pp. 210-238). Cambridge: Cambridge University Press.

Chunpir, H. I., & Ludwig, T. (2017, July). A Software to Capture Mental Models. In *International Conference on Universal Access in Human-Computer Interaction* (pp. 393-409). New York: Springer

Cheng, P.W. (1997). From covariation to causation: A causal power theory. *Psychological Review*, *104*(2), 367.

Cheng, P. W. (2000). Causality in the Mind: Estimating Contextual and Conjunctive Causal Power. In *Explanation and Cognition* (pp. 227–253). MIT Press.

Chi, M. T., Bassok, M., Lewis, M. W., Reimann, P., & Glaser, R. (1989). Self-explanations: How students study and use examples in learning to solve problems. *Cognitive Science*, *13*(2), 145–182.

Chi, M.T., Leeuw, N., Chiu, M.-H., & LaVancher, C. (1994). Eliciting self-explanations improves understanding. *Cognitive Science*, *18*(3), 439–477.

Chi, M.T., & VanLehn, K.A. (1991). The content of physics self-explanations. *The Journal of the Learning Sciences*, *1*(1), 69–105.

Chin-Parker, S., & Bradner, A. (2010). Background shifts affect explanatory style: how a pragmatic theory of explanation accounts for background effects in the generation of explanations. *Cognitive Processes*, *11*, 227–249.

Chin-Parker, S., & Cantelon, J. (2016). Contrastive constraints guide explanation-based category learning. *Cognitive Science*, *40*, 1-11. [DOI: 10.1111/cogs.12405].

Clancey, W.J. (1981). Methodology for building an intelligent tutoring system. Retrieved from http://dl.acm.org/citation.cfm?id=891745

Clancey, W.J. (1983). The epistemology of a rule-based expert system—a framework for explanation. *Artificial Intelligence*, *20*(3), 215–251.

Clancey, W.J. (1984a). Details of the revised therapy algorithm. *Rule-Based Expert Systems: The MYCIN Experiments of the Stanford Heuristic Programming Project*. *Reading MA: Addison-Wesley*, 133–146.

Clancey, W.J. (1984b). Methodology for building an intelligent tutoring system. *Methods and Tactics in Cognitive Science*, 51–84.

Clancey, W.J. (1986a). From GUIDON to NEOMYCIN and HERACLES in twenty short lessons. *AI Magazine*, *7*(3), 40.

Clancey, W.J. (1986b). *Intelligent Tutoring Systems: A Tutorial Survey.* Stanford University Department of Computer Science. Retrieved from http://www.dtic.mil/docs/citations/ADA187066

Clancey, W.J. (1987). *Knowledge-based tutoring: the GUIDON program*. MIT Press. Retrieved from http://dl.acm.org/citation.cfm?id=SERIES9773.28590

Clancey, W.J. (1988). The knowledge engineer as student: Metacognitive bases for asking good questions. In *Learning issues for intelligent tutoring systems* (pp. 80–113). Springer. Retrieved from https://link.springer.com/chapter/10.1007/978-1-4684-6350-7_5

Clancey, W.J. (1992). Model construction operators. *Artificial Intelligence*, *53*(1), 1–115. https://doi.org/10.1016/0004-3702(92)90037-X

Clancey, W.J. (1993). Notes on "Epistemology of a rule-based expert system." *Artificial Intelligence*, *59*(1–2), 197–204.

Clancey, W.J., & Letainger, R. (1984). NEOMYCIN: Reconfiguring a rule-based expert system for application to teaching. In *Readings in medical artificial intelligence: The first decade* (pp. 361–381).

Clark, H.H., Brennan, S.E., & others. (1991). Grounding in communication. *Perspectives on Socially Shared Cognition*, *13*(1991), 127–149.

Clarke, A.A., & Smyth, M.G.G. (1993). A co-operative computer based on the principles of human co-operation. *International Journal of Man-Machine Studies*, *38*(1), 3–22.

Clausner, T. (1994). Commonsense knowledge and conceptual structure in container metaphors. *In Proceedings of Sixteenth Annual Conference of the Cognitive Science Society*, 189–194.

Clement, J. (1988). Observed methods for generating analogies in scientific problem solving. *Cognitive Science*, *12*(4), 563–586.

Clos, J., Wiratunga, N., & Massie, S. (2017). Towards Explainable Text Classification by Jointly Learning Lexicon and Modifier Terms. *IJCAI-17 Workshop on Explainable Artificial Intelligence (XAI)*. Retrieved from http://home.earthlink.net/~dwaha/research/meetings/ijcai17-xai/3.%20(Clos+%20XAI-17)%20Towards%20Explainable%20Text%20Classification%20by%20Jointly%20Learning%20Lexicon%20and%20Modifier%20Terms.pdf

Codella, N. C., Hind, M., Ramamurthy, K. N., Campbell, M., Dhurandhar, A., Varshney, K. R., ... & Mojsilovic, A. (2018). TED: Teaching AI to Explain its Decisions. *arXiv preprint arXiv:1811.04896*.

Conant, R. C., & Ashby, W. R. (1970). Every good regulator of a system must be a model of that system. *International Journal of Systems Science*, *1*(2), 89–97.

Cooper, A. (2004). *The inmates are running the asylum*. Carmel, IN: Sams Publishing.

Corbett, A.T., & Anderson, J.R. (1991). Feedback control and learning to program with the CMU Lisp Tutor. Presented at the Annual Meeting of the American Educational Research Association, Chicago, IL.

Core, M.G., Lane, H.C., Van Lent, M., Gomboc, D., Solomon, S., & Rosenberg, M. (2006). Building explainable artificial intelligence systems. In *AAAI* (pp. 1766–1773). Retrieved from http://www.aaai.org/Papers/IAAI/2006/IAAI06-010.pdf

Craik, K.J.W. (1967). *The nature of explanation* (Vol. 445). CUP Archive. Retrieved from https://books.google.com/books?hl=en&lr=&id=wT04AAAAIAAJ&oi=fnd&pg=PA1&dq=Craik,+K.+(1943).+The+nature+of+explanation.+Cambridge+University+Press.&ots=07rYVqgWB4&sig=2m_qg42W8Q6AQDeA-tAcS_kQD4w

Crandall, B., Klein, G.A., & Hoffman, R.R. (2006). *Working minds: A practitioner's guide to cognitive task analysis*. MIT Press. Retrieved from

https://books.google.com/books?hl=en&lr=&id=ZfcVGsJlyhMC&oi=fnd&pg=PR5&dq=Crandall,+B.,+Klein,+G.,+%26+Hoffman,+R.R.+(2006).+Working+minds:+A+practition

er%27s+guide+to+cognitive+task+analysis.+MIT+Press.&ots=oh9NAdHbc7&sig=ZKvp 1Zusoy5sZbOwiIlAW-YsPcc

Craven, M.W., & Shavlik, J.W. (1996). *Extracting comprehensible models from trained neural networks*. University of Wisconsin, Madison. Retrieved from http://research.cs.wisc.edu/machine-learning/shavlik-group/craven.thesis.pdf

Croskerry, P. (2009). A universal model of diagnostic reasoning. *Academic Medicine*, *84*(8), 1022–1028.

Darden, L., Machamer, P., & Crave, C (2000). Thinking about mechanisms. *Philosophy of Science*, *67*, 1-25.

Darlington, K. (2013). Aspects of intelligent systems explanation. *Universal Journal of Control and Automation, 1*, 40-51.

Davies, M. (2011). Concept mapping, mind mapping and argument mapping: what are the differences and do they matter? *Higher Education*, *62*(3), 279–301.

Degani, A. (2004). *Taming HAL: Designing interfaces beyond 2001*. Springer. Retrieved from https://books.google.com/books?hl=en&lr=&id=DMQWDAAAQBAJ&oi=fnd&pg=PP1 &dq=Degani,+A.+(2013).+Taming+HAL:+Designing+interfaces+beyond+2001.+New+ York:+Macmillan.&ots=lMbNp1lpm3&sig=dQKTKE2pZGVOTDnRA5g8x3PSpi0

De Greef, H.P., & Neerincx, M.A. (1995). Cognitive support: Designing aiding to support human knowledge. *International Journal of Human-Computer Studies, 42*, 531-571.

Dhaliwal, J. S., & Benbasat, I. (1996). The use and effects of knowledge-based system explanations: theoretical foundations and a framework for empirical evaluation. *Information Systems Research*, *7*(3), 342–362.

De Kleer, J., & Brown, J. S. (2014). Assumptions and ambiguities in mechanistic mental models. In *Mental models* (pp. 163-198). Psychology Press.

Dodge, J., Penney, S., Hilderbrand, C., Anderson, A., & Burnett, M. (2018). How the Experts do it: Assessing and explaining agent behaviors in real-time strategy games. In *Proceedings of CHI 2018*, Paper 562. New York: Association for Computing Machinery.

Dodson, T., & Goldsmith, J. (2011). A natural language argumentation interface for explanation generation in Markov decision processes. In *Proceedings of the 22nd International Joint Conference on Artificial Intelligence* IJCAI-11 [DOI: 10.1007/978-3-642-24873-3_4]

Dodson, T., Mattei, N., & Goldsmith, J. (2011, October). A natural language argumentation interface for explanation generation in Markov decision processes. In *International Conference on Algorithmic Decision Theory* (pp. 42-55). Springer, Berlin, Heidelberg.

Doignon, J., & Falmagne, J. (1985). Spaces for the assessment of knowledge. *International Journal of Man-Machine Studies*, *23*(2), 175–196. https://doi.org/10.1016/S0020-7373(85)80031-6

Donahue, J., Hendricks, A., L., Guadarrama, S., Rohrbach, M., Venugopalan, S., Saenko, K., & Darrell, T. (2015). Long-term recurrent convolutional networks for visual recognition and description. . In *Proceedings of Computer Vision and Pattern Recognition 2015*. [https://arxiv.org/abs/1411.4389]. (pp. 2625–2634).

Doran, D., Schulz, S., & Besold, T. R. (2017). What does explainable AI really mean? A new conceptualization of perspectives. *ArXiv Preprint ArXiv:1710.00794*.

Doshi-Velez, F., & Kim, B. (2017). A Roadmap for a Rigorous Science of Interpretability. *ArXiv Preprint ArXiv:1702.08608*. Retrieved from https://arxiv.org/abs/1702.08608

Doshi-Velez, F., & Kim, B. (2017a). Towards a rigorous science of interpretable machine learning [arXiv:1702.08608v2].

Doshi-Velez, F., Kortz, M., Budish, R., Bavitz, C., Gershman, S., O'Brien, D., … Wood, A. (2017). Accountability of AI under the law: The role of explanation. *ArXiv:1711.01134 [Cs, Stat]*. Retrieved from http://arxiv.org/abs/1711.01134

Douven, I. (2011). Abduction. In *Stanford Encyclopedia of Philosophy*. Retrieved from https://plato.stanford.edu/entries/abduction/

Doyle, D., Tsymbal, A., & Cunningham, P. (2003). *A review of explanation and explanation in case-based reasoning*. Trinity College Dublin, Department of Computer Science. Retrieved from http://ai2-s2-pdfs.s3.amazonaws.com/f3aa/7f2e9c820527cff2dca84227dd5a37011c35.pdf

Doyle, J., Radzicki, M., & Trees, W. (2008). Measuring change in mental models of complex dynamic systems. *Complex Decision Making*, 269–294.

Doumas, L.A., & Hummel, J.E. (2005). Approaches to modeling human mental representations: What works, what doesn't and why. *The Cambridge Handbook of Thinking and Reasoning, Ed. KJ Holyoak & RG Morrison*, 73–94.

Dourish, P. (1995). Accounting for system behaviour: Representation, reflection and resourceful action. In *Proceedings of CIC'95 Conference on Computers in Context* (pp. 145-170). Aarhus University, Aarhus Denmark.

Druzdzel, M.J. (1990). *Scenario based explanations for Bayesian decision support systems*. Technical Report CMU–EPP–1990–03–04, Department of Engineering and Public Policy, Carnegie Mellon University, Pittsburgh, PA.

Druzdzel, M. J. (1996a). Explanation in probabilistic systems: Is it feasible? will it work. In *Proceedings of the Fifth International Workshop on Intelligent Information Systems (WIS-96)* (pp. 12–24). Retrieved from https://pdfs.semanticscholar.org/3081/81b6b83dab171a2f1f9cf0afb8562b6c982d.pdf

Druzdzel, M.J. (1996b). Qualitative verbal explanations in Bayesian belief networks. *AISB QUARTERLY*, 43–54.

Druzdzel, M.J., & Henrion, M. (1990). Using scenarios to explain probabilistic inference. In *Working notes of the AAAI-90 Workshop on Explanation* (pp. 133–141). Retrieved from http://pitt.edu/~druzdzel/psfiles/explan.pdf

Duckstein, L., & Nobe, S.A. (1997). Q-analysis for modeling and decision making. *European Journal of Operational Research*, *103*(3), 411–425. https://doi.org/10.1016/S0377-2217(97)00308-1

Dunbar, K. (1995). How scientists really reason: Scientific reasoning in real-world laboratories. *The Nature of Insight*, *18*, 365–395.

Dwork, C., Hardt, M., Pitassi, T., Reingold, O., & Zemel, R. (2012). Fairness through awareness. In *Proceedings of ITCS '12: The 3rd Innovations in Theoretical Computer Science Conference* (pp. 214-226). New York: Association for Computing Machinery.

Dzindolet, M.T., Peterson, S.A., Pomranky, R.A., Pierce, L.G., & Beck, H.P. (2003). The role of trust in automation reliance. *International Journal of Human-Computer Studies*, *58*(6), 697–718.

Dzindolet, M.T., Pierce, L.G., Beck, H.P., & Dawe, L.A. (2002). The perceived utility of human and automated aids in a visual detection task. *Human Factors, 44*, 79-94.

Dzindolet, M.T., Pierce, L., Pomranky, R., Peterson, S., & Beck, H. (2001). Automation reliance on a combat identification system. In *Proceedings of the Human Factors and Ergonomics Society Annual Meeting* (Vol. 45, pp. 532–536). SAGE Publications. https://doi.org/10.1177/154193120104500456

Eaves Jr, B.S., & Shafto, P. (2016). Toward a general, scaleable framework for Bayesian teaching with applications to topic models. *ArXiv Preprint ArXiv:1605.07999*. Retrieved from https://arxiv.org/abs/1605.07999

Ebel, R. (1974). And still the dryads linger. *American Psychologist*, *29*(7), 485–492. https://doi.org/10.1037/h0036780

Ebener, S., Khan, A., Shademani, R., Compernolle, L., Beltran, M., Lansang, M. A., & Lippman, M. (2006). Knowledge mapping as a technique to support knowledge translation. *Bulletin of the World Health Organization*, *84*(8), 636–642.

Eberle, R., Kaplan, D., & Montague, R. (1961). Hempel and Oppenheim on explanation. *Philosophy of Science*, *28*(4), 418–428. https://doi.org/10.1086/287828

Erhan, D., Bengio, Y., Courville, A., & Vincent, P. (2009). Visualizing higher-layer features of a deep network. *University of Montreal*, *1341*, 3.

Einhorn, H.J., & Hogarth, R. M. (1986). Judging probable cause. *Psychological Bulletin*, *99*(1), 3.

Eining, M. M., & Dorr, P. B. (1991). The impact of expert system usage on experiential learning in an auditing setting. *Journal of Information Systems*, *5*(1), 1–16.

Eiter, T., & Lukasiewicz, T. (2002). Complexity results for structure-based causality. *Artificial Intelligence, 142, 53-89.*

Eiter, T., & Lukasiewicz, T. (2006). Causes and explanations in the structural-mode approach: tractable cases. *Artificial Intelligence, 170,* 542-580.

Elenberg, E.R., Dimakis, A.G., Feldman, M., & Karbashi, A. (2017). Streaming weak submodularity: Interpreting neural networks on the fly. In *Proceedings of the 31st Conference on Neural Information Processing Systems (NIPS 2017)*. City, State*:* NIPS Foundation.[arXiv:1703.02647v3].

Eliot, L.B. (1986). Analogical problem-solving and expert systems. *IEEE Expert*, *1*(2), 17–28.

Elizalde, F., Sucar, E., Reyes, A., & deBuen, P. (2007). "An MDP approach for explanation generation." Workshop on Explanation-aware Computing. Menlo Park, CA: Association for the Advancement of Artificial Intelligence.

Ellis, C. (1989). Explanation in intelligent systems. In C. Ellis (Ed.), *Expert knowledge and explanation* (pp. 108-128). New York: John Wiley.

Erlich, K. (1996). Applied mental models in human–computer interaction. *Mental Models in Cognitive Science: Essays in Honour of Phil Johnson-Laird*, 223.

Facione, P. A., & Gittens, C. A. (2015). Mapping decisions and arguments. *Inquiry: Critical Thinking Across the Disciplines*, *30*(2), 17–53.

Fails, J. A., & Olsen Jr, D. R. (2003). Interactive machine learning. In *Proceedings of the 8th international conference on Intelligent user interfaces* (pp. 39–45). ACM.

Falkenhainer, B. (1990). A unified approach to explanation and theory formation. In J. Shrager and P. Langley (Eds.), *Computational models of scientific discovery and theory formation.* (pp. 157-196). San Mateo, CA: Morgan Kaufman.

Falkenhainer, B., Forbus, K.D., & Gentner, D. (1989). The structure-mapping engine: Algorithm and examples. *Artificial Intelligence*, *41*(1), 1–63.

Fallon, C.K., & Blaha, L.M. (2018). Improving automation transparency: Addressing some of machine learning's unique challenges. In D. Schmorrow and C.M. Fidopiastis (Eds.) *Augmented cognition: Users and contexts: Proceedings of the 12th International Conference, part II* (pp. 245-254). New York: Springer.

Falls, J.A., & Olsen, D.R. (2003). Interactive machine learning. Presentation at the International Conference on Intelligent User Interfaces (IUI'03), New York: Association for Computing Machinery.

Féraud, R., & Clérot, F. (2002). A methodology to explain neural network classifications. *Neural Networks, 15*, 237-246.

Feigl, H., & Maxwell, G. (1962). Scientific Explanation, Space, and Time: Minnesota Studies in the Philosophy of Science. Retrieved from https://philpapers.org/rec/FEISES

Feldmann, F. (2018). Measuring machine learning interpretability. Seminar Report, University of Heidelberg.                                        [https://hci.iwr.uni-heidelberg.de/system/files/private/downloads/860270201/felix_feldmann_eml2018_report.pdf]

Felten, E. (2017). What does it mean to ask for an "explainable" algorithm? Retrieved August 29, 2017, from https://freedom-to-tinker.com/2017/05/31/what-does-it-mean-to-ask-for-an-explainable-algorithm/

Feltovich, P. J., Coulson, R. L., & Spiro, R. J. (2001). Learners'(mis) understanding of important and difficult concepts: A challenge to smart machines in education. *Smart Machines in Education*, 349–375.

Feltovich, P.J., Hoffman, R.R., Woods, D., & Roesler, A. (2004). Keeping it too simple: How the reductive tendency affects cognitive engineering. *IEEE Intelligent Systems*, *19*(3), 90–94.

Fernbach, P. M., Sloman, S. A., Louis, R. S., & Shube, J. N. (2012). Explanation fiends and foes: How mechanistic detail determines understanding and preference. *Journal of Consumer Research*, *39*(5), 1115–1131.

Fiebrink, R., Cook, P.R., & Trueman, D. (2012). Human model evaluation using interactive supervised learning. Presentation at Computer-Human Interaction (CHI 2002). New York: Association for Computing Machinery.

Fonseca, B.A., & Chi, M.T.H. (2011). Instruction based on self-explanation. In R.E. Mayer and P.A. Alexander (Eds.), *Handbook of research on learning and instruction* (296-321). Boca Raton, Taylor and Francis.

Forbus, K. D., & Feltovich, P. J. (2001). *Smart machines in education*. AAAI press.

Forbus, K.D., Gentner, D., & Law, K. (1995). MAC/FAC: A model of similarity-based retrieval. *Cognitive Science*, *19*(2), 141–205.

Ford, K.M., & Adams-Webber, J.R. (1992). Knowledge acquisition and constructivist epistemology. *The Psychology of Expertise: Cognitive Research and Empirical AI*, 121–136.

Ford, K.M., Cañas, A.J., & Coffey, J. (1993). Participatory explanation (pp. 111–115). Presented at the FLAIRS 93: Sixth Florida Artificial Intelligence Research Symposium, Ft. Lauderadale, FL: Institute for Human and Machine Cognition.

Ford, K.M., Cañas, A.J., Coffey, J.W., Andrews, E.J., Schad, N., & Stahl, H. (1992). Interpreting functional images with NUCES: Nuclear Cardiology Expert System. In *Fifth Florida Artificial Intelligence Research Symposium (FLAIRS)* (pp. 85–90).

Ford, K.M., Coffey, J.W., Cañas, A.J., Andrews, E.J., & Turner, C.W. (1996). Diagnosis and Explanation by a Nuclear Cardiology Expert System. *International Journal of Expert Systems*, *9*, 499-506.

Fox, J., Glasspool, D., Grecu, D., Modgil, S., South, M., & Patkar, V. (2007). Argumentation-based inference and decision making–A medical perspective. *IEEE Intelligent Systems*, *22*(6).

Fox, M., Long, D., & Magazzeni, D. (2017). Explainable Planning. In *IJCAI-17 Workshop on Explainable Artificial Intelligence (XAI)*. Retrieved from http://home.earthlink.net/~dwaha/research/meetings/ijcai17-xai/4.%20(Fox+%20XAI-17)%20Explainable%20Planning.pdf

Fox, J., & Patkar, V. (2007, November). Argumentation-based inference and decision making--A medical perspective. *IEEE: Intelligent Systems* [DOI: 10.1109/MIS.2007.102 ].

Francis, G. (2000). Designing multifunction displays: An optimization approach. *International Journal of Cognitive Ergonomics, 4,* 107-122.

Frank, J., McGuire, K., Moses, H., & Stephenson, J. (2016). Developing decision aids to enable human spaceflight autonomy. *AI Magazine*, *37*(4), 46–54.

Frederiksen, J.R., White, B.Y., Collins, A., & Eggan, G. (1988). Intelligent tutoring systems for electronic troubleshooting. *Intelligent Tutoring Systems: Lessons Learned*. Mahwah, NJ: Erlbaum.

Freitas, A. A. (2014). Comprehensible classification models: a position paper. *ACM SIGKDD Explorations Newsletter*, *15*(1), 1–10.

Freuder , E.C., Likitvivatanavong, C., & Wallace, R.J. (2001). Deriving explanations and implications for constraint satisfaction problems. In Proceedings of the *International Conference on Principles and Practice of Constraint Programming* (pp. 585-589). New York: Springer.

Friedman, B., & Nissenbaum, H. (1996). Bias in computer systems. *ACM Transactions on Information Systems (TOIS)*, *14*(3), 330–347. https://doi.org/10.1145/230538.230561

Friedman, S., Forbus, K., & Sherin, B. (2018). Representing, Running, and Revising Mental Models: A Computational Model. *Cognitive Science*, 42, 1110-1145.

Fukui, A., Park, D. H., Yang, D., Rohrbach, A., Darrell, T., & Rohrbach, M. (2016). Multimodal compact bilinear pooling for visual question answering and visual grounding. *ArXiv Preprint ArXiv:1606.01847*.

Galinsky, A.D., Maddux, W. W., Gilin, D., & White, J. B. (2008). Why it pays to get inside the head of your opponent: The differential effects of perspective taking and empathy in negotiations. *Psychological Science*, *19*(4), 378–384.

Gan, C., Wang, N., Yang, Y., Yeung, D.-Y., & Hauptmann, A. G. (2015). Devnet: A deep event network for multimedia event detection and evidence recounting. In *Proceedings of the IEEE Conference on Computer Vision and Pattern Recognition* (pp. 2568–2577). Retrieved from http://www.cv-foundation.org/openaccess/content_cvpr_2015/html/Gan_DevNet_A_Deep_2015_CVPR_paper.html

Gardner, A. v.d.L. (1987). Machine learning. In *Encyclopedia of Artificial Intelligence* (Vol. 1, pp. 464–488).

Gary, M.S., & Wood, R.E. (2011). Mental models, decision rules, and performance heterogeneity. *Strategic Management Journal*, *32*(6), 569–594.

Gautier, P.O., & Gruber, T. R. (1993). *Generating explanations of device behavior using compositional modeling and causal ordering*. Knowledge Systems Laboratory, Computer Science Department, Univ.

Geldard, F. (1939). "Explanatory principles" in psychology. *The Psychological Review*, *46*(5), 411–424. https://doi.org/10.1037/h0059714

Gentner, D., Holyoak, K.J., & Kokinov, B.N. (2001). *The analogical mind: Perspectives from cognitive science*. MIT press. Retrieved from

https://books.google.com/books?hl=en&lr=&id=RfQX9wuf2cC&oi=fnd&pg=PR7&dq=Gentner,+D.,+Holyoak,+K.J.,+%26+Kokinov,+B.N.+(Eds.).+(2001).+The+analogical+m

ind:+Perspectives+from+cognitive+science.+Cambridge,+MA:+Bradford+Books.&ots=
MujLlRNODk&sig=6vQfE8vjufDdwVgiWMGRFzhvW8I

Gentner, D., & Schumacher, R. M. (1986). Use of structure mapping theory for complex systems. In *Proceedings of the 1986 IEEE international conference on systems, man, and cybernetics* (Vol. 1, pp. 14–17). IEEE Atlanta, GA. New York. Retrieved from http://groups.psych.northwestern.edu/gentner/newpdfpapers/GentnerSchumacher86.pdf

Gentner, D., & Stevens, A. (1983). *Mental models*. New York, Psychology Press.

George, F.H. (1953). Logical constructs and psychological theory. The Psychological Review, 60(1), 1–6. https://doi.org/10.1037/h0057812

Ghallab, M., Nau, D., & Traverso, P. (2004). *Automated Planning: theory and practice*. Elsevier. ISBN: 9780080490519

Giffin, C., Wilkenfeld, D. A., & Lombrozo, T. (2017). The explanatory effect of a label: Explanations with named categories are more satisfying. *Cognition*, *168*, 357–369. https://doi.org/10.1016/j.cognition.2017.07.011

Gigerenzer, G. (1993). The bounded rationality of probabilistic mental models. In *This chapter is based on a lecture delivered at Harvard University, Oct 2, 1991*. Taylor & Frances/Routledge.

Gigerenzer, G., Hoffrage, U., & Kleinbölting, H. (1991). Probabilistic mental models: A Brunswikian theory of confidence. *Psychological Review*, *98*(4), 506-528.

Gilpin, L.H., et al. (2018). Explaining explanation: An approach to evaluating interpretability of machine learning.[arXiv.org > cs > arXiv:1806.00069v2].

Ginet, C. (2008). In defense of a non-causal account of reasons explanations. *The Journal of Ethics, 12,* 229-237.

Giordano, L., & Schwind, C. (2004). Conditional logic of actions and causation. *Artificial Intelligence, 157,* 239-279.

Gkatzia, D., Lemon, O., & Rieser, V. (2016). Natural Language Generation enhances human decision-making with uncertain information. *ArXiv Preprint ArXiv:1606.03254*. Retrieved from https://arxiv.org/abs/1606.03254

Glennan, S. (2002). Rethinking mechanistic explanation. *Philosophy of Science, 69,* S342-S353.

Goetz, E.T., & Sadoski, M. (1995). Commentary: The perils of seduction: Distracting details or incomprehensible abstractions? *Reading Research Quarterly*, 500–511.

Goguen, J.A., Weiner, J.L., & Linde, C. (1983). Reasoning and natural explanation. *International Journal of Man-Machine Studies*, *19*(6), 521–559. https://doi.org/10.1016/S0020-7373(83)80070-4

Goodfellow, I.J., Shlens, J., & Szegdy, C. (2015). Explaining and harnessing adversarial examples. In Proceedings of the International Conference on Learning Representations, ICLR 2015. [arXiv:1412.6572v3].

Goodman, B., & Flaxman, S. (2016). European Union regulations on algorithmic decision-making and a "right to explanation." Presented at the ICML Workshop on Human Interpretability in Machine Learning, New York, NY.

Gopnik, A. (2000). Explanation as orgasm and the drive for causal knowledge: The function, evolution, and phenomenology of the theory-formation system. In F. Keil & R. Wilson (Eds.), *Cognition and Explanation* (pp. 299–323). Cambridge: MIT Press..

Goyal, Y., Khot, T., Summers-Stay, D., Batra, D., & Parikh, D. (2016a). Making the V in VQA matter: Elevating the role of image understanding in Visual Question Answering. *ArXiv Preprint ArXiv:1612.00837*.

Goyal, Y., Mohapatra, A., Parikh, D., & Batra, D. (2016b). Interpreting visual question answering models. In *ICML Workshop on Visualization for Deep Learning* (Vol. 2). Retrieved from https://pdfs.semanticscholar.org/72ce/bd7d046080899703ed3cd96e3019a9f60f13.pdf

Goyal, Y., Mohapatra, A., Parikh, D., & Batra, D. (2016c). Towards Transparent AI Systems: Interpreting Visual Question Answering Models. *ArXiv Preprint ArXiv:1608.08974*. Retrieved from https://arxiv.org/abs/1608.08974

Graesser, A.C., Baggett, W., & Williams, K. (1996). Question-driven explanatory reasoning. *Applied Cognitive Psychology*, *10*(7), 17–31.

Graesser, A. C., & Franklin, S. P. (1990). QUEST: A cognitive model of question answering. *Discourse Processes*, *13*(3), 279–303.

Graesser, A. C., Gordon, S. E., & Brainerd, L. E. (1992). QUEST: A model of question answering. *Computers & Mathematics with Applications*, *23*(6–9), 733–745.

Graesser, A. C., Lang, K. L., & Roberts, R. M. (1991). Question answering in the context of stories. *Journal of Experimental Psychology: General*, *120*(3), 254.

Greenbaum, J., & Kyng, M. (1991). *Design at work: Cooperative design of computer systems*. Boca Raton, FL: CRC Press.

Greer, J.E., Falk, S., Greer, K.J., & Bentham, M.J. (1994). Explaining and justifying recommendations in an agriculture decision support system. *Computers and Electronics in Agriculture*, *11*(2), 195–214.

Gregor, S., and Benbasat, I. (1999). Explanations from Intelligent Systems: Theoretical Foundations and Implications for Practice. *MIS Quarterly, 23*, 497-530.

Grice, H.P. (1975). Logic and conversation. In P. Cole and J. Morgan (eds.) *Syntax and semantics Volume 3: Speech acts*. Academic Press, New York.

Groce, A., Kulesza, T., Zhang, C., Shamasunder, S., Burnett, M., Wong, W.-K., et al. (2014). You are the only possible oracle: Effective test selection for end users of interactive machine learning systems. *IEEE Transactions on Software Engineering*, *40*(3), 307–323.

Grotzer, T. A. (2003). Learning to understand the forms of causality implicit in scientifically accepted explanations. *Studies in Science Education, 39*, 1-74.

Gruber, T. (1991). Learning why by being told what. *IEEE Expert*, *6*(4), 65–75. https://doi.org/10.1109/64.85922

Guerlain, S. (1995, October). Using the critiquing approach to cope with brittle expert systems. In *Proceedings of the Human Factors and Ergonomics Society Annual Meeting* (Vol. 39, No. 4, pp. 233-237). Sage CA: Los Angeles, CA: SAGE Publications.

Gupta, M., et al. (2016). Monotonic calibrated interpolated look-up tables. *Journal of Machine Learning Research, 17*, 1-47.

Haddawy, P., Jacobson, J., & Kahn Jr, C. E. (1997). BANTER: a Bayesian network tutoring shell. *Artificial Intelligence in Medicine*, 10(2), 177–200.

Hajan, S., Domingo-Ferrer, J., Monreale, A., & Pedreschi, D. (2014). Discimination- and privacy-aware patterns. *Data Mining and Knowledge Discovery, 29*, 1733-1782.

Halasz, F. G., & Moran, T. P. (1983). Mental models and problem solving in using a calculator. In *Proceedings of the SIGCHI conference on Human Factors in Computing Systems* (pp. 212–216). ACM.

Halpern, J. Y., & Pearl, J. (2005a). Causes and explanations: A structural-model approach. Part I: Causes. *The British Journal for the Philosophy of Science*, *56*(4), 843–887.

Halpern, J. Y., & Pearl, J. (2005b). Causes and explanations: A structural-model approach. Part II: Explanations. *The British Journal for the Philosophy of Science*, *56*(4), 889–911.

Hancock, P. A., Billings, D. R., Schaefer, K. E., Chen, J. Y., De Visser, E. J., & Parasuraman, R. (2011). A meta-analysis of factors affecting trust in human-robot interaction. *Human Factors*, *53*(5), 517–527.

Hankinson, R. J. (2001). *Cause and explanation in ancient Greek thought*. Oxford University Press.

Hardt, S. L. (1987). Connection machines. In *Encyclopedia of Artificial Intelligence* (Vol. 1, pp. 199–200).

Harford, T. (2014, March 28). Big data: Are we making a big mistake? *The Financial Times*.

Harman, G.H. (1965). The Inference to the Best Explanation. *Philosophical Review*, *74*(1), 88–95.

Hasling, D.W., Clancey, W.J., & Rennels, G. (1984). Strategic explanations for a diagnostic consultation system. *International Journal of Man-Machine Studies*, *20*(1), 3–19.

Hawkins, J. (2017). Can we copy the brain? What intelligent machines need to learn from the Neocortex. *IEEE Spectrum*, *54*(6), 34–71.

Hayes, B., and Shah, J.A. (2017). Improving robot controller transparency through autonomous policy explanation. In P*roceedings of Human-Robot Interaction, HRI'17*. New York: Association for Computing Machinery.

Hayes-Roth, F. (1987). Explanation. In *Encyclopedia of Artificial Intelligence* (Vol. 1, pp. 298–300).

Hayes-Roth, F., Waterman, D., & Lenat, D. (1984). Building expert systems. Boston: Addison-Wesley.

Haynes, S.R., Cohen, M.A., & Ritter, F.E. (2008). Designs for explaining intelligent agents. *International Journal of Human-Computer Studies*, *67*, 90-110.

Heckerman, D., & Shachter, R. (1994). A decision-based view of causality. In *Proceedings of the Tenth international conference on Uncertainty in artificial intelligence* (pp. 302–310). Morgan Kaufmann Publishers Inc. Retrieved from http://dl.acm.org/citation.cfm?id=2074433

Hegarty, M. (2004). Mechanical reasoning by mental simulation. *Trends in Cognitive Sciences*, *8*(6), 280–285.

Hegarty, M., Just, M. A., & Morrison, I. R. (1987). *Mental models of mechanical systems: individual differences in qualitative and quantitative reasoning*. Report, Department of Psychology, Carnegie-Mellon University, Pittsburgh PA

Heller, J. (1994). Semantic structures. In *Knowledge structures* (pp. 117–149). Springer. Retrieved from http://link.springer.com/chapter/10.1007/978-3-642-52064-8_4

Hempel, C. G., & Oppenheim, P. (1948). Studies in the Logic of Explanation. *Philosophy of Science*, *15*(2), 135–175.

Hendricks, L. A., Akata, Z., Rohrbach, M., Donahue, J., Schiele, B., & Darrell, T. (2016). Generating visual explanations. In *European Conference on Computer Vision* (pp. 3–19). Springer. Retrieved from http://link.springer.com/chapter/10.1007/978-3-319-46493-0_1

Henrion, M., & Druzdzel, M. J. (1990a). Qualitative and linguistic explanations of probabilistic reasoning in belief networks. In *Proceedings of the Third International Conference on Information Processing and Management of Uncertainty in Knowledge-Based Systems (IPMU)* (pp. 225–227).

Henrion, M., & Druzdzel, M.J. (1990b). Qualitative propagation and scenario-based scheme for exploiting probabilistic reasoning. In *Proceedings of the Sixth Annual Conference on Uncertainty in Artificial Intelligence* (pp. 17–32). Elsevier Science Inc. Retrieved from http://dl.acm.org/citation.cfm?id=757035

Henschen, L. (1987). Causal reasoning. In *Encyclopedia of Artificial Intelligence* (Vol. 2, pp. 827–832).

Herlocker, J.L., Konstan, J.A., & Riedl, J. (2000). Explaining collaborative filtering recommendations. In *Proceedings of the 2000 ACM conference on Computer supported cooperative work* (pp. 241–250). ACM. Retrieved from http://dl.acm.org/citation.cfm?id=358995

Hesslow, G. (1988). The problem of causal selection. In D.J. Hilton (Ed.), *Contemporary science and natural explanation: commonsense conceptions of causality* (pp. 11-32). London: Harvester Press.

Hilton, D.J. (1990). Conversational processes and causal explanation. *Psychological Bulletin*, *107*(1), 65.

Hilton, D.J. (1996). Mental models and causal explanation: Judgements of probable cause and explanatory relevance *Thinking & Reasoning, 2*(4), 273-308.

Hilton, D.J. (2016). Social attribution and explanation. In M. Waldman (Ed.), *Oxford Handbook of causal reasoning*. Oxford: Oxford University Press.

Hilton, D.J., & Erb, H.-P. (1996). Mental models and causal explanation: Judgements of probable cause and explanatory relevance. *Thinking and Reasoning*, *2*(4), 273–308. https://doi.org/10.1080/135467896394447

Hilton, D.J., McClure, J.L., & Slugoski, B.R. (2005). The course of events: Counterfactuals, causal sequences and explanation. In D.R. Mandel, D.J. Hilton, & P. Catellani (Eds.), *The psychology of counterfactual thinking* (pp. 44-60). Routledge.

Hilton, D.J., McClure, J., & Slugoski, B. (2005). Counterfactuals, conditionals and causality: A social psychological perspective. In D.R. Mandel, D.J. Hilton, and P. Catellani (Eds.), *The psychology of counterfactual thinking* (pp. 4-60). Bristol, UK: Routledge.

Hilton, D.J., McClure, J., & Sutton, R.M. (2009). Selecting explanations from causal chains: Do statistical principles explain preferences for voluntary causes? *European Journal of Social Psychology, 39*, 1-18.

Hilton, D.J., & Slugoski, B.R. (1986) Knowledge-based causal attribution: The abnormal conditions focus model. *Psychological Review, 93*(1) 75-88.

Hinton, G., Vinyals, O., & Dean, J. (2015). Distilling the knowledge in a neural network. *ArXiv Preprint ArXiv:1503.02531*. Retrieved from https://arxiv.org/abs/1503.02531

Hitchcock, C. (2003)**.** Contrastive Explanation. In Martijn Blaauw (Ed.), *Contrastivism in Philosophy: New Perspectives*. New York: Routledge.

Hitchcock, C., & Woodward, J. (2003). Explanatory generalizations, part II: Plumbing explanatory depth. *Noûs, 37*(2), 181–199.

Hoffman, R.R. (1995). Monster Analogies. *AI Magazine, 16*(3), 11–35.

Hoffman, R. R. (1998). AI models of verbal/conceptual analogy. *Journal of Experimental & Theoretical Artificial Intelligence, 10*(2), 259–286.

Hoffman, R.R. (2017a). A taxonomy of emergent trusting in the human-machine relationship. In P. Smith & R. R. Hoffman (Eds.), *Cognitive systems engineering: The future for a changing world*. Boca Raton, FL: Taylor & Francis.

Hoffman, R.R. (2017b). Extending and integrating macrocognition models and concepts. Presented at the Thirteenth International Conference on Naturalistic Decision Making., University of Bath, England. Retrieved from

https://www.eventsforce.net/uob/media/uploaded/EVUOB/event_2/GoreWard_NDM13Pr oceedings_2017.pdf

Hoffman, R. R., Eskridge, T., & Shelley, C. (2009). A naturalistic exploration of forms and functions of analogizing. *Metaphor and Symbol, 24*(3), 125–154.

Hoffman, R. R., Johnson, M., Bradshaw, J. M., & Underbrink, A. (2013). Trust in automation. *IEEE Intelligent Systems, 28*(1), 84–88.

Hoffman, R.R., & Klein, G. (2017b). Explaining Explanation, Part 1: Theoretical Foundations. *IEEE Intelligent Systems, 32*(3), 68–73. https://doi.org/10.1109/MIS.2017.54

Hoffman, R.R., Klein, G. A., & Miller, J. (2011). Naturalistic investigations and models of reasoning about complex indeterminate causation. *Information Knowledge Systems Management*, *10*(1–4), 397–425.

Hoffman, R.R., Klein, G., & Miller, J.E. (2011). Naturalistic investigations and models of reasoning about complex indeterminate causation. *Information and Knowledge Systems Management, 10*, 397-425.

Hoffman, R.R., LaDue, D., & Mogil, H.M., Roebber, P., and Trafton, J.G. (2017). *Minding the Weather: How Expert Forecasters Think*. Cambridge, MA: MIT Press.

Hoffman, R.R., Miller, T. Klein, G., & Litman, J. (2018). "Metrics for Explainable AI: Challenges and Prospects." A report on the DARPA Explainable AI Program, DARPA, Washington, DC.

Hoffman, R., Miller, T., Mueller, S. T., Klein, G. & Clancey, W. J. (2018). Explaining Explanation, Part 4: A Deep Dive on Deep Nets. IEEE Intelligent Systems, 33(3), 87–95.

Hoffman, R.R., Mueller, S.T., & Klein, G. (2017). Explaining Explanation, Part 2: Empirical Foundations. *IEEE Intelligent Systems*, *32*(4), 78–86. https://doi.org/10.1109/MIS.2017.3121544

Hoffman, R.R., Mueller, S.T., Klein, G. & Litman, J. (2018). "Metrics for Explainable AI: Challenges and Prospects." Deliverable on the DARPA XAI Program.

Hoffman, R.R., Ward, P., DiBello, L., Feltovich, P.J., Fiore, S.M., & Andrews, D. (2014). *Accelerated Expertise: Training for High Proficiency in a Complex World*. Boca Raton, FL: Taylor and Francis/CRC Press.

Hoffmann, J., Kissmann, P., & Torralba, A. (2014). "Distance"? Who cares? Tailoring merge-and-shrink heuristics to detect unsolvability. In *Proceedings of the 2014 European Conference on Artificial Intelligence*. IOS Press. [DOI:10.3233/978-1-61499-419-0-441]

Hollan, J.D., Hutchins, E.L., & Weitzman, L. (1984). STEAMER: An interactive inspectable simulation-based training system. *The AI Magazine, 5* (2) 15-27.

Hospers, J. (1946). On explanation. *The Journal of Philosophy*, *43*(13), 337–356. https://doi.org/10.2307/2019810

Hu, R., Andreas, J., Rohrbach, M., Darrell, T., & Saenko, K. (2017). Learning to reason: End-to-end module networks for visual question answering. *ArXiv Preprint ArXiv:1704.05526*. Retrieved from https://arxiv.org/abs/1704.05526

Huber, D. E., Shiffrin, R. M., Lyle, K. B., & Ruys, K. I. (2001). Perception and preference in short-term word priming. *PSYCHOLOGICAL REVIEW-NEW YORK-*, *108*(1), 149–182.

Hume, D. (1748). *Philosophical Essays Concerning Human Understanding (1 ed.). London: A. Millar.*

Jackson, P. (1985). Reasoning about belief in the context of advice-giving systems. In M.A. Bramer (Ed.), *Research and development in expert systems* (pp. 73-83). Cambridge University Press.

Jain, A., Zamir, A. R., Savarese, S., & Saxena, A. (2016). Structural-RNN: Deep learning on spatio-temporal graphs. In *Proceedings of the IEEE Conference on Computer Vision and Pattern Recognition* (pp. 5308–5317). Retrieved from http://www.cv-foundation.org/openaccess/content_cvpr_2016/html/Jain_Structural-RNN_Deep_Learning_CVPR_2016_paper.html

Jakulin, A., Možina, M., Demšar, J., Bratko, I., & Zupan, B. (2005). Nomograms for visualizing support vector machines. In *Proceedings of the eleventh ACM SIGKDD international conference on Knowledge discovery in data mining* (pp. 108–117). ACM.

James, W. (2003). Scientific explanation. In *The Stanford Encyclopedia of Philosophy (Spring 2017 Edition)*.

Jamieson, G. A. (1996). Using the Conant method to discover and model human-machine structures. In *Proceedings of the Human Factors and Ergonomics Society Annual Meeting* (Vol. 40, pp. 173–176). SAGE Publications Sage CA: Los Angeles, CA. Retrieved from http://journals.sagepub.com/doi/abs/10.1177/154193129604000406

Jarvelin, K., & Kekalainen, J. (2002). "Cumulated gain-based indicators of IR performance." Research Note, Department of Information Sciences, University of Tampere. Tampere, Finland.

Jha, S., Raman, V., Pinto, A., Sahai, T., & Francis, M. (2017). On learning sparse Boolean formulae for explaining AI decisions. In *NASA Formal Methods Symposium* (pp. 99–114). Springer.

Jian, J.-Y., Bisantz, A.M., & Drury, C. G. (1998). Towards an empirically determined scale of trust in computerized systems: distinguishing concepts and types of trust. In *Proceedings of the Human Factors and Ergonomics Society Annual Meeting* (Vol. 42, pp. 501–505). SAGE Publications Sage CA: Los Angeles, CA.

Jian, J.-Y., Bisantz, A. M., & Drury, C. G. (2000). Foundations for an empirically determined scale of trust in automated systems. *International Journal of Cognitive Ergonomics*, *4*(1), 53–71.

Johnson, D.S. (2007). Achieving customer value from electronic channels through identity commitment, calculative commitment, and trust in technology. *Journal of Interactive Marketing*, *21*(4), 2–22.

Johnson, H., & Johnson, P. (1993). Explanation Facilities and Interactive Systems. In *Proceedings of the 1st International Conference on Intelligent User Interfaces* (pp. 159–166). New York, NY, USA: ACM. https://doi.org/10.1145/169891.169951

Johnson, M., Bradshaw, J. M., Hoffman, R. R., Feltovich, P. J., & Woods, D. D. (2014). Seven cardinal virtues of human-machine teamwork: examples from the DARPA robotic challenge. *IEEE Intelligent Systems*, *29*(6), 74–80.

Johnson, P., Johnson, H., Waddington, R., & Shouls, A. (1988). Task-related knowledge structures: analysis, modelling and application. In *BCS HCI* (pp. 35–62). Retrieved from https://pdfs.semanticscholar.org/803b/42e4f5a7a52a70e73fc141c7f87c7c8829f2.pdf

Johnson, W. L. (1994). Agents that learn to explain themselves. In *AAAI 1994 Proceedings* (pp. 1257–1263). Menlo Park, CA: Association for the Advancement of Artificial Intelligence.

Johnson, W.L., & Lewis, J.C. (2018, Summer). Pedagogical agents: Back to the future. *The AI Magazine*, 33 -44.

Johnson-Laird, P. N. (1980). Mental models in cognitive science. *Cognitive Science*, *4*(1), 71–115.

Johnson-Laird, P.N. (1983). *Mental models: Towards a cognitive science of language, inference, and consciousness*. Harvard University Press.

Jones, R.M., & VanLehn, K. (1992). A fine-grained model of skill acquisition: fitting cascade to individual subjects. In *Proceedings of the Fourteenth Annual Conference of the Cognitive Science Society* (pp. 873-877). Boca Raton, F: CRC Press.

Joseph, M., Kearns, M., Morgenstern, J., Neel, S., & Roth, A. (2016a). Rawlsian fairness for machine learning. *ArXiv Preprint ArXiv:1610.09559*.

Joseph, M., Kearns, M., Morgenstern, J., Neel, S., & Roth, A. (2016b). Fair algorithms for infinite and contextual bandits. *arXiv preprint arXiv:1610.09559*.

Khan, O.Z., Poupart, P. & Black, J.P. (2009). Minimal sufficient explanations for factored Markov decision processes. In *Proceedings of the Nineteenth International Conference on Automated Planning and Scheduling* (pp. 194-200). Menlo Park, CA: Association for the Advancement of Artificial Intelligence.

Kahneman, D., & Varey, C.A. (1990). Propensities and counterfactuals: The loser that almost won. *Journal of Personality and Social Psychology, 59,* 1101-1110.

Kaplan, A. (1964). *The Conduct of Inquiry: Methodology for Behavioral Science*. Chandler Publishing Company.

Karakul, M., & Qudrat-Ullah, H. (2008). How to improve dynamic decision making? Practice and promise. *Complex Decision Making*, 3–24.

Karpathy, A., Johnson, J., & Fei-Fei, L. (2015). Visualizing and understanding recurrent networks. *ArXiv Preprint ArXiv:1506.02078*. Retrieved from https://arxiv.org/abs/1506.02078

Karsenty, L., & Brezillon, P. J. (1995). Cooperative problem solving and explanation. *Expert Systems With Applications*, *8*(4), 445–462.

Kass, R., & Finin, T. (1988). The Need for User Models in Generating Expert System Explanation. *Int. J. Expert Syst.*, *1*(4), 345–375.

Kass, A., & Leake, D. (1987). "Types of explanations." Report. Department of Computer Science, Yale University, New Haven, CT. [http://www.dtic.mil/dtic/tr/fulltext/u2/a183253.pdf].

Katz, S., & Lesgold, A. (1992). Use of Fuzzy Modeling Techniques in a Coached Practice Environment for Electronics Troubleshooting. Retrieved from https://eric.ed.gov/?id=ED349964

Kay, J. (2000). Stereotypes, student models and scrutability. In G. Gauthier, C. Frasson, and K. VanLehn (eds.), *Lecture Notes in Computer Science 1839* (pp. 19-30). Berlin: Springer.

Kay, J. (2006). Scrutable adaptation: Because we can and must. In V. Wade, H. Ashman, and B. Smyth (eds.), *Lecture Notes in Computer Science 4018* (pp. 11 – 19). Berlin: Springer.

Kefalidou, G. (2017). When immediate interactive feedback boosts optimization problem solving: A 'human-in-the-loop' approach for solving Capacitated Vehicle Routing Problems. *Computers in Human Behavior*, *73*, 110–124. https://doi.org/10.1016/j.chb.2017.03.019

Kehoe, C., Stasko, J., & Taylor, A. (2001). Rethinking the evaluation of algorithm animations as learning aids: an observational study. *International Journal of Human-Computer Studies*, *54*, 265–284.

Keil, F.C. (2006). Explanation and understanding. *Annual Review of Psychology 57*, 227–254.

Keil, F., Rozenblit, L., & Mills, C.M. (2004). What lies beneath? Understanding the limits of understanding. In D.T. Levin (Ed.), *Thinking and seeing* (pp. 227-249). Cambridge: MIT Press.

Khan, O.Z., Poupart, P. & Black, J.P. (2009). Minimal sufficient explanations for factored Markov decision processes. In *Proceedings of the Nineteenth International Conference* *on Automated Planning and Scheduling* (pp. 194-200). Menlo Park, CA: Association for the Advancement of Artificial Intelligence.

Khemlani, S., & Johnson-Laird, P. N. (2010). Explanations make inconsistencies harder to detect. In *Proceedings of the 32 annual meeting of the cognitive science society*. Portland, Oregon.

Kieras, D. E., & Bovair, S. (1984). The role of a mental model in learning to operate a device. *Cognitive Science*, *8*(3), 255–273.

Kieras, D. & Polson, P.G. (1985). An approach to the formal analysis of user complexity. *International Journal of Man-Machine Studies, 22*, 365-394.

Kilgore, R., & Voshell, M. (2014). Increasing the transparency of unmanned systems: Applications of ecological interface design. In *International Conference on Virtual, Augmented and Mixed Reality* (pp. 378–389). Springer. Retrieved from http://link.springer.com/chapter/10.1007/978-3-319-07464-1_35

Kim, B., Chacha, C.M., & Shah, J. (2013). Inferring Robot Task Plans from Human Team Meetings: A Generative Modeling Approach with Logic-Based Prior. In *Proceedings of the Twenty-Seventh AAAI Conference on Artificial Intelligence* (pp. 1394-1400). Menlo Park, CA: Association for the Advancement of Artificial Intelligence.

Kim, B. Glassman, E., Johnson, B, & Shah, J. (2015). "iBCM: Interactive Bayesian Case Model Empowering Humans via Intuitive Interaction." Report MIT-CSAIL-TR-2015-010, Computer Science and Artificial Intelligence Laboratory, MIT, Cambridge, MA.

Kim, B., Khanna, R., & Koyejo, O. O. (2016). Examples are not enough, learn to criticize! criticism for interpretability. In *Advances in Neural Information Processing Systems* (pp. 2280–2288). Retrieved from http://papers.nips.cc/paper/6300-examples-are-not-enough-learn-to-criticize-criticism-for-interpretability

Kim, B., Lin, X., Collins, C., Taylor, G.W., & Amer, M.R. (2016). Learn, generate, rank, explain: A case study of explanation by generation. In *Proceedings of CHI'16*. New York: Association for Computing Machinery.

Kim, B., Shah, J., & Doshi-Velez, F. (2015). Mind the gap: A generative approach to interpretable feature selection and extraction. *Advances in Neural Information Processing Systems, 28*, 2251–2259.

Kim, C., Lin, X., Collins, C., Taylor, G.W., & Amer, M. (2016). Learn, generate, rank, explain: A case study of explanation by generation. In *Proceedings of CHI '16*. New York: Association for Computing Machinery.

Kim, M.-J., Kim, K.-J., Kim., S., & Dey, A.K. (2016). Evaluation of Starcraft artificial intelligence competition bots by experienced human players. In *Proceedings of the 2016 CHI Conference Extended Abstracts on Human Factors in Computing Systems* (pp. 1915-1921. New York: Association for Computing Machinery.

Kim, T. W. (2018). Explainable artificial intelligence (XAI), the goodness criteria and the graspability test. *arXiv preprint arXiv:1810.09598*.

Kisa, D., Van den Broeck, G., Choi, A., & Darwiche, A. (2014). Probabilistic Sentential Decision Diagrams. In *KR*. Retrieved from http://www.aaai.org/ocs/index.php/KR/KR14/paper/download/8005/7969

Klahr, D., & Dunbar, K. (1988). Dual space search during scientific reasoning. *Cognitive Science*, *12*(1), 1–48. https://doi.org/10.1207/s15516709cog1201_1

Klein, G.A. (1989). Recognition-primed decisions. In Rouse, W. B. (Ed.), *Advances in man machine system research* (Vol. 5, pp. 47–92). Greenwich, CT: JAI Press.

Klein, G.A. (1993). *A recognition-primed decision (RPD) model of rapid decision making*. Ablex Publishing Corporation New York. Retrieved from https://pdfs.semanticscholar.org/0672/092ecc507fb41d81e82d2986cf86c4bff14f.pdf

Klein, G.A. (2007). Flexecution as a paradigm for replanning, part 1. *IEEE Intelligent Systems*, *22*(5). Retrieved from http://ieeexplore.ieee.org/abstract/document/4338498/

Klein, G. & Crandall, B. (1995). The role of mental simulation in problem solving and decision making. In In P. Hancock, J. Flach, J. Caird and K. Vicente (Eds.), *Local applications of the ecological approach to human-machine systems* (pp. 324-358). Mahwah, NJ: Erlbaum.

Klein, G. A., Feltovich, P. J., Bradshaw, J. M., & Woods, D. D. (2005). Common ground and coordination in joint activity. *Organizational Simulation*, *53*, 139–184.

Klein, G.A., & Hoffman, R.R. (2008). Macrocognition, mental models, and cognitive task analysis methodology. In *Naturalistic Decision Making and Macrocognition* (pp. 57–80).

Klein, G., Moon, B., & Hoffman, R. R. (2006a). Making sense of sensemaking 1: Alternative perspectives. *IEEE Intelligent Systems*, *21*(4), 70–73.

Klein, G.A., Moon, B., & Hoffman, R.R. (2006b). Making sense of sensemaking 2: A macrocognitive model. *IEEE Intelligent Systems*, *21*(5), 88–92.

Klein, G.A., Orasanu, J., Calderwood, R., & Zsambok, C. E. (1993). *Decision Making in Action: Models and Methods*. Norwood, NJ: Ablex.

Klein, G.A., Phillips, J.K., Rall, E. L., & Peluso, D.A. (2007). A data-frame theory of sensemaking. In *Expertise out of context: Proceedings of the sixth international conference on naturalistic decision making* (pp. 113–155). New York, NY, USA: Lawrence Erlbaum.

Klein, G. A., Rasmussen, L., Lin, M.-H., Hoffman, R. R., & Case, J. (2014). Influencing preferences for different types of causal explanation of complex events. *Human Factors*, *56*(8), 1380–1400.

Kline, D. (2017). Where human intelligence outperforms AI. Retrieved December 12, 2017, from https://techcrunch.com/2017/09/30/where-human-intelligence-outperforms-ai/

Kment, B. (2014). *Modality and explanatory reasoning*. Oxford University Press.

Ko, A. J., Myers, B., & Aung, H. (2004). Six learning barriers in end-user programming systems. In *Proceedings of VL/HCC 2004: The IEEE Symposium on Visual Languages and Human-Centric Computing* (pp. 199-206). New York: IEEE.

Koehler, D. (1991). Explanation, imagination , and confidence in judgement. *Psychological Bulletin*, *110*(3), 499–519.

Kofod-Petersen, A., Cassens, J., & Aamodt, A. (2008). Explanatory capabilities in the creek knowledge-intensive case-based reasoner. *Frontiers in Artificial Intelligence and Applications*, *173*, 28.

Koh, P.W., & Liang, P. (2017). Understanding black-box predictions via influence functions. In *Proceedings of the 34th International Conference on Machine Learning* (pp. 1995-1894) [arXiv:1703.04730v2].

Kononenko, I., Štrumbelj, E., Bosnić, Z., Pevec, D., Kukar, M., & Robnik-Šikonja, M. (2013). Explanation and reliability of individual predictions. *Informatica*, *37*(1), 48.

Koustanai, A., Mas, A., Cavallo, V., & Delhomme, P. (2010). Familiarization with a Forward Collision Warning on driving simulator: cost and benefit on driver system interactions and trust.

Krizhevsky, A., Sutskever, I., & Hinton, G. (2012). ImageNet classification with deep convolutional neural networks. In *Proceedings of the 25th International Conference on Neural Information Processing Systems* (pp. 1097–1105). Curran Associates Inc.

Kroll, J. A., Barocas, S., Felten, E. W., Reidenberg, J. R., Robinson, D. G., & Yu, H. (2016). Accountable algorithms. *University of Pennsylvania Law Review, 165,* 633-705.

Krull, D. S., & Anderson, C. A. (1997). The process of explanation. *Current Directions in Psychological Science*, *6*(1), 1–5.

Kuang, C. (2017). Can A.I. be taught to explain itself? *The New York Times*. Retrieved from https://www.nytimes.com/2017/11/21/magazine/can-ai-be-taught-to-explain-itself.html

Kuhn, D. (2001). How do people know. *Psychological Science*, *12*(1), 1–8.

Kulesza, T., Stumpf, S., Burnett, M., Wong, W.-K., Riche, Y., Moore, T., … McIntosh, K. (2010). Explanatory debugging: Supporting end-user debugging of machine-learned programs. In *Proceedings of the IEEE Symposium on Visual Languages and Human-Centric Computing* (pp. 41-48) New York: IEEE. [doi: 10.1109/VLHCC.2010.15]

Kulesza, T., Burnett, M., Stumpf, S., Wong, W.-K., Das, S., Groce, A., … McIntosh, K. (2011). Where are my intelligent assistant's mistakes? A systematic testing approach. *End-User Development*, 171–186.

Kulesza, T., Burnett, M., Wong, W.-K., & Stumpf, S. (2015). Principles of explanatory debugging to personalize interactive machine learning. In *Proceedings of the 20th International Conference on Intelligent User Interfaces* (pp. 126–137). Atlanta, Georgia. https://doi.org/10.1145/2678025.2701399

Kulesza, T., Stumpf, S., Burnett, M., & Kwan, I. (2012). Tell me more?: The effects of mental model soundness on personalizing an intelligent agent. In *Proceedings of the SIGCHI Conference on Human Factors in Computing Systems* (pp. 1–10). ACM. Retrieved from http://dl.acm.org/citation.cfm?id=2207678

Kulesza, T., Stumpf, S., Burnett, M., Yang, S., Kwan, I., & Wong, W.-K. (2013). Too much, too little, or just right? Ways explanations impact end users' mental models. In *Proceedings of IEEE Symposium on Visual Languages and Human-Centric Computing* (pp. 3-10). New York: IEEE. [doi: 10.1109/VLHCC.2013.6645235].

Kulesza, T., Stumpf, S., Wong, W-K., Burnett, M., Perona, S., Ko, A. & Oberst, I. (2011). Why-oriented end-user debugging of naïve Bayes text classification. *Transactions on Interactive Intelligent Systems, 1* New York: Association for Computing Machinery. [doi: 10.1145/2030365.2030367]

Kulesza, T., Wong, W.-K., Burnett, M., & Stumpf, S. (2012b). The role of explanations in assessing and correcting personalized intelligent agents. In *Proceedings of CHI'12: Computer-Human Interaction*. New York: Association for Computing Machinery. [https://dropline.net/wp-content/uploads/2012/02/CHI-2012-Workshop.pdf]

Kumar, D., Wong, A., & Taylor, G.W. (2017). Explaining the unexplained: A Class-Enhanced Attentive Response (CLEAR) approach to understanding deep neural networks. In *Proceedings of the Computer Vision and Patter Recognition Workshop* (CVPR-W), New York: IEEE  [arXiv:1704.04133v2]

Lacave, C., & Díez, F.J. (2002). A review of explanation methods for Bayesian networks. *The Knowledge Engineering Review*, *17*(2), 107–127.

Lajoie, S.P. (2009). Developing professional expertise with a cognitive apprenticeship model: Examples from avionics and medicine. *Development of Professional Expertise: Toward Measurement of Expert Performance and Design of Optimal Learning Environments*, 61–83.

Lake, B.M., Ullman, T.D., Tennenbaum, J.B., & Gershman, S.J. (2017). Building machines that learn and think like people. *Behavioral and Brain Sciences, 40*, E253. [doi:10.1017/S0140525X16001837].

Lakkaraju, H., Kamar, E., Caruana, R., & Leskovec, J. (2017). Interpretable & Explorable Approximations of Black Box Models. *ArXiv Preprint ArXiv:1707.01154.*

Lamberti, D.M., & Walace, W.A. (1990). Intelligent interface design: An empirical assessment of knowledge presentation in expert systems. *MIS Quarterly, 14*, 279-311.

Landauer, T. K. (1988). Research methods in human-computer interaction. In *Handbook of human-computer interaction* (pp. 905–928). New York: Elsevier.

Landecker, W., Thomure, M. D., Bettencourt, L. M., Mitchell, M., Kenyon, G. T., & Brumby, S. P. (2013). Interpreting individual classifications of hierarchical networks. In *Proceedings of the IEEE Symposium on Computational Intelligence and Data Mining* (CIDM) New York: IEEE. [DOI: 10.1109/CIDM.2013.6597214].

Lane, H.C., Core, M. G., Van Lent, M., Solomon, S., & Gomboc, D. (2005). *Explainable artificial intelligence for training and tutoring*. Institute For Creative Technologies, University Of Southern California, Marina Del Rey Ca. Retrieved from http://www.dtic.mil/docs/citations/ADA459148

Langley, P. (1987). *Scientific discovery: Computational explorations of the creative processes*. MIT press. Retrieved from https://books.google.com/books?hl=en&lr=&id=VPxc-uPB7LEC&oi=fnd&pg=PP11&dq=Langley,+P.,+Simon,+H.A.,+Bradshaw,+G.L.,+and+Zytkow,+J.M.+(1987).+Scientific+discovery:+Computational+explorations+of+the+creative+process.+Cambridge,+MA:+MIT+Press.+&ots=LZY-AwrN9k&sig=UU0X_KwwGeBz4jqCOKTCym6QmwE

Langley, P., Meadows, B., Sridharan, M., & Choi, D. (2017). Explainable Agency for Intelligent Autonomous Systems. In *AAAI* (pp. 4762–4764). Retrieved from http://www.aaai.org/ocs/index.php/IAAI/IAAI17/paper/download/15046/13734

Langlotz, C. P., & Shortliffe, E. H. (1989). The critiquing approach to automated advice and explanation: rationale and examples. In *Expert Knowledge an Explanation: The Knowledge-Language Interface, Charlie Ellis (Ed)*. New York: Ellis Horwood Limited.

Lazar, J., Feng, J. H., & Hochheiser, H. (2017). *Research methods in human-computer interaction*. Morgan Kaufmann.

Lazaridou, A., Peysakhovich, A., & Baroni, M. (2017). Multi-agent cooperation and the emergence of (natural) language. In *Proceedings of the International Conference on Learning Representations*, ICLR 2017. [arXiv:1612.07182v2].

LeCun, Y., Bengio, Y., and Hinton, G. (2015). Deep learning. *Nature, 512,* 436-444.

Leddo, J., Abelson, R.P., & Gross, P.H. (1984). Conjunctive explanations: When two reasons are better than one. *Journal of Personality and Social Psychology, 47,* 933-943.

Lee, J.D., & See, K.A. (2004). Trust in automation: Designing for appropriate reliance. *Human Factors, 46,* 50-80.

Lee, J.D., & Moray, N. (1994) Trust, self-confidence, and operators' adaptation to automation. *International Journal of Human-Computer Studies, 40,* 153-184.

Lei, T., Barzilay, R., & Jaakkola, T. (2016). Rationalizing neural predictions. In *Proceedings of the 2016 Conference on Empirical Methods in Natural Language Processing* (pp. 107-117). Stroudsburgh,PA: Association for Computational Linguistics [arXiv:1606.04155v2].

Lerch, F. J., Prietula, M. J., & Kulik, C. T. (1997). The Turing effect: The nature of trust in expert systems advice. In *Expertise in context* (pp. 417–448). MIT Press.

Lerner, J., & Tetlock, P. (1999). Accounting for the effects of accountability. *Psychological Bulletin*, *125*(2), 255–275. https://doi.org/10.1037/0033-2909.125.2.255

Lesgold, A., Lajoie, S., Bunzo, M., & Eggan, G. (1988). *SHERLOCK: A coached practice environment for an electronics troubleshooting job*. Report, Learning Research and Development Center, University of Pittsburgh, Pittsburgh, PA. Retrieved from https://eric.ed.gov/?id=ED299450

Letham, B., Rudin, C., McCormick, T. H., Madigan, D., & others. (2015). Interpretable classifiers using rules and Bayesian analysis: Building a better stroke prediction model. *The Annals of Applied Statistics*, *9*(3), 1350–1371.

Lewis, C. (1986). A model of mental model construction. *ACM SIGCHI Bulletin*, *17*(4), 306–313.

Li, J., Chen, X., Hovy, E., & Jurafsky, D. (2015). Visualizing and understanding neural models in NLP. In *Proceedings of The 15th Annual Conference of the North American Chapter of* *the Association for Computational Linguistics* (pp. 681–691). Stroudsburgh, PA: Association for Computational Linguistics. [arXiv:1506.01066v2].

Li, Y., Yosinski, J., Clune, J., Lipson, H., & Hopcroft, J. (2015). Convergent learning: Do different neural networks learn the same representations? *Journal of Machine Learning, 44*, 196-212.

Liang, Y., & Van den Broeck, G. (2017). Towards Compact Interpretable Models: Shrinking of Learned Probabilistic Sentential Decision Diagrams. *IJCAI-17 Workshop on Explainable Artificial Intelligence (XAI).* Retrieved from http://home.earthlink.net/~dwaha/research/meetings/ijcai17-xai/5.%20(Liang%20&%20Van%20den%20Broeck%20XAI-17)%20Towards%20Compact%20Interpretable%20Models.pdf

Lim, B.Y., & Dey, A.K. (2009). Assessing demand for intelligibility in context-aware applications. In P*roceedings of the 11th International Conference on Ubiquitous Computing* (pp. 195-204). New York: Association for Computing Machinery.

Lim, B.Y., & Dey, A.K. (2010). Toolkit to support intelligibility in context-aware applications. In P*roceedings of the 12th International Conference on Ubiquitous Computing* (pp. 13-22). New York: Association for Computing Machinery.

Lim, B.Y., Dey, A.K., & Avrahami, D. (2009). Why and Why Not explanations improve the intelligibility of context-aware intelligent systems. In *Proceedings of the SIGCHI Conference on Human Factors in Computing System*s (pp. 2119-2128). New York: Association for Computing Machinery.

Linegang, M. P., Stoner, H. A., Patterson, M. J., Seppelt, B. D., Hoffman, J. D., Crittendon, Z. B., & Lee, J. D. (2006). Human-automation collaboration in dynamic mission planning: A challenge requiring an ecological approach. In *Proceedings of the Human Factors and Ergonomics Society Annual Meeting* (Vol. 50, pp. 2482–2486). SAGE Publications Sage CA: Los Angeles, CA.

Lipton, P. (1990). Contrastive explanation. *Royal Institute of Philosophy Supplements*, *27*, 247–266.

Lipton, Z. C. (2016). The mythos of model interpretability. Presented at the 2016 ICML Workshop on Human Interpretability in Machine Learning.

Litman, J. A. and Silvia, P. J. (2006). The latent structure of trait curiosity: Evidence for interest and deprivation curiosity dimensions. *Journal of Personality Assessment, 86*, 318-328.

Liu, M., Shi, J., Li, Z., Li, C., Zhu, J., & Liu, S. (2017). Towards better analysis of deep convolutional neural networks. *IEEE Transactions on Visualization and Computer Graphics*, *23*(1), 91–100.

Llinas, J., Bisantz, A., Drury, C., Seong, Y., & Jian, J.-Y. (1998). *Studies and analyses of aided adversarial decision making. Phase 2: Research on human trust in automation*. DTIC Document.

Lobato, J., & Siebert, D. (2002). Quantitative reasoning in a reconceived view of transfer. *Journal of Mathematical Behavior, 21*, 87-116.

Loksa, D., Ko, A. J., Jernigan, W., Oleson, A., Mendez, C. J., & Burnett, M. M. (2016). Programming, Problem Solving, and Self-Awareness: Effects of Explicit Guidance. In *Proceedings of the 2016 CHI Conference on Human Factors in Computing Systems* (pp. 1449–1461). ACM. Retrieved from http://dl.acm.org/citation.cfm?id=2858252

Lomas, M., Chevalier, R., Cross II, E. V., Garrett, R. C., Hoare, J., & Kopack, M. (2012). Explaining robot actions. In *Proceedings of the seventh annual ACM/IEEE international conference on Human-Robot Interaction* (pp. 187–188). ACM. Retrieved from http://dl.acm.org/citation.cfm?id=2157748

Lombrozo, T. (2006). The structure and function of explanations. *Trends in Cognitive Sciences*, *10*(10), 464–470. https://doi.org/10.1016/j.tics.2006.08.004

Lombrozo, T. (2007). Simplicity and probability in causal explanation. *Cognitive Psychology*, *55*, 232–257. https://doi.org/10.1016/j.cogpsych.2006.09.006

Lombrozo, T. (2009a). Explanation and categorization: How "why?" informs "what?" *Cognition*, *110*(2), 248–253. https://doi.org/10.1016/j.cognition.2008.10.007

Lombrozo, T. (2009b). Explanation and categorization: How "why?" informs "what?" *Cognition*, *110*(2), 248–253.

Lombrozo, T. (2010). Causal–explanatory pluralism: How intentions, functions, and mechanisms influence causal ascriptions. *Cognitive Psychology*, *61*(4), 303–332.

Lombrozo, T. (2011). The instrumental value of explanations. *Philosophy Compass*, *6*(8), 539–551.

Lombrozo, T. (2012). Explanation and abductive inference. In *Oxford handbook of thinking and reasoning* (pp. 260–276).

Lombrozo, T. (2016). Explanatory preferences shape learning and inference. *Trends in Cognitive Sciences*, *20*(10), 748–759. https://doi.org/10.1016/j.tics.2016.08.001

Lombrozo, T. (2016). Explanation. In *A Companion to Experimental Philosophy* (1st ed., pp. 491–503),

Lombrozo, T., & Carey, S. (2006). Functional explanation and the function of explanation. *Cognition*, *99*, 167–204. https://doi.org/10.1016/j.cognition.2004.12.009

Lombrozo, T., & Gwynne, N. Z. (2014a). Explanation and inference: mechanistic and functional explanations guide property generalization. *Frontiers in Human Neuroscience*, *8*. Retrieved from https://www.ncbi.nlm.nih.gov/pmc/articles/PMC4160962/

Lombrozo, T., & Gwynne, N. Z. (2014b). Explanation and inference: mechanistic and functional explanations guide property generalization. *Frontiers in Human Neuroscience*, *8*, 700.

Lombrozo, T., & Rutstein, J. (2004). Simplicity in explanation. *Cognitive Science Society*.

Lombrozo, T., Sloman, S., Strevens, M., Trout, J.D., & Skolnick Weisberg, D. (2008). Understanding why: The cognitive science of explanation. *Proceedings of the Annual Meeting of the Cognitive Science Society, 30*, 693-694.

Lou, Y., Caruana, R., & Ghrke, J. (2012). Intelligible models for classification and regression. In Proceedings of *Knowledge Discovery and Data Mining* KDD'12 (pp. 150-158). New York: Association for Computing Machinery. [doi 10.1145/2339530.2339556].

Lulia Nastasia, D., & Rakow, L. (2010). What is theory? Puzzles and maps as metaphors in communication theory. *TripleC*, *B*(1), 1–17.

Lum, K., and Isaac, W. (2016, October). To predict and serve? *Significance Magazine* (pp. 14-19).

Lyons, J.B. (2013). Being transparent about transparency: A model for human-robot inter-action. In D. Sofge, G.J. Kruijff, & W.F. Lawless (Eds.) *Trust and Autonomous Systems: Papers from the AAAI Spring Symposium* (Technical Report SS-13-07). Menlo Park, CA: AAAI

Lyons, J.B., Clark, M.A., Wagner, A.R., & Schuelke, M.J. (2017). Certifiable trust in autonomous systems: Making the intractable tangible. *AI Magazine*, *38*(3), 37–49. https://doi.org/10.1609/aimag.v38i3.2717

Lyons, J.T. Ho, N., Lee Van Abel, A., Hoffmann, L., et al. (2017). Comparing trust in Auto-GCAS between experienced and novice Air Force pilots. *Ergonomics in Design*, *25*, 4–9. https://doi.org/10.1177/1064804617716612

van der Maaten, L., & Hinton, G. (2008). Visualizing data using t-SNE. *Journal of Machine Learning Research, 9,* 2579-2605.

Machamer, P., Darden, L., & Craver, C.F. (2000). thinking about mechanisms. *Philosophy of Science*, *67*, 1-25.

Mahendran, A., & Vedaldi, A. (2015). Understanding deep image representations by inverting them. (n proceedings of Computer vision (CVPR 2015), New York: IEEE. [https://www.cv-foundation.org/openaccess/CVPR2015_search.py]

Maheswaran, D., & Chaiken, S. (1991). Promoting systematic processing in low-motivation settings: Effect of incongruent information on processing and judgment. *Journal of Personality and Social Psychology*, *61*(1), 13.

Maida, A.S., and Deng, M. (1989). A language to allow expert systems to have beliefs about their users. In C. Ellis (Ed.), *Expert knowledge and explanation* (pp. 127-143). New York: John Wiley.

Malle, B.F. (1999). How people explain behavior: A new theoretical framework. *Personality and Social Psychology Review, 3*, 23-48.

Malle, B. F. (2004). H*ow the mind explains behavior.* Cambridge, MA: MIT Press.

Malle, B.F. (2011). Time to give up the dogmas of attribution: An alternative theory of behavior explanation. In J. M. Olson & M. P. Zanna (Eds.), *Advances in experimental social psychology. Advances in experimental social psychology, 44*, 297-352. San Diego, CA, US: Academic Press.

Mandel, D.R. (2003a). Effect of counterfactual and factual thinking on causal judgments. *Thinking & Reasoning, 9*, 246-265.

Mandel, D.R. (2003b). Judgment dissociation theory: An analysis of differences in causal, counterfactual, and covariational reasoning. *Journal of Experimental Psychology: General, 132*, 419-434.

Mandel, D.R. (2011). Mental simulation and the nexus of causal and counterfactual explanation. In C. Hoerl, T. McCormack, and S.R. Beck (Eds.), *Understanding counterfactuals, understanding causation: Issues in philosophy and psychology* (pp. 147-170). Oxford: Oxford University Press.

Mandel, D.R., Hilton, D.J., & Catellani P. (Eds.) (2011). *The psychology of counterfactual thinking* Bristol, UK: Routledge.

Mao, J., Benbasat, I., & Dhaliwal, J. S. (1996). Enhancing explanations in knowledge-based systems with hypertext. *Journal of Organizational Computing and Electronic Commerce*, *6*(3), 239–268. https://doi.org/10.1080/10919399609540279

Manicas, P., & Secord, P. (1983). Implications for psychology of the new philosophy of science. *American Psychologist*, 399–926.

Marcus, G. (2017). "Deep learning: A critical appraisal." [arXiv:1801.00631v1].

Mark, W.S., & Simpson, R.L. (1991, June). Knowledge-based systems. *IEEE Expert*, pp. 12-17.

Marr, D. & Poggio, T. (1976). "From understanding computation to understanding neural circuitry." MIT AI Memo 357, Artificial Intelligence Laboratory, MIT, Cambridge, MA.

Martens, D., Huysmans, J., Setiono, R., Vanthienen, J., & Baesens, B. (2008). Rule extraction from support vector machines: An overview of issues and application in credit scoring. S*tudies in Computational Intelligence (SCI) 80*, 33–63. Berlin: Springer-Verlag.

Martens, D., & Provost, F. (2014). Explaining data-driven document classification. *MIS Quarterly, 398*, A1- A6.

Martens, S., et al. (2008). Rule extraction from support vector machines: An overview of issues and application in credit scoring. S*tudies in Computational Intelligence (SCI) 80*, 33–63. Berlin: Springer-Verlag.

Martin, J.L., & Wiley, J. A. (2000). Algebraic representations of beliefs and attitudes II: Microbelief models for dichotomous belief data. *Sociological Methodology*, *30*, 123–164.

Maxwell, G. (1974). Corroboration without demarcation. In *The philosophy of Karl Popper* (Vol. 14, pp. 292–321). Open Court Publishing Co.

May, J., Barnard, P. J., & Blandford, A. (1993). Using structural descriptions of interfaces to automate the modelling of user cognition. *User Modeling and User-Adapted Interaction*, *3*(1), 27-64.

Maze, J. R. (1954). Do intervening variables intervene. *The Psychological Review*, *61*(4), 226–234.

McBride, S. E., Rogers, W. A., & Fisk, A. D. (2014). Understanding human management of automation errors. *Theoretical Issues in Ergonomics Science*, *15*(6), 545–577.

McCloy, R., & Byrne, R. M. (2000). Counterfactual thinking about controllable events. *Memory & Cognition*, *28*(6), 1071–1078.

McClure, J. (2002). Goal-based explanations of actions and outcomes. *European Review of Social Psychology, 12*, 201-235.

McClure, J., & Hilton, D. (1997). For you can't always get what you want: When preconditions are better explanations than goals. *British Journal of Social Psychology, 36*,223-240.

McClure, J., & Hilton, D. (1998). Are goals or preconditions better explanations? It depends on the question. *European Journal of Social Psychology, 28*, 897-911.

McClure, J. L., Sutton, R. M., & Hilton, D. J. (2003). The Role of Goal-Based Explanations. *Social Judgments: Implicit and Explicit Processes*, *5*.

McGill, A.L. (1991). The influence of the causal background on the selection of causal explanations. *British Journal of Social Psychology*, *30*(1), 79–87.

McGuinness, D.l. & Borgida, A. (1999). Explaining description subsumption logics. *Proceedings of IJCAI, International Joint Conference on Artificial Intelligence*. [https://www.semanticscholar.org].

McKeown, K.R., & Swartout, W. R. (1987). Language generation and explanation. *Annual Review of Computer Science*, *2*(1), 401–449.

Meissner, W.W. (1960). Intervening constructs: Dimensions of controversy, *67*(1), 51–72. https://doi.org/10.1037/h0042855

Mercado, J. E., Rupp, M. A., Chen, J. Y., Barnes, M. J., Barber, D., & Procci, K. (2016). Intelligent agent transparency in human–agent teaming for Multi-UxV management. *Human Factors*, *58*(3), 401–415.

Merrill, D. (1987). An expert system for instructional design. *Journal of Computer-Based Instruction*, *16*(3), 25–37.

Merritt, S.M. (2011). Affective processes in human–automation interactions. *Human Factors: The Journal of the Human Factors and Ergonomics Society*, *53*(4), 356–370.

Merritt, S.M., Heimbaugh, H., LaChapell, J., & Lee, D. (2013). I trust it, but I don't know why effects of implicit attitudes toward automation on trust in an automated system. *Human Factors: The Journal of the Human Factors and Ergonomics Society*, *55*(3), 520–534.

Metoyer, R., Stumpf, S., Neumann, C., Dodge, J., Cao, J., & Schnabel, A. (2010). Explaining how to play real-time strategy games. *Knowledge-Based Systems*, *23*(4), 295–301.

Miller, J.R. (1988). The role of human-computer interaction in intelligent tutoring systems. *Foundations of Intelligent Tutoring Systems*, 143–189.

Miller, T. (2017). Explanation in Artificial Intelligence: Insights from the Social Sciences. *ArXiv:1706.07269 [Cs]*. Retrieved from http://arxiv.org/abs/1706.07269

Miller, T., Howe, P., & Sonenberg, L. (2017). Explainable AI: Beware of Inmates Running the Asylum. *IJCAI-17 Workshop on Explainable Artificial Intelligence (XAI)*. Retrieved from http://home.earthlink.net/~dwaha/research/meetings/ijcai17-xai/6.%20(Miller+%20XAI-17%20extended)%20Explainable%20AI%20-%20Beware%20of%20Inmates%20Running%20the%20Asylum.pdf

Mills, C. M., & Keil, F. C. (2004). Knowing the limits of one's understanding: The development of an awareness of an illusion of explanatory depth. *Journal of Experimental Child Psychology*, *87*(1), 1–32.

Mitchell, D., Russo, E., & Rennington, N. (1989). Back to the future: Temporal perspective in the explanation of events. *Journal of Behavioral Decision Making*, *2*, 25–38. https://doi.org/10.1002/bdm.3960020103

Mnih, V., et al. (2013). Playing Atari with deep reinforcement learning. In *Proceedings of the NIPS Deep Learning Workshop*. [arXiv:1312.5602v1].

Mnih, V., Kavukcuoglu, K., Silver, D., Rusu, A. A., Veness, J., Bellemare, M. G., … others. (2015). Human-level control through deep reinforcement learning. *Nature*, *518*(7540), 529–533

Moeyersoms, J., d'Alessandro, B., Provost, F., & Martens, D. (2016). Explaining Classification Models Built on High-Dimensional Sparse Data. *ArXiv:1607.06280 [Cs, Stat]*. Retrieved from http://arxiv.org/abs/1607.06280

Molnar, C. (2018). *Interpretable machine learning*. [https://christophm.github.io/interpretable-ml-book/].

Montague, E. (2010). Validation of a trust in medical technology instrument. *Applied Ergonomics*, *41*(6), 812–821.

Monroe, D. (2018). AI, Explain Yourself. *Communications of the ACM, 61* (11), 11-13.

Montavon, G., Lapuschkin, S., Binder, A., Samek, W., & Müller, K.-R. (2017). Explaining nonlinear classification decisions with deep Taylor decomposition. *Pattern recognition*, *65*, 211-222.

Moon, B.M., Hoffman, R.R., Cañas, A.J., and Novak, J.D. (Eds.) (2011). *Applied concept mapping: Capturing, Analyzing and Organizing Knowledge*. Boca Raton, FL: Taylor and Francis.

Moore, D.T., & Hoffman, R.R. (2011). "A practice of understanding." Chapter 5 In D. T. Moore (Au.), *Sensemaking: A structure for an intelligence revolution*. Washington DC: National Defense Intelligence College Press.

Moore, J. D., & Swartout, W. R. (1988). *Explanation in expert systems: A survey*. University of Southern California Marina Del Rey Information Sciences Institute. Retrieved from http://www.dtic.mil/docs/citations/ADA206283

Moore, J. D., & Swartout, W. R. (1990). Pointing: A Way Toward Explanation Dialogue. In *AAAI* (Vol. 90, pp. 457–464). Retrieved from https://ocs.aaai.org/Papers/AAAI/1990/AAAI90-069.pdf

Moore, J. D., & Swartout, W. R. (1991). A reactive approach to explanation: Taking the user's feedback into account. In *Natural language generation in artificial intelligence and* *computational linguistics* (pp. 3–48). Springer. Retrieved from http://link.springer.com/chapter/10.1007/978-1-4757-5945-7_1

Moray, N. (1987a). Intelligent aids, mental models, and the theory of machines. *International Journal of Man-Machine Studies*, *27*(5–6), 619–629.

Moray, N. (1987b). Intelligent aids, mental models, and the theory of machines. *International Journal of Man-Machine Studies*, *27*(5–6), 619–629.

Moray, N. (1990). A lattice theory approach to the structure of mental models. *Philosophical Transactions of the Royal Society of London B: Biological Sciences*, *327*(1241), 577–583.

Moray, N. (1996). A taxonomy and theory of mental models. In *Proceedings of the Human Factors and Ergonomics Society Annual Meeting* (Vol. 40, pp. 164–168). SAGE Publications Sage CA: Los Angeles, CA. Retrieved from http://journals.sagepub.com/doi/abs/10.1177/154193129604000404

Moray, N. (1998). Identifying mental models of complex human–machine systems. *International Journal of Industrial Ergonomics*, *22*(4), 293–297.

Moray, N., & Inagaki, T. (1999). Laboratory studies of trust between humans and machines in automated systems. *Transactions of the Institute of Measurement and Control*, *21*, 203–211.

Moulin, B., Irandoust, H., Bélanger, M., & Desbordes, G. (2002). Explanation and Argumentation Capabilities: Towards the Creation of More Persuasive Agents. *Artificial Intelligence Review*, *17*(3), 169–222. https://doi.org/10.1023/A:1015023512975

Moulton, S. T., & Kosslyn, S. M. (2009). Imagining predictions: mental imagery as mental emulation. *Philosophical Transactions of the Royal Society of London B: Biological Sciences*, *364*(1521), 1273-1280.

Mueller, S.T. (2009). A Bayesian recognitional decision model. *Journal of Cognitive Engineering and Decision Making*, *3*(2), 111–130.

Mueller, S.T., & Klein, G.A. (2011). Improving Users' Mental Models of Intelligent Software Tools. *IEEE Intelligent Systems*, *26*(2), 77–83. https://doi.org/10.1109/MIS.2011.32

Mueller, S.T., & Shiffrin, R.M. (2006). REM II: A model of the developmental co-evolution of episodic memory and semantic knowledge. In *International Conference on Learning and Development (ICDL), Bloomington, IN*. Retrieved from https://pdfs.semanticscholar.org/ce57/291e001741c1ca5a1f2d1f92a290dddde51.pdf

Mueller, S. T., & Tan, Y. Y. S. (2018). Cognitive perspectives on opinion dynamics: the role of knowledge in consensus formation, opinion divergence, and group polarization. *Journal of Computational Social Science, 1*(1), 15-48.

Mueller, S.T., & Thanasuan, K. (2014). Associations and manipulations in the mental lexicon: a model of word-stem completion. *Journal of Mathematical Psychology*, *59*, 30–40.

Muir, B. M. (1994). Trust in automation: Part I. Theoretical issues in the study of trust and human intervention in automated systems. *Ergonomics*, *37*(11), 1905–1922. https://doi.org/10.1080/00140139408964957

Muir, B.M., & Moray, N. (1996). Trust in automation. Part II. Experimental studies of trust and human intervention in a process control simulation. *Ergonomics*, *39*(3), 429–460.

Muramatsu, J., & Pratt, W. (2001). Transparent Queries: investigation users' mental models of search engines. In *Proceedings of the 24th annual international ACM SIGIR conference on Research and development in information retrieval* (pp. 217–224). ACM. Retrieved from http://dl.acm.org/citation.cfm?id=383991

Murphy, D.S. (1990). Expert system use and the development of expertise in auditing: A preliminary investigation. *Journal of Information Systems, 4,* 18-35.

Murphy, G., & Medin, D. (1985). The role of theories in conceptual coherence. *Psychological Review*, *92*(3), 289–316.

Narayanan, M., Chen, E., He, J., Kim, S., Gershman, S., & Doshi-Velez, F. (2018). How do humans understand explanations from machine learning systems? An Evaluation of human-interpretability of explanation. [ArXiv:1802.00682].

Nataksu, R.T. (2004). Explanatory power of intelligent systems: A research framework. In *Proceedings of Decision Support in an Uncertain and Complex World: The IFIP TC8/WG8.3 International Conference 2004*. [https://pdfs.semanticscholar.org/8cff/ab3d8abb2f0c6014b1121c5ac77e1195d933.pdf].

Neches, R., Swartout, W. R., & Moore, J. D. (1985a). Enhanced maintenance and explanation of expert systems through explicit models of their development. *IEEE Transactions on Software Engineering*, (11), 1337–1351.

Neches, R., Swartout, W. R., & Moore, J. D. (1985b). Explainable (and Maintainable) Expert Systems. In *IJCAI* (Vol. 85, p. 382). Retrieved from http://www.academia.edu/download/30756822/10.1.1.81.9841.pdf

Niehaus, J., & Young, R. M. (2014). Cognitive models of discourse comprehension for narrative generation. *Literary and Linguistic Computing*, *29*(4), 561–582. https://doi.org/10.1093/llc/fqu056

Nott, G. (2017). 'Explainable Artificial Intelligence': Cracking open the black box of AI. Retrieved from https://www.computerworld.com.au/article/617359/explainable-artificial-intelligence-cracking-open-black-box-ai/

Novak, J. D. (1998). *Learning, creating, and using knowledge: Concept maps as facilitative tools in schools and corporations*. Mahwah, NJ: Lawrence Erlbaum Associates.

Nguyen, A., Yosinski, J., & Clune, J. (2015). Deep neural networks are easily fooled: High confidence predictions for unrecognizable images. In *Proceedings of the IEEE Conference on Computer Vision and Pattern Recognition* (pp. 427–436). Retrieved from http://www.cv-foundation.org/openaccess/content_cvpr_2015/html/Nguyen_Deep_Neural_Networks_2015_CVPR_paper.html

Nugent, C., Doyle, D., & Cunningham, P. (2009). Gaining insight through case-based explanation. *Journal of Intelligent Information Systems 32*, 267-295.

O'Neill, E., & Johnson, P. (1999). *Task knowledge structures and the design of collaborative systems*. AAAI Technical Report FS-99-03. Compilation copyright\copyright 1999, AAAI. Retrieved from http://www.aaai.org/Papers/Symposia/Fall/1999/FS-99-03/FS99-03-012.pdf

O'Reilly, T., Symons, S., & MacLatchy-Gaudet, H. (1998). A comparison of self-explanation and elaborative interrogation. *Contemporary Educational Psychology*, *23*(4), 434–445.

Otte, C. (2013). Safe and interpretable machine learning: A methodological review. In C. Moewes and A. Nürnberger (Eds.), *Computational intelligence in intelligent data analysis* (pp. 111-122). New York: Springer.

Overton, J. A. (2013). "Explain" in scientific discourse. *Synthese*, *190*(8), 1383–1405. https://doi.org/10.1007/s11229-012-0109-8

Pace, G.l., & Rosner, M. (2014). Explaining violation traces with finite state natural language generation models. In *Proceedings of the International Workshop on Controlled Natural Language CNL 2014: Controlled Natural Language* (pp 179-189). New York: Springer. [arXiv:1406.2298v1].

Pacer, M., Williams, J. J., Chen, X., Lombrozo, T., & Griffiths, T. (2013). Evaluating computational models of explanation using human judgments. In *Proceedings of the Twenty-Ninth Conference on Uncertainty in Artificial Intelligence*. Retrieved from https://arxiv.org/abs/1309.6855

Papamichail, K. N., & French, S. (2003). Explaining and justifying the advice of a decision support system: a natural language generation approach. *Expert Systems with Applications*, *24*(1), 35–48.

Papadimitriou, A., Symeonidis, P., & Manolopoulos, Y. (2016). A generalized taxonomy of explanations styles for traditional and social recommender systems. *Data Mining and Knowledge Discovery*. New York: Springer. [DOI 10.1007/s10618-011-0215-0]

Papernot, N., McDaniel, P., Goodfellow, I., Jha, S., Celik, B., & Swami, A. (2017). Practical black-box attacks against machine learning. In *ASIA CCS '17* (pp. 506–519). Abu Dhabi, UAE. https://doi.org/10.1145/3052973.3053009

Parasuraman, R., & Riley, V. (1997) Humans and automation: Use, misuse, disuse, abuse. *Human Factors, 39*, 230-253.

Parasuraman, R., Molloy, R., & Singh, I. L. (1993). Performance consequences of automation-induced 'complacency'. *The International Journal of Aviation Psychology*, *3*(1), 1–23.

Park, D.H., Hendricks, L. A., Akata, Z., Schiele, B., Darrell, T., & Rohrbach, M. (2016). *Attentive explanations: Justifying decisions and pointing to the evidence*. Retrieved from https://arxiv.org/abs/1612.04757

Pavlus, J. (2017, September 6). Stop pretending you really know what AI is and read this instead. Retrieved September 6, 2017, from https://qz.com/1067123/stop-pretending-you-really-know-what-ai-is-and-read-this-instead/

Pearl, J. (1988). *Probabilistic reasoning in intelligent systems: Networks of plausible inference*. San Mateo, CA: Morgan Kaufmann.

Pearl, J., & Mackenzie, D. (2018). *The book of why: The new science of cause and effect*. New York: Basic Books.

Peirce, C.S. (1891). Review of William James's *Principles of Psychology*, *Nation, 53*, 32.

Peirce, C.S. (1903). *Harvard Lectures on Pragmatism*, Ch. 5 (pp. 171–174). Cambridge, MA: Harvard University Press.

Phillips, E., Ososky, S., Grove, J., & Jentsch, F. (2011). From tools to teammates: Toward the development of appropriate mental models for intelligent robots. In *Proceedings of the Human Factors and Ergonomics Society Annual Meeting* (Vol. 55, pp. 1491–1495). SAGE Publications Sage CA: Los Angeles, CA. Retrieved from http://journals.sagepub.com/doi/abs/10.1177/1071181311551310

Pinker, S. (2017). Uncommon insights into common knowledge. In *APS Award Address* (Vol. 30).

Pirolli, P., & Card, S. (1999). Information foraging. *Psychological Review*, *106*(4), 643–675. https://doi.org/10.1037/0033-295X.106.4.643

Pirolli, P., & Card, S. (2005). The sensemaking process and leverage points for analyst technology as identified through cognitive task analysis. In *Proceedings of international conference on intelligence analysis* (pp. 2–4).

Polson, M.C., & Richardson, J.J. (Eds.). (1988). *Foundations of intelligent tutoring systems*. Hillsdale, NJ: Lawrence Erlbaum Associates.

Pomranky, R.A., Dzindolet, M.T., & Peterson, S.A. (2001). *Violations of expectations: A study of automation use*. Citeseer.

Poursabzi-Sangdeh, D.G. Goldstein, D.G., Hofman, J.M., Vaughn, J.W., & Wallace, H. (2018). Manipulating and measuring model interpretability. [arXiv: 1802.07810]

Preston, J., & Epley, N. (2005). Explanations versus applications: The explanatory power of valuable beliefs. *Psychological Science*, *16*(10), 826–832.

Prietula, M., Feltovich, P., & Marchak, F. (2000). Factors influencing analysis of complex cognitive tasks: A framework and example from industrial process control. *Human Factors: The Journal of the Human Factors and Ergonomics Society*, *42*(1), 54–74. https://doi.org/10.1518/001872000779656589

Psotka, J., Massey, L.D., & Mutter, S.A. (1988). *Intelligent tutoring systems: Lessons learned*. Psychology Press. Retrieved from

https://books.google.com/books?hl=en&lr=&id=sFtYUgHeEqUC&oi=fnd&pg=PR13&dq=Psotka,+J.,+Massey,+L.D.+%26+Mutter,+S.A.+(Eds.)+(1988).+Intelligent+tutoring+systems:+Lessons+learned.+Hillsdale,+BJ:+Erlbaum&ots=sS7DWf3F&sig=KvMZcqUpedo5cjVVYSlkrnmIFzM

Rajani, N.F., & Mooney, R. J. (2017). Using explanations to improve ensembling of visual question answering systems. *Training*, *82*, 248–349.

Rasmussen, J. (1983). Skills, rules, and knowledge; signals, signs, and symbols, and other distinctions in human performance models. *IEEE Transactions on Systems, Man, and Cybernetics*, (3), 257–266.

Rasmussen, J., Nixon, P., & Warner, F. (1990). Human error and the problem of causality in analysis of accidents. *Philosophical Transactions of the Royal Society of London B: Biological Sciences*, *327*(1241), 449–462.

Reed, S., Akata, Z., Lee, H., & Schiele, B. (2016). Learning deep representations of fine-grained visual descriptions. In *Proceedings of the IEEE Conference on Computer Vision and*

*Pattern Recognition* (pp. 49–58). Retrieved from http://www.cv-foundation.org/openaccess/content_cvpr_2016/html/Reed_Learning_Deep_Representations_CVPR_2016_paper.html

Rehder, B. (2003). A causal-model theory of conceptual representation and categorization. J*ournal of Experimental Psychology: Learning, Memory, and Cognition, 29,* 1141–1159.

Riedl, M. O., & Young, R. M. (2010). Narrative planning: Balancing plot and character. *Journal of Artificial Intelligence Research*, *39*, 217–268.

Reiss, J. (2012). The explanation paradox. *Journal of Economic Methodology, 19*(1), 43–62.

Rey, G.D. (2012). A review of research and a meta-analysis of the seductive detail effect. *Educational Research Review*, *7*(3), 216–237.

Ribeiro, M.T., Singh, S., & Guestrin, C. (2016a). Model-agnostic interpretability of machine learning. *ArXiv Preprint ArXiv:1606.05386*. Retrieved from https://arxiv.org/abs/1606.05386

Ribeiro, M.T., Singh, S., & Guestrin, C. (2016b). Why Should I Trust You?: Explaining the Predictions of Any Classifier. In *Proceedings of the 22nd ACM SIGKDD International Conference on Knowledge Discovery and Data Mining* (pp. 1135–1144). ACM. Retrieved from http://dl.acm.org/citation.cfm?id=2939778

Rips, L.J., Brem, S. K., & Bailenson, J.N. (1999). Reasoning dialogues. *Current Directions in Psychological Science*, *8*(6), 172–177.

Ritter, F., & Feurzeig, W. (1988). Teaching real-time tactical thinking. *Intelligent Tutoring Systems: Lessons Learned*, 285–301.

Rittle-Johnson, B. (2006). Promoting transfer: Effects of self-explanation and direct instruction. *Child Development*, *77*(1), 1–15.

Robnik-Šikonja, M., & Kononenko, I. (2008). Explaining classifications for individual instances. *IEEE Transactions on Knowledge and Data Engineering*, *20*(5), 589–600.

Robnik-Šikonja, M., Likas, A., Constantinopoulos, C., Kononenko, I., & Štrumbelj, E. (2011). Efficiently explaining decisions of probabilistic RBF classification networks In A. Dobnikar, U. Lotric, & B. Ster (Eds.), *Proceedings of the International Conference on Adaptive and Natural Computing Algorithms,* Part I, LNCS 6593, pp. 169–179. Berlin: Springer-Verlag.

Rook, F.W., & Donnell, M.I. (1993). Human cognition and the expert system interface: mental models and inference explanations. *IEEE Transactions on Systems, Man, and Cybernetics, 23*, 1649-1661.

Rosenberg, J. (1981). On understanding the difficulty in understanding. In H. Parret & J. Bouveresse (Eds.), *Meaning and understanding*. Berlin: De Gruyter.

Rozenblit, L., & Keil, F. (2002). The misunderstood limits of folk science: an illusion of explanatory depth. *Cognitive Science, 26*, 521–562.

Rosenthal, S., Selvaraj, S. P., & Veloso, M. M. (2016). Verbalization: Narration of Autonomous Robot Experience. In *IJCAI* (pp. 862–868). Retrieved from https://www.ijcai.org/Proceedings/16/Papers/127.pdf

Ross, A., Hughes, M.C., & Doshi-Velez, F. (2017). Right for the right reasons: Training differentiable models by constraining their explanations. In *Proceedings of the Twenty-Sixth International Joint Conference on Artificial Intelligence* (pp. 2662-2670). [arXiv:1703.03717v2].

Roth, E. M., Butterworth, G., & Loftus, M. J. (1985). The problem of explanation: placing computer generated answers in context. *Proceedings of the Human Factors and Ergonomics Society Annual Meeting, 29*, 861–863.

Rothermel, G., Burnett, M., Li, L., Dupuis, C., & Sheretov, A. A methodology for testing spreadsheets (2001). *ACM Transactions on Software Engineering and Methodology 10*, 110-147.

Rouse, W. B., & Morris, N. M. (1986). On looking into the black box: Prospects and limits in the search for mental models. *Psychological Bulletin, 100*(3), 349.

Royce, J. (1963). Factors as theoretical constructs. *American Psychologist, 18*(8), 522–528. https://doi.org/10.1037/h0044493

Rozeboom, W. (1956). Mediation variables in scientific theory. *The Psychological Review, 63*(4), 249–264. https://doi.org/10.1037/h0043718

Rozenblit, L., & Keil, F. (2002). The misunderstood limits of folk science: an illusion of explanatory depth. *Cognitive Science, 26*, 521–562.

Sacks, H. & Schegloff, E. (1974). A simplest systematics for the organization of turn-taking for conversation. *Language, 50*, 696-735

Sadeghi, F., Kumar Divvala, S.K., & Farhadi, A. (2015). Viske: Visual knowledge extraction and question answering by visual verification of relation phrases. In *Proceedings of the IEEE conference on computer vision and pattern recognition* (pp. 1456–1464).

Sala, G., & Gobet, F. (2017). Does far transfer exist? *Current Directions in Psychological Science, 26*, 515-520.

Salmon, W.C. (1989). *Four decades of scientific explanation*. Minneapolis: University of Minnesota Press.

Samek, W., Binder, A., Montavon, G., Lapuschkin, S., & Müller, K.-R. (2017). Evaluating the visualization of what a deep neural network has learned. *IEEE Transactions on Neural Networks and Learning Systems*, *28*(11), 2660–2673.

Samurçay, R., & Hoc, J. M. (1996). Causal versus topographical support for diagnosis in a dynamic situation. *Le Travail Humain*, 45-68.

Santos, E.E., Santos, E., Wilkinson, J.T., & Xia, H. (2009). On a framework for the prediction and explanation of changing opinions. In *Systems, Man and Cybernetics, 2009. SMC 2009. IEEE International Conference on* (pp. 1446–1452). IEEE. Retrieved from http://ieeexplore.ieee.org/abstract/document/5346294/

Sarner, M.H., & Carberry, S. (1992). Generating tailored definitions using a multifaceted user model. *User Modeling and User-Adapted Interaction, 2*, 181–210.

Sauro, J., & Dumas, J.S. (2009). Comparison of three one-question, post-task usability questionnaires. In *Proceedings of the SIGCHI Conference on Human Factors in Computing Systems* (pp. 1599–1608). ACM. Retrieved from http://dl.acm.org/citation.cfm?id=1518946

Schaefer, K.E. (2013). *The perception and measurement of human-robot trust*. University of Central Florida Orlando, Florida.

Schaffernicht, M., & Groesser, S. (2011). A comprehensive method for comparing mental models of dynamic systems. *European Journal of Operational Research*, *210*, 57–67. https://doi.org/10.1016/j.ejor.2010.09.003

Schank, R.P. (2013). *Explanation patterns: Understanding mechanically and creatively*. Psychology Press. Retrieved from

https://books.google.com/books?hl=en&lr=&id=EZh_AAAAQBAJ&oi=fnd&pg=PR5&dq=Schank,+R.C.+(1986).+Explanation+patterns:+Understanding+mechanically+and+creatively.++Mahwah,+NJ:+Erlbaum.&ots=p_Dw5_plm4&sig=S9S8PxJwbgMGNZWTbtXFUSf1b4A

Schmidhuber, J. (2015). Deep learning in neural networks: An overview. *Neural Networks*, *61*, 85–117.

Schwiep, J. (2017, May 3). The state of explainable AI. [https://medium.com/@jschwiep/the-state-of-explainable-ai-e252207dc46b]

Scott, C., Clancey, W. J., Davis, R., & Shortliffe, E. H. (1977). *Explanation capabilities of production-based consultation systems* (Stanford Heuristic Programming Project No. STAN-CS-77-593) (pp. 1–30).

Scott, C., William J. Clancey, Davis, R., & Shortliffe, E. (1984). Methods for generating explanations. In *Rule-based expert systems: The MYCIN experiments of the Stanford Heuristic programming project* (pp. 338–362).

Scriven, M. (1958). Definitions, explanations, and theories. *Concepts, Theories, and the Mind-Body Problem*, *2*, 99–195.

Searle, J. R. (1985). *Expression and meaning: Studies in the theory of speech acts*. Cambridge University Press.

Sebok, A., & Wickens, C. (2017). Implementing lumberjacks and black swans Into model-based tools to support human-automation interaction. *Human Factors*, *59*(2), 189–203. https://doi.org/10.1177/0018720816665201

Seegebarth, B., Muller, F., Schattenberg, B., & Biundo, S. (2012). Making hybrid plans more clear to human users — A formal approach for generating sound explanations. In *Proceedings of the Twenty-Second International Conference on Automated Planning and Scheduling* (pp. 225-233). Menlo Park, CA: Association for the Advancement of Computer Science.

Selfridge, M., Daniell, J., & Simmons, D. (1985). Learning causal models by understanding real-world natural language explanations. In *Proceedings of the Second Conference on Artificial Intelligence Application*s (pp. 378-383). New York: IEEE.

Selvaraju, R. R., Das, A., Vedantam, R., Cogswell, M., Parikh, D., & Batra, D. (2016). Grad-cam: Why did you say that? visual explanations from deep networks via gradient-based localization. *ArXiv Preprint ArXiv:1610.02391*. Retrieved from https://arxiv.org/abs/1610.02391

Shafto, P., & Goodman, N. (2008). Teaching games: Statistical sampling assumptions for learning in pedagogical situations. In *Proceedings of the 30th annual conference of the Cognitive Science Society* (pp. 1632–1637). Cognitive Science Society Austin, TX.

Shafto, P., Goodman, N. D., & Griffiths, T. L. (2014). A rational account of pedagogical reasoning: Teaching by, and learning from, examples. *Cognitive Psychology*, *71*, 55–89.

Shank, R. C. (1986). *Explanation patterns*. Hillsdale, NJ: Lawrence Erlbaum Associates.

Shank, G. (1998). The extraordinary powers of abductive reasoning. *Theory and Psychology, 8*, 841-860. [https://doi.org/10.1177/0959354398086007]

Sheh, R.K. (2017). "Why did you do that?" Explainable intelligent robots. In *Proceedings of The AAAI-17 Workshop on Human-Aware Artificial Intelligence* (pp. 628-634). Menlo Park, CA: Association for the Advancement of Artificial Intelligence.

Sheh, R.K, & Monteath, I. (2008). Defining explainable AI for Requirements Analysis. *Künstliche Intelligenz* [https://doi.org/10.1007/s13218-018-0559-3].

Shinsel, A., Kulesza, T., Burnett, M., Curran, W., Groce, A., Stumpf, S., & Wong, W.-K. (2011). Mini-crowdsourcing end-user assessment of intelligent assistants: A cost-benefit study. In *Visual Languages and Human-Centric Computing (VL/HCC), 2011 IEEE Symposium on* (pp. 47–54). IEEE. Retrieved from http://ieeexplore.ieee.org/abstract/document/6070377/

Shortliffe, E. H. (1976). *MYCIN: Computer-based medical consultations*. Elsevier, New York.

Shwartz-Ziv, R., & Tishby, N. (2017). Opening the Black Box of Deep Neural Networks via Information. *ArXiv Preprint ArXiv:1703.00810*. Retrieved from https://arxiv.org/abs/1703.00810

Silverman, B. G. (1992). Building a better critic-recent empirical results. *IEEE Expert*, *7*(2), 18–25.

Simon, H. A. (1992). What is an "explanation" of behavior? *Psychological Science, 3*, 150-161.

Simonyan, K., Vedaldi, A., & Zisserman, A. (2013). Deep inside convolutional networks: Visualising image classification models and saliency maps. In *Proceedings of the Conference on Advanced in Neural Information Processing Systems*. [arXiv:1312.6034v2].

Singh, I.L., Molloy, R., & Parasuraman, R. (1993a). Automation-induced" complacency": Development of the complacency-potential rating scale. *The International Journal of Aviation Psychology*, *3*(2), 111–122.

Singh, I.L., Molloy, R., & Parasuraman, R. (1993b). Individual differences in monitoring failures of automation. *The Journal of General Psychology*, *120*(3), 357–373.

Singh, S., Lewis, R. L., Barto, A. G., & Sorg, J. (2010). Intrinsically motivated reinforcement learning: An evolutionary perspective. *IEEE Transactions on Autonomous Mental Development*, *2*(2), 70–82.

Sinha, R., & Swearingen, K. (2002). The role of transparency in recommender systems. In *CHI'02 extended abstracts on Human factors in computing systems* (pp. 830–831). ACM. Retrieved from http://dl.acm.org/citation.cfm?id=506619

Sleeman, D., & Brown, J. S. (1982). *Intelligent tutoring systems*. London: Academic Press. Retrieved from https://hal.archives-ouvertes.fr/hal-00702997/

Sloman, S.A. (1994). When explanations compete: The role of explanatory coherence on judgements of likelihood. *Cognition*, *52*(1), 1–21.

Slugoski, B.R., Lalljee, M., Lamb, R. & Ginsburg, G.P. (1993). Attribution in conversational context: Effect of mutual knowledge on explanation-giving *European Journal of Social Psychology, 23,* 219-238.

Smith, D.E. (2012). Planning as an iterative process. In *Proceedings of the Twenty-Sixth AAAI Conference on Artificial Intelligence* (pp. 2180-2185). Menlo Park, CA: Association for the Advancement of Artificial Intelligence.

Smith, R. (2014). Explanation, understanding, and control. *Synthese*, *191*(17), 4169–4200.

Sørmo, F., Cassens, J., & Aamodt, A. (2005). Explanation in case-based reasoning–perspectives and goals. *Artificial Intelligence Review*, *24*(2), 109–143.

Sohrabi, S. , Baier, J.A., & MAcIllraith, S.A. (2011).  Preferred Explanations: Theory and Generation via Planning. In *Proceedings of the Twenty-Fifth AAAI Conference on Artificial Intelligence* (pp. 261-267). Menlo Park, CA: Association for the Advancement of Artificial Intelligence.

Southwick, R.W. (1988). Topic explanation in expert systems. In *8. Annual Technical Conference of the British Computer Society Specialist Group on Expert Systems* (pp. 47–57).

Southwick, R.W. (1991). Explaining reasoning: An overview of explanation in knowledge-based systems. *The Knowledge Engineering Review, 6,* 1-19.

Spector, J.M. (2008). Expertise and dynamic tasks. In *Complex decision making* (pp. 25–37). Springer. Retrieved from https://link.springer.com/content/pdf/10.1007/978-3-540-73665-3_2.pdf

Spiro, R.J., Feltovich, P.J., Coulson, R. L., & Anderson, .Daniel K. (1988). Multiple Analogies for Complex Concepts: Antidotes for Analogy-Induced Misconception in Advanced Knowledge Acquisition. Technical Report No. 439. Retrieved from https://eric.ed.gov/?id=ED301873

Springenberg, J.T., Dosovitskiy, A., Brox, T., & Riedmiller, M. (2014). Striving for simplicity: The all convolutional net. *ArXiv Preprint ArXiv:1412.6806*. Retrieved from https://arxiv.org/abs/1412.6806

Strauch, B. (2017). The automation-by-expertise-by-training interaction. *Human Factors*, *59*(2), 204–228. https://doi.org/10.1177/0018720816665459

Strevens, M. (2004). The causal and unification approaches to explanation unified—causally. *Noûs*, *38*(1), 154–176.

St-Cyr, O., & Burns, C. M. (2002, September). Mental models and ecological interface design: An experimental investigation. In *Proceedings of the Human Factors and Ergonomics Society Annual Meeting* (Vol. 46, No. 3, pp. 270-274). Sage CA: Los Angeles, CA: SAGE Publications.

Stubbs, K., Hinds, P. J., & Wettergreen, D. (2007). Autonomy and common ground in human-robot interaction: A field study. *IEEE Intelligent Systems*, *22*(2), 42-50.

Stumpf S., Rajaram V., Li L., Burnett M., Dietterich T., Sullivan E., Drummond R., & Herlocker J. 2007. Toward harnessing user feedback for machine learning. In *Proceedings of the Conference on Intelligent User Interfaces* (pp. 82-91). New York: Association for Computing Machinery.

Sturm, I., Lapuschkin, S., Samek, W., & Müller, K.-M. (2016). Interpretable deep neural networks for single-trial EEG classification. *Journal of Neuroscience Methods, 274*, 141–145.

Suermondt, H. J. (1992). Explanation in Bayesian Belief Networks (PhD Thesis). Stanford University, Stanford, CA, USA.

Suermondt, J. & Cooper, G. (1992) An evaluation of explanations of probabilistic inference. In *Proceedings of the Symposium on Computer Applications in Medical Care* (pp. 579-585). Bethesda, MD: American Medical Informatics Association.

Suthers, D., Woolf, B., & Cornell, M. (1992). Steps form Explanation planning to model construction dialogs. In *Proceedings of AAAI-92* (pp., 24-30). Menlo Park, CA: Association for the Advancement of Artificial Intelligence.

Swartout, W.R. (1977). A digitalis therapy advisor with explanations. In *Proceedings of the 5th international joint conference on Artificial intelligence-Volume 2* (pp. 819–825). Morgan Kaufmann Publishers Inc. Retrieved from http://dl.acm.org/citation.cfm?id=1623009

Swartout, W.R. (1981). Producing explanations and justifications of expert consulting programs. Retrieved from http://dl.acm.org/citation.cfm?id=889859

Swartout, W.R. (1983). XPLAIN: A system for creating and explaining expert consulting programs. *Artificial Intelligence*, *21*(3), 285–325.

Swartout, W. R. (1985). Explaining and justifying expert consulting programs. In *Computer-assisted medical decision making* (pp. 254–271). Springer. Retrieved from https://link.springer.com/chapter/10.1007/978-1-4612-5108-8_15

Swartout, W.R. (1987). Explanation. *Encyclopedia of Artificial Intelligence*, *1*, 298–300.

Swartout, W.R., & Moore, J. D. (1993). Explanation in second generation expert systems. *Second Generation Expert Systems*, *543*, 585.

Swartout, W.R., Paris, C., & Moore, J. (1991). Explanations in knowledge systems: Design for explainable expert systems. *IEEE Expert*, *6*(3), 58–64.

Sweeney, L. (2013). Discrimination in online ad delivery: Google ads, black names and white names, racial discrimination, and click advertising. *Queue*, *11*(3), 1–19.

Swinney, L. (1995). The explanation facility and the explanation effect. *Expert Systems With Applications, 9*, 557-567.

Symeonidis, P., Nanopoulos, A., & Manolopoulos, Y. (2009). MoviExplain: a recommender system with explanations. In *Proceedings of the third ACM conference on Recommender systems* (pp. 317–320). ACM. Retrieved from http://dl.acm.org/citation.cfm?id=1639777

Szpunar, K. K., Spreng, R. N., & Schacter, D. L. (2014). A taxonomy of prospection: Introducing an organizational framework for future-oriented cognition. *Proceedings of the National Academy of Sciences*, *111*(52), 18414-18421.

Tanner, M.C., & Keuenke, A.M. (1991, June). The roles of task structure and domain functional models. *IEEE Expert*, pp. 50-57.

Tapaswi, M., Zhu, Y., Stiefelhagen, R., Torralba, A., Urtasun, R., & Fidler, S. (2016). MovieQA: Understanding stories in movies through question-answering. In *Computer Vision and Pattern Recognition*. Las Vegas, NV.

Tate, D., Grier, R., A. Martin, C., L. Moses, F., & Sparrow, D. (2016). *A Framework for Evidence Based Licensure of Adaptive Autonomous Systems*. https://doi.org/10.13140/RG.2.2.11845.86247

Teach, R. L., & Shortliffe, E. H. (1981). An analysis of physician attitudes regarding computer-based clinical consultation systems. *Computers and Biomedical Research*, *14*(6), 542–558.

Thagard, P. (1989). Explanatory coherence. *Behavioral and Brain Sciences*, *12*(03), 435–467.

Thanasuan, K., & Mueller, S. (2014). Crossword Expertise as Recognitional Decision Making: An Artificial Intelligence Approach. *Cognitive Science*, *5*, 1018. https://doi.org/10.3389/fpsyg.2014.01018

Theis, L., Oord, A. van den, & Bethge, M. (2015). A note on the evaluation of generative models. *ArXiv Preprint ArXiv:1511.01844*. Retrieved from https://arxiv.org/abs/1511.01844

Thelisson, E., Padh, K., & Elisa Celis, L. (2017). Regulatory mechanisms and algorithms towards trust in AI/ML. In *IJCAI-17 Workshop on Explainable Artificial Intelligence (XAI)*.

Thompson, V. A., & Byrne, R. M. (2002). Reasoning counterfactually: Making inferences about things that didn't happen. *Journal of Experimental Psychology: Learning, Memory, and Cognition*, *28*(6), 1154.

Thorpe, D. A. (1974). The Quartercentenary model of D-N explanation. *Philosophy of Science*, *41*(2), 188–195. https://doi.org/10.1086/288583

Thrun, S. (1995). Extracting rules from artificial neural networks with distributed representations. In *Proceedings of the 7th International Conference on Neural Information Processing Systems,* (pp. 505-512). New York: Association for Computing Machinery.

Timmer, S. T., Meyer, J.-J. C., Prakken, H., Renooij, S., & Verheij, B. (2017). A two-phase method for extracting explanatory arguments from Bayesian networks. *International Journal of Approximate Reasoning*, *80*, 475–494.

Tintarev, N. (2007). Explaining recommendations. In *International Conference on User Modeling* (pp. 470–474). Springer. Retrieved from http://link.springer.com/chapter/10.1007/978-3-540-73078-1_67

Treu, S. (1998). Need for multi-aspect measures to support evaluation of complex human-computer interfaces. In *Proceedings of the Fourth Annual Symposium on Human Interaction With Complex Systems* (pp. 82-191). Washington, DC: IEEE.

Trout, J.D. (2002). Scientific explanation and the sense of understanding, *Philosophy of Science, 69,* 212-233.

Tullio, J., Dey, A. K., Chalecki, J., & Fogarty, J. (2007). How it works: a field study of non-technical users interacting with an intelligent system. In *Proceedings of the SIGCHI Conference on Human Factors in Computing Systems* (pp. 31–40). ACM. Retrieved from http://dl.acm.org/citation.cfm?id=1240630

Tworek, C., & Cimpian, A. (2016). Why do people tend to infer "ought" from "is"? The role of biases in explanation. *Psychological Science*, *27*(8), 1109–1122.

https://doi.org/10.1177/0956797616650875

Usunier, N., Synnaeve, G., Lin, Z., & Chintala, S. (2016). Episodic Exploration for Deep Deterministic Policies: An Application to StarCraft Micromanagement Tasks. *ArXiv Preprint ArXiv:1609.02993*. Retrieved from https://arxiv.org/abs/1609.02993

Van Der Linden, J. (2002). Meta-constraints to aid interaction and to provide explanations. In *Proceedings of the Joint Workshop of the European Research Consortium for Informatics and Mathematics*. Downloaded from [http://oro.open.ac.uk/4914/]

Van Der Linden, S., Leiserowitz, A., Rosenthal, S., & Maibach, E. (2017). Inoculating the public against misinformation about climate change. *Global Challenges*, *1*(2), 1–7. https://doi.org/10.1002/gch2.201600008

Van Fraassen, B. C. (1977). The pragmatics of explanation. *American Philosophical Quarterly*, *14*(2), 143–150.

van Lent, M., Fisher, W., & Mancuso, M. (2004). An explainable artificial intelligence system for small-unit tactical behavior. In *Proceedings of the National Conference on Artificial Intelligence* (pp. 900–907). Menlo Park, CA; Cambridge, MA; London; AAAI Press; MIT Press; 1999.

VanLehn, K., Ball, W., & Kowalski, B. (1990). *Explanation-based learning of correctness: Towards a model of the self-explanation effect*. Department of Psychology, Carnegie Mellon University. Retrieved from http://www.dtic.mil/docs/citations/ADA222364

Varsheny, K.R., & Alemzadeh, H. (2017). One the safety of machine learning: Cyber-physical systems, decision sciences, and data products. *Big Data Journal 5,* 246–255. [arXiv:1610.01256v1].

Vasilyeva, N.Y., & Lombrozo, T. (2015). Explanations and causal judgments are differentially sensitive to covariation and mechanism information. In D.C. Noelle, R. Dale, A. S. Warlaumont, J. Yoshimi, T. Matlock, C. D. Jennings, & P. P. Maglio (Eds.), *Proceedings of the 37th Annual Conference of the Cognitive Science Society* (p. 2475-2480). Austin, TX: Cognitive Science Society.

Vasilyeva, N., Wilkenfeld, D., & Lombrozo, T. (2015). Goals affect the perceived quality of explanations. In *Proceedings of the 37th Annual Conference of the Cognitive Science Society,* (pp. 2469-2474). Austin, TX: Cognitive Science Society.

Vermeulen, J., Luyten, K., van den Howen, E., & Coninx, K. (2013). Crossing the bridge over Norman's gulf of execution: Revealing feedforward's true identity. In *Proceedings of CHI 2013*. (pp. 1931–1940). New York: Association for Computing Machinery.

Voosen, P. (2017). How AI detectives are cracking open the black box of deep learning. *Science | AAAS*, *357*(No. 6346), 22–27.

Wachter, S., Mittelstadt, B., & Russell, C. (2017). Counterfactual explanations without opening the black box: Automated decisions and the GDPR, 1–44.

Wallace, R.J., & Freuder, E.C. (2001). Explanation for whom? In *Proceedings of CP2000 Workshop on Analysis and Visualization*. (pp. 119-129). [http://www.cs.ucc.ie/~osullb/cp01/papers/wallace.ps]

Wallis, J. W., & Shortliffe, E. H. (1981). *Explanatory power for medical expert systems: studies in the representation of causal relationships for clinical consultations*. DTIC Document.

Wallis, J. W., & Shortliffe, E. H. (1984). Customized explanations using causal knowledge. In *Rule-Based Expert Systems: The MYCIN Experiments of the Stanford Heuristic Programming Project* (pp. 371–388). Retrieved from http://www.aaai.org/Papers/Buchanan/Buchanan22.pdf

Walton, D. (2004). A new dialectical theory of explanation. *Philosophical Explorations, 7*(1), 71–89.

Walton, D. (2007). Dialogical models of explanation. Association for the Advancement of Artificial Intelligence. [https://pdfs.semanticscholar.org/f86b/ed742f7b8458ef76f69e448aa9a542340d1a.pdf]

Walton, D. (2011). A dialogue system specification for explanation. *Synthese*, *182*(3), 349–374.

Wang, D., Yang, Q., Abdul, A., & Lim, B.Y. (2019). Designing theory-drive user-centric explainable AI. In *Proceedings of the 2019 CHI Conference on Human Factors in Computing Systems*. New York: Association for Computing Machinery.

Wang, H., & Yeung, D.-Y. (2016). Towards Bayesian deep learning: A survey. *IEEE Transactions On Knowledge And Data Engineering* [arXiv:1604.01662v2]

Wang, L., Jamieson, G. A., & Hollands, J. G. (2009). Trust and reliance on an automated combat identification system. *Human Factors*, *51*(3), 281–291.

Wang, T., Rudin, C., Velez-Doshi, F., Liu, Y., Klampfl, E., & MacNeille, P. (2016). Bayesian rule sets for interpretable classification. In P*roceedings of the IEEE 16th International Conference on Data Mining* (pp. 1269-1274). New York: IEEE.

Wang, T., Rudin, C., Doshi-Velez, F., Liu, Y., Klampfl, E., & MacNeille, P. (2015). Or's of And's for interpretable classification, with application to context-aware recommender systems. *Journal of Machine Learning Research 18,* 1-37.

Weinberger, D. (2017, April 18). Our Machines Now Have Knowledge We'll Never Understand. Retrieved August 29, 2017, from https://www.wired.com/story/our-machines-now-have-knowledge-well-never-understand/

Weiner, J. L. (1980). BLAH, A system that explains its reasoning. *Artificial Intelligence*, *15*, 19–48.

Weiner, J.L. (1989). The effect of user models on the production of explanations. In C. Ellis (Ed.), *Expert knowledge and explanation* (pp. 144-146). New York: John Wiley.

Weston, J., Bordes, A., Chopra, S., Rush, A. M., van Merriënboer, B., Joulin, A., & Mikolov, T. (2015). Towards AI-complete question answering: A set of prerequisite toy tasks. *ArXiv Preprint ArXiv:1502.05698*. Retrieved from https://arxiv.org/abs/1502.05698

Wick, M. R., & Slagle, J. (1989). The partitioned support network for expert system justification. *IEEE Transactions on Systems, Man, and Cybernetics*, *19*(3), 528–535. https://doi.org/10.1109/21.31059

Wick, M. R., & Thompson, W. B. (1992). Reconstructive expert system explanation. *Artificial Intelligence*, *54*(1–2), 33–70.

Wickens, C.D. (1994). Designing for situation awareness and trust in automation. In *Proceedings of IFAC Integrated Systems Engineering Conference*, Baden, Germany, pp. 365-370.

Wilkenfeld, D. A. (2014). Functional explaining: A new approach to the philosophy of explanation. *Synthese*, *191*(14), 3367–3391.

Wilkenfeld, D.A. (2016). Understanding without believing. *Explaining Understanding: New Perspectives from Epistemology and Philosophy of Science*.

Wilkenfeld, D. A., & Lombrozo, T. (2015). Inference to the best explanation (IBE) versus explaining for the best inference (EBI). *Science & Education*, *24*(9), 1059–1077. https://doi.org/10.1007/s11191-015-9784-4

Wilkenfeld, D.A., Plunkett, D., & Lombrozo, T. (2016a). Depth and deference: When and why we attribute understanding. *Philosophical Studies*, *173*(2), 373–393.

Wilkenfeld, D. A., Plunkett, D., & Lombrozo, T. (2016b). Folk attributions of understanding: is there a role for epistemic luck? *Episteme*, 1–26.

Wilkison, BD., Fisk, A. D., & Rogers, W. A. (2007). Effects of mental model quality on collaborative system performance. In *Proceedings of the Human Factors and Ergonomics Society Annual Meeting* (Vol. 51, pp. 1506–1510). Sage Publications Sage CA: Los Angeles, CA. Retrieved from http://journals.sagepub.com/doi/abs/10.1177/154193120705102208

Wille, R. (1997). Conceptual graphs and formal concept analysis. *Conceptual Structures: Fulfilling Peirce's Dream*, 290–303.

Williams, M. D., Hollan, J. D., & Stevens, A. L. (1983). Human reasoning about a simple physical system. *Mental models*, 131-154.

Williams, J.J., Kim, J., Rafferty, A., Maldonado, S., Gajos, K. Z., Lasecki, W. S., & Heffernan, N. (2016). AXIS: Generating explanations at scale with learnersourcing and machine learning. In *Proceedings of the Third (2016) ACM Conference on Learning* (pp. 379-388). New York: Association for Computing Machinery.

Williams, J.J., & Lombrozo, T. (2010). The role of explanation in discovery and generalization: Evidence from category learning. *Cognitive Science*, *34*(5), 776–806.

Williams, J.J., & Lombrozo, T. (2013). Explanation and prior knowledge interact to guide learning. *Cognitive Psychology*, *66*(1), 55–84.

Williams, J.J., Lombrozo, T., & Rehder, B. (2012). Why does explaining help learning? Insight from an explanation impairment effect. In *Proceedings of the 32nd annual meeting of the cognitive science society* (pp. 11–14). Portland, Oregon.

Williams, J.J., Lombrozo, T., & Rehder, B. (2013). The hazards of explanation: Overgeneralization in the face of exceptions. *Journal of Experimental Psychology: General*, *142*(4), 1006.

Wilson, A.G., Dann, C., Lucas, C.G., and Zing, Z.p. (2015). The human kernel. In *Proceedings of the 20th Annual Conference on Neural Information Processing Systems (NIPS)* (pp. 1-10). New York: Association for Computing Machinery. [rXiv:1510.07389v3].

Wilson, J. R., & Rutherford, A. (1989). Mental models: Theory and application in human factors. *Human Factors*, *31*(6), 617-634.

Wood, S. (1986). Expert systems for theoretically ill-formulated domains. In *Research and development in expert systems III* (pp. 132–139). Cambridge University Press.

Woodward, J. (2005). *Making things happen: A theory of causal explanation*. Oxford university press. Retrieved from https://books.google.com/books?hl=en&lr=&id=vvj7frYow6IC&oi=fnd&pg=PR7&dq=Woodward,+J.+(2005).+Making+things+happen:+A+theory+of+causal+explanation&ots=xScMdhyQXR&sig=MPmC6rQPN5XkLcV5xmv3XxDlsL0

Woolf, B. (2007). *Building Intelligent Interactive Tutors: Student-centered strategies for revolutionizing e-learning*. Morgan Kaufmann Publishers Inc.

Woolf, B., Beck, J., Eliot, C., & Stern, M. (2001). Growth and Maturity of Intelligent Systems: A status report. In *Smart machines in education: The coming revolution in educational technology* (pp. 99–144). MIT press.

Wright, J. L., Chen, J. Y. C., Barnes, M. J., & Hancock, P. A. (2016). Agent reasoning transparency's effect on operator workload. *Proceedings of the Human Factors and*

*Ergonomics      Society      Annual      Meeting*,      *60*(1),      249–253. https://doi.org/10.1177/1541931213601057

Xu, H., & Saenko, K. (2016). Ask, attend and answer: Exploring question-guided spatial attention for visual question answering. In *European Conference on Computer Vision* (pp. 451–466). Springer. Retrieved from http://link.springer.com/chapter/10.1007/978-3-319-46478-7_28

Yap. G.-E., Tan, A.-H., & Pang, H.H. (2008). Explaining Inferences in Bayesian Networks. (2008). A*pplied Intelligence, 29*, 263-278.

Yates, J. F., Veinott, E. S., & Patalano, A. L. (2003). Hard decisions, bad decisions: On decision quality and decision aiding. *Emerging Perspectives on Judgment and Decision Research*, 13–63.

Ye, L. R., & Johnson, P. E. (1995). The impact of explanation facilities on user acceptance of expert systems advice. *MIS Quarterly*, *19*(2), 157–172. https://doi.org/10.2307/249686

Yosinski, J., Clune, J., Nguyen, A., Fuchs, T., & Lipson, H. (2015). Understanding neural networks through deep visualization. *ArXiv Preprint ArXiv:1506.06579*. Retrieved from https://arxiv.org/abs/1506.06579

Young, R. M. (2014). Surrogates and mappings: Two kinds of conceptual models for interactive devices. In *Mental models* (pp. 43-60). Psychology Press.

Zachary, W. & Eggleston, R.G. (2002). A development environment and methodology for the design of work-centered user interface systems. In *Proceedings of the Human Factors and Ergonomics Society 46th Annual Meeting* (pp. 656-660). Santa Monica, CA: Human Factors and Ergonomics Society.

Zahavy, T., Ben-Zrihem, N., & Mannor, S. (2016). Graying the black box: Understanding DQNs. In *International Conference on Machine Learning* (pp. 1899–1908). Retrieved from http://www.jmlr.org/proceedings/papers/v48/zahavy16.pdf

Zavod, M., Rickert, D.E., & Brown, S.H. (2002). The automated card sort as an interface design tool: A comparison of products. In *Proceedings of the Human Factors and Ergonomics Society 46th Annual Meeting* (pp. 646-650). Santa Monica, CA: Human Factors and Ergonomics Society.

Zeiler, M. D., & Fergus, R. (2014). Visualizing and understanding convolutional networks. In *European conference on computer vision* (pp. 818–833). Springer. Retrieved from http://link.springer.com/chapter/10.1007/978-3-319-10590-1_53

Zeiler, M. D., Krishnan, D., Taylor, G. W., & Fergus, R. (2010). Deconvolutional networks. In *Computer Vision and Pattern Recognition (CVPR), 2010 IEEE Conference on* (pp. 2528–2535). IEEE.

Zeiler, M.D., Taylor, G.W., & Fergus, R. (2011). Adaptive deconvolutional networks for mid and high level feature learning. In *Proceedings of the International Conference on Computer Vision* (pp. 2018-2025). New York: IEEE. [ DOI:10.1109/ICCV.2011.6126474].

Zhang, K., & Wickens, C. D. (1987). A study of the mental model of a complex dynamic system: The effect of display aiding and contextual system training. In *Proceedings of the Human Factors Society Annual Meeting* (Vol. 31, pp. 102–106). SAGE Publications Sage CA: Los Angeles, CA.

Zhang, Y., Sreedharan, S., Kulkarni, A., Chakraborti, T., Zhuo, H. H., & Kambhampati, S. (2016). Plan explicability and predictability for robot task planning. In *Robotics and Automation (ICRA), 2017 IEEE International Conference on* (pp. 1313–1320). IEEE. Retrieved from http://ieeexplore.ieee.org/abstract/document/7989155/

Zhou, B., Khosla, A., Lapedriza, A., Oliva, A., & Torralba, A. (2016). Learning deep features for discriminative localization (pp. 2921–2929). Presented at the Proceedings of the IEEE Conference on Computer Vision and Pattern Recognition.

Zwaan, R. A., Langston, M. C., & Graesser, A. C. (1995). The construction of situation models in narrative comprehension: An event-indexing model. *Psychological Science*, *6*(5), 292–297. https://doi.org/10.1111/j.1467-9280.1995.tb00513.x

Zwaan, R. A., Magliano, J. P., & Graesser, A. C. (1995). Dimensions of situation model construction in narrative comprehension. *Journal of Experimental Psychology: Learning, Memory, and Cognition*, *21*(2), 386–397. https://doi.org/10.1037/0278-7393.21.2.386

## APPENDIX
## Evaluations of XAI System Performance Using Human Participants

| Authors | Bansal et al., 2014 |
| --- | --- |
| AI System | The system determines the failure modes for computer vision systems, and uses this information to automatically predict failure. It explains the causes of different failures in human-understandable language, which additionally allows for human-aided failure prediction. |
| Research Participants | At least 10 Mechanical Turkers. |
| Explanation Interface | Not specified. |
| Explanation Form | Explanations are either lists or trees of textual descriptions which explain the conditions under which failure may arise. |
| Study Design or Comparison Made | All subjects were in the same experimental condition. |
| Manipulation of Explanation Goodness | Explanations are treated as superior based on the ability of a human user to use them to accurately predict system failure. Thus, explanation goodness is measured by the percentage of a user's predictions that were correct. |
| Evaluation of Mental Models | Not evaluated. |
| Performance (Task or Evaluation Measure) | Same as explanation goodness. |
| Trust/Reliance Measures | Discussed but not evaluated (increasing user trust in predictive systems is a motivation of the work). |

| Authors | Berry & Broadbent, 1987 |
| --- | --- |
| AI System | AI explanations of a river pollutant search task. |
| Research Participants | University students. |
| Explanation Interface | Two forms of explanations:<br>1. A single block explanation at the beginning of the task<br>2. Multiple "why" explanations which the user may request at-will throughout the task |
| Explanation Form | (L) Textual descriptions advising users on their task |
| Study Design or Comparison Made | Explanation versus No Explanation. |
| Manipulation of Explanation Goodness | Only the manipulation of explanation type. |
| Evaluation of Mental Models | Participants who were required to verbalize their reasoning following their receipt of an explanation performed better than participants who were required to verbalize but who did not receive an explanation. |

| Performance Evaluation | The role of explanation in the user's actions on a complex river pollutant search task. Participants had to decide which of a set of factories was responsible for pollution based on computer suggestions about what pollutants to test for. Participants who were able to ask "why" questions about the computer recommendation performed better at the task. |
|---|---|
| Trust/Reliance Measures | Only evaluated in terms of task performance. |

| **Authors** | **Chang et al., 2009** |
|---|---|
| AI System | Text classifier. |
| Research Participants | Eight Mechanical Turkers. |
| Explanation Interface | Not described, presumably text. |
| Explanation Form | Not directly evaluated. |
| Study Design or Comparison Made | Single group design. |
| Manipulation of Explanation Goodness | Not directly evaluated. |
| Evaluation of Mental Models | Not directly evaluated. |
| Performance Evaluation | Human judgments of the "interpretability" (understandability) of the probabilistic topic models developed by a machine learning system.<br>Word Intrusion task—Judgments of whether the machine-derived topics correspond to topics developed by humans. Participants are presented with six words and have to decide which is off topic.<br>Topic Intrusion Task—Participants are shown the title of a document, a snippet from the document. and four topics (represented by the highest-probability words within that topic). The participant chooses the topic that does not belong to the document. |
| Trust/Reliance Measures | Not directly evaluated.<br>*"For three topic models, we demonstrated that traditional metrics do not capture whether topics are coherent or not. Traditional metrics are, indeed, negatively correlated with the measures of topic quality developed in this paper. Our measures enable new forms of model selection and suggest that practitioners developing topic models should thus focus on evaluations that depend on real-world task performance rather than optimizing likelihood-based measures"* (p. 8). |

| Authors | Dzindolet et al., 2003; see also Dzindolet et al., 2002 |
|---|---|
| AI System | Simulated decision aid. |
| Research Participants | 219 University students (in three experiments). |
| Explanation Interface | Text showing the decisions of the decision aids, and text explaining the decision errors that were made by the decision aid (i.e., false alarms, such as "Since shading from a tree sometimes take human-like forms, mistakes can be made"). |
| Explanation Form | Decisions (soldier present vs. soldier absent) made by a hypothetical "contrast detector" computer system. |
| Study Design or Comparison Made | In Experiment 2 there were two conditions: The decisions of the decision aid were either presented or not. A bar graph showed the running error rate of the decision aid and the participants across trials (continuous feedback). Decision aid performed either better (half of the number of errors made by participants in Experiment 1) or worse (twice the number of errors made by participants in Experiment 1). Some participants saw the decision aid's decision after reaching their own decision. In Experiment 3, after the first 100 trials participants were shown the decision aid's cumulative performance. |
| Manipulation of Explanation Goodness | Not evaluated. |
| Evaluation of Mental Models | Initial instructions provided a global explanation of how the "contrast detector" worked, and that is was not always accurate. |
| Performance (Task or Evaluation Measure) | Participants viewed 200 slides of terrain and had to decide whether a camouflaged soldier was in each photo. Participants rated their confidence (in both themselves and the decision aid), trust in the decision aid and degree of reliance on the decision aid, both after the instructions (but before the test trials) and once again after the test trials. Participants were also asked to explain their ratings and justify their decisions. |
| Trust/Reliance Measures | Focus of the work was on the relationship among judgments of automation reliability and trust, and their relation to reliance. Reliable decision aid was rated as more trustworthy. After observing the automated aid make errors, participants distrusted even reliable aids, and showed greater self-reliance, unless an explanation was provided regarding why the aid might err. Reliance increased if reasons for the aid's errors was provided. Reliance can be predicted from trust, but reliance and trust depend on whether system errors can be well-explained. |

| Authors | Fails & Olsen, 2003 |
|---|---|
| AI System | The system, Crayons, creates image classifiers through an interactive machine learning process. Its intended audience is UI designers or other programmers who may have varying levels of ML experience. |
| Research Participants | 10 subjects (no further detail provided), |
| Explanation Interface | Explanations are given via an interactive camera-based interface. |
| Explanation Form | Visual explanations which describe the current status of the classifier (based on user contributions, what pixels in an image are classified as belonging to an object or class of objects? Such pixels are visually highlighted). |
| Study Design or Comparison Made | Subjects divided into two groups based solely on the order they used one of the two algorithms (CWSS first then MSSS, vs. MSSS first then CWSS). |
| Manipulation of Explanation Goodness | Not directly evaluated, though it may be seen as a function of human-machine performance (below) (better performance = better explanations). |
| Evaluation of Mental Models | Not evaluated. |
| Performance (Task or Evaluation Measure) | Human-machine performance was measured based on the amount of time and training iterations it took for the system to achieve 97.5% agreement with the gold standard non-interactive classifier. |
| Trust/Reliance Measures | Not evaluated. |

| Authors | Fiebrink et al., 2011 |
|---|---|
| AI System | Interactive supervised learning system for real-world gesture analysis problems. The AI system supports human-computer interaction where human users are engaged in tasks such as choosing training algorithms, evaluating/comparing models, and supplying training data. Users are engaged in tasks such as choosing training algorithms, evaluating/comparing models, and supplying training data. |
| Research Participants | Students and faculty member of music department, professional cellist/composer. In the first study ("A"), human subjects were seven composers. In the second ("B"), they were 21 undergraduate students from a variety of majors and years. In the third ("C"), the subject was a single professional cellist/composer. |
| Explanation Interface | The interface was a GUI, but explanations usually involved purely sound-based feedback. Explanations are model outputs (textual/visual/aural feedback), used to improve user understanding of a machine learning system. |
| Explanation Form | In studies A and B, explanations were the sonic outputs produced by their interactions with the model. In study C, the participant was also |

| | able to view textual or graphical visualizations of output. |
|---|---|
| Study Design or Comparison Made | The differences between the three studies involved the number and type of participants, and with what goal they interacted with the system. In all studies, the same software was used (Wekinator), though participants in different studies used this software differently. |
| Manipulation of Explanation Goodness | Explanation goodness per se was not evaluated. Researchers examined what model criteria are important to users in interactive music performance systems. Evaluation criteria were correctness, cost, decision boundary shape, label confidence and posterior shape, and complexity and unexpectedness |
| Evaluation of Mental Models | User mental models were not evaluated for soundness. |
| Performance (Task or Evaluation Measure) | Human-machine performance was measured by the users via direct evaluation criteria and subjective evaluation criteria. |
| Trust/Reliance Measures | Trust and reliance were not evaluated. |

| Authors | Goguen et al., 1983 |
|---|---|
| AI System | The BLAH explanation system. |
| Research Participants | Indirectly, via the analysis of transcripts. |
| Explanation Interface | Tree structures of statements and justifications; Three explanation sources (i.e., Watergate transcripts, interview of participants' career choice, interview of income tax preparation, |
| Explanation Form | Linguistic unit whose overt structure consists of statement(s) to be justified; Tree structure of statements and justifications ("reasons"). |
| Study Design or Comparison Made | Three types of justification: giving a reason, giving examples, and eliminating alternatives. Analysis of dialog structures, in the form of a transformational grammar. |
| Manipulation of Explanation Goodness | Not evaluated. |
| Evaluation of Mental Models | Evaluated in the sense of being a naturalistic investigation of argument (embeddings) structures and dialogs. |
| Performance (Task or Evaluation Measure) | Analysis of three examples of human-to-human explanation to reveal the various forms of explanation (e.g., a statement followed by a reason, such as "*I used the tax short form because I didn't have much money*"). Emphasis is on assessing explanation as a collaborative dialog. |
| Trust/Reliance Measures | Not evaluated. Purpose is to support the design of expert systems with better artificial reasoning and explanation capabilities. |

| Authors | de Greef and Neerincx, 1995 |
|---|---|
| AI System | The statistics program HOMALS and a mock-up of a decision aid. |
| Research Participants | Twenty novice to apprentice HOMALS users, plus experts (number unspecified) to develop the task model. |
| Explanation Interface | "Aiding" interfaces identify the knowledge a user lacks, and present contextually relevant information to fill this gap when/if it is needed. What information is provided as aiding is determined via procedural task knowledge obtained from human experts. |
| Explanation Form | Has accompanying statistical scatterplots that show the data being statistically analyzed, but these are not key to the explanations. Interface includes a text window which displays the aiding information in a written form. |
| Manipulation of Explanation Goodness | Not evaluated, but it is argued that the best way to develop task models is to study the performance and reasoning of experts. Interviews were conducted to derive the "expert task model." |
| Evaluation of Mental Models | Argue that user experience and knowledge must be considered in the design of explanation systems. |
| Performance (Task or Evaluation Measure) | Using a mock-up of the statistical analysis program, HOMALS, participants had to modify the depiction of data sets so as to answer questions that were asked of the data. Main measure was correctness of the imposed data transformations. Performance of introductory statistics students was not affected by the aiding system. Partitipants whose knowledge of statistics was somewhat greater than introductory level performed better and learned more. |
| Trust/Reliance Measures | Broad goal is is to create a comprehensive method for designing interfaces. Using experts, one develops the "aiding functions" that would make the interface easy to learn and use. Iterate and assess the interface redesign and the aiding functions (usability analysis). |

| Authors | Groce et al., 2014 |
|---|---|
| AI System | The system is designed to be able to interactively assist users in the detection of faults in machine learning classifiers. |
| Research Participants | 48 university students with little-to-no background in computer science, |
| Explanation Interface | Interactive GUI incorporating textual and graphical information, |
| Explanation Form | Explanations communicate classifier prioritization strategies to users via small graphical or numerical icons. |
| Study Design or Comparison Made | All participants received four experimental conditions: one control, and one for each of the system's three prioritization methods, |
| Manipulation of Explanation Goodness | Explanations were evaluated for goodness based on human-machine performance (below) relative to the control. |

| Evaluation of Mental Models | Not evaluated. |
|---|---|
| Performance (Task or Evaluation Measure) | Human-machine performance evaluated based on 1) the ability of users to find classifier faults; 2) the efficiency of user test coverage; and 3) subjective user attitudes toward the system. |
| Trust/Reliance Measures | Not evaluated. |

| Authors | Herlocker et al., 2000 |
|---|---|
| AI System | The system provides explanatory interfaces to justify the black-box decisions made by movie recommendation systems (namely, MovieLens). |
| Research Participants | Experiment 1 used 78 participants selected from the userbase of the MovieLens online movie recommendation system; Experiment 2 used 210 participants from the same pool of users. |
| Explanation Interface | Explanations delivered to users via the MovieLens website GUI. |
| Explanation Form | Twenty-one different types (or elements) of explanations, including textual, graphical, and numerical information (see Table 1 in the paper for more details). |
| Study Design or Comparison Made | Experiment 1 was a randomized block design, where blocks were users and the treatments were the different forms of explanation; Experiment 2 divided participants into 7 groups (two controls with no explanations, and five others each with different types of explanations). |
| Manipulation of Explanation Goodness | Evaluated in Experiment 1 by presenting to users different explanations for the same movie recommendation and seeing which explanations most increased user likelihood to see the movie, |
| Evaluation of Mental Models | Discussed but not evaluated. |
| Performance (Task or Evaluation Measure) | Human-machine performance was evaluated in Experiment 2 by measuring filtering performance (percent of correct decisions made by users). |
| Trust/Reliance Measures | Not evaluated. |

| Authors | Kass & Finin, 1988 |
|---|---|
| AI System | The system, GUMAC, operates in tandem with interactive advisory expert systems to build user models representing the user's domain knowledge. |
| Research Participants | 2 users cases with varying knowledge of the domain. |
| Explanation Interface | Not explicated, but likely to be a purely text-based dialogue system. |
| Explanation Form | No explanations given. |

| Study Design or Comparison Made | No research design presented. The only difference between the two users is their level of domain knowledge. |
|---|---|
| Manipulation of Explanation Goodness | Discussed but not evaluated. |
| Evaluation of Mental Models | GUMAC produces representations of user mental models, but the work does not empirically evaluate them. |
| Performance (Task or Evaluation Measure) | Not evaluated. |
| Trust/Reliance Measures | Not evaluated. |

| Authors | Kim et al., 2015 |
|---|---|
| AI System | The system is an inference algorithm that generates plans from the conversations of human planners. It specifically caters to deploying autonomous systems in time-critical domains (where speech data may be scarce and noisy) and makes use of a hybrid probabilistic and logical method for plan generation to compensate for the poor data quality. |
| Research Participants | Participants were 46 Mechanical Turkers. They were divided into pairs. |
| Explanation Interface | No explanations (the paper simply describes an algorithm). |
| Explanation Form | No explanations. |
| Study Design or Comparison Made | Three conditions were compared: 1) perfect PDDL (plan validator) files; 2) perfect domain file, but problem file has some missing information; 3) perfect problem file, but domain file has some missing information; and 4) no PDDL files. |
| Manipulation of Explanation Goodness | Not evaluated. |
| Evaluation of Mental Models | Not evaluated. |
| Performance (Task or Evaluation Measure) | Participants' task was to validate the performance of the plan with a metric of 83% accuracy. The algorithm's performance evaluated in terms of: 1) task allocation accuracy (the system's ability to infer what aspects of the plan are to be retained and what aspects should be rejected); 2) plan sequencing accuracy (the system's ability to correctly order plan subcomponents); and 3) overall plan accuracy (arithmetic mean of the two other metrics). |
| Trust/Reliance Measures | Not evaluated. |

| Authors | Kulesza, et al., 2010 |
|---|---|
| AI System | The system works to debug machine-learned programs by exchanging explanations with its user, such that the explanation the system gives describes how the program arrived to an incorrect solution, and the user uses this to explain in-turn where the program went wrong. The specific domain focus here is in transcript coding (hereafter just called "coding"). |
| Research Participants | Study 1 used nine psychology and HCI students with some experience programming, but no experience in machine learning; Study 2 used 74 local students and nearby residents with neither experience in machine learning nor (except one person) programming. |
| Explanation Interface | Explanations were delivered in Study 1 via paper printouts of a coded transcript which had been annotated with explanations. In Study 2, explanations were delivered via a GUI prototype. Explanations detail program logic (in lieu of source code) and runtime data; user interacts with explanations to verify that the system logic is correct. |
| Explanation Form | Heavy use of graphs and charts to summarize summaries of runtime output (via charts). Explanations in Study 1 were textual annotations explaining what semantic info was most influential in performing a text-categorization task. In Study 2, explanations generated by the system were graphical summaries of run-time output and visual highlighting of segments or words in the transcript. |
| Study Design or Comparison Made | Study 1 involved no software, but Study 2 involved the software prototype. Study 2 isolated the impact of explanations about logic and those about runtime output by having four different versions of the prototype software (one control, one with logic explanations, one with runtime explanations, and one with logic and runtime explanations) and recording how participant performance varied depending on which version was used to debug. |
| Manipulation of Explanation Goodness | Study 1 does not touch on this. In Study 2, the effectiveness of explanations was evaluated based on: 1) objective improvement of the accuracy of the program (before the system-aided debugging vs. after the debugging), and 2) based on participant attitudes towards the system (what types of debugging aid they thought were most useful). |
| Evaluation of Mental Models | Study 1 explored how users reason about and correct a text-classification program; how explanations might influence users' existing mental model; participant attitudes toward explanatory debugging; how user mental models are observed to flexibly accommodate new information as it is presented to them via explanations, thus becoming more complex as the user internalizes the |
| | system's logical process. Study #2 does not touch on this. |
| Performance (Task or Evaluation Measure) | Asked whether explanatory debugging works, exposing a learned program's logic to end users and for eliciting user corrections to improve the program's predictions and enable end users to debug learned programs via a Natural Programming methodology. Human performance in Study #1 was measured by the degree by which users' mental models grow to incorporate system logic after explanations. Human-machine performance in Study 2 was measured by the degree by which system accuracy improves due to the interactive debugging. |
| Trust/Reliance Measures | Trust and reliance were not evaluated. |

| Authors | Kulesza, Burnett et al., 2011 |
|---|---|
| AI System | WYSIWYT/ML (What You See Is What You Test for Machine Learning). The system (1) advises the user about which predictions to test, then (2) contributes more tests "like" the user's, (3) measures how much of the assistant's reasoning has been tested, and (4) continually monitors over time whether previous testing still "covers" new behaviors the assistant has learned. |
| Research Participants | Forty-eight university students. |
| Explanation Interface | Why-oriented approach: explains system logic/state and provides a debugging mechanism which allows an end user to directly influence system behavior; examination of barriers to user efficacy in the context of such a system. |
| Explanation Form | Pie chart depicts the Confidence of the intelligent assistant in its classifications, aligned with a list of the possible classifications. A different window showed "oddball" messages—those least similar to the messages that the assistant has learned from (Similarity). A numerical widget showed the number of relevant words (0 to 20) to explain the reason for the priority assigned to the message (relevance). The interface features were designed so as to satisfy the goodness criteria (see below). |
| Study Design or Comparison Made | Study is essentially a test of method for end users to assess whether and when to rely on their intelligent assistants. Primary task of the intelligent assistant is to determine the major topic of messages (cars, computers, religion), using a support vector mechanism. (trained to 85% correct performance). In a counterbalanced within-subject design, participants worked on message classification (200 messages in total) with the interface showing one of three widgets on each trial, either confidence, similarity or relevance, or none of these (control condition). |
| Manipulation of Explanation | WYSIWYT/ML prioritizes the intelligent assistant's topic predictions that are most likely to be wrong, based on the assistant's certainty that |

| Goodness | the topic it predicted is correct. Researchers relied on "cognitive dimensions": The user should not have to manually track or compute things. There should not be hidden dependencies. Users should not have to make a decision before they have information about the decision's consequences. the interface should allow users to annotate, change the screen layout, and communicate informally with themselves or with other users.  Users should be able to see how a component relates to the whole. *"This was initially a problem for our priority widget because it had too many roles: a single widget communicated the priority of assessing the message, explained why it had that priority, and how the message had been assessed—all in one small icon. Thus, we changed the prototype so that no widget had more than one role."* |
|---|---|
| Evaluation of Mental Models | User mental models were not evaluated directly. |
| Performance (Task or Evaluation Measure) | WYSIWYT/ML helped end users find software assistants' mistakes significantly more effectively than ad hoc testing (i.e., serendipitous discovery or "gut instincts").<br>Participants found significantly more incorrect classifcations when they could see the Confidence widget. |
| Trust/Reliance Measures | Study is essentially a test of method for end users to assess whether and when to rely on their intelligent assistants.<br>Participants reported greater satisfaction from using the explanation widgets.  Using the explanation widgets enabled participants to get through more of the 200 test messages. |

| Authors | Kulesza, Stumpf et al., 2011; Stumpf et al., 2007 |
|---|---|
| AI System | Bayesian classification algorithm for text classification. |
| Research Participants | Twenty-two college students. |
| Explanation Interface | Text-based with pull-down menu options and histobar figures showing feature weights. |
| Explanation Form | Interactive explanations are intended to serve as a means of supporting the debugging of machine-learned logic (intelligent assistants). Explanation is regarded as a dialog, in which users can ask why questions about machine decisions, receive explanations of the machine's logic and execution state (i.e., local explanations), and change the explanations by direct manipulation, to see how the machine's determinations change.<br>A generative grammar was used to generate questions, that operated over a restricted ontology (words, messages, files, folders). The range of possibilities was reduced to nine: Why was <message> filed to <folder>? Why wouldn't the [message] be filed to [folder]? Why does |

|  | [word] matter to the [folder]?, etc.  Answers were expressed relative to the intelligent assistant's logic.<br>Histobar diagram shows the importance of features to the given classification. |
|---|---|
| Study Design or Comparison Made | Think aloud task in which pairs of participants collaborated. in the classification of email messages. Protocols were analyzed using a particular coding scheme that referenced the sorts of cognitive difficulties people encounter while debugging programs (Ko, et al., 2004). |
| Manipulation of Explanation Goodness | Not explicitly evaluated, though the work ins premised on notions of what counts as a good explanation in this domain. "*[Answers to] Why-questions of statistical machine learning systems to clarify the program's current behavior provide faithful explanations of the current logic in terms of representations of the current source code and execution state of the program… These answers can help the end user debug intelligent assistants*" (p. 46). |
| Evaluation of Mental Models | Analysis of the protocols focused on the kinds of cognitive barriers participants encountered, statements about the information that would help them when in a given barrier situation (e.g., "Why didn't it do x?"), and the kinds of queries they entered into the system about the system's classifications. |
| Performance (Task or Evaluation Measure) | Participants had to evaluate the categorizations made by the intelligent assistant with regard to the folders to which messages were assigned. User changes to the displayed  debugging features would cause the classifier to recalculate its predictions. |
| Trust/Reliance Measures | Results include a thorough listing of the kinds of questions that users asked that relate to system usefulness, usability, and trust/reliance. Examples are:  Can we do this quicker?,  Can the system handle phrases?,  Why are more messages being filed to [folder]? Was our past action wrong? |

| Authors | Kulesza et al., 2012 |
|---|---|
| AI System | Music recommender system that emulates an adaptable internet radio station hat allows users to create custom "stations" and personalize their preferences.  Database of 36,000 songs. |
| Research Participants | Sixty-two university students; little or no prior computer experience. |
| Explanation Interface | Users can provide feedback about particular songs ("debugging"). Feedback could be with respect to such factors as song tempo and song popularity, artist similarity. Interactivity is suggestive of the notion of explanation as a human-machine dialog. |
| Explanation Form | Aural tutorial on the internal functional *and* structural workings of the agent's reasoning.  Structural information can be considered global |
| | whereas the functional information and examples could be considered local. |
| Study Design or Comparison Made | Two groups, received either No Scaffolding or Scaffolding. Scaffolding was a 15-minute tutorial on how the recommender worked. It included illustrated examples of how the recommender determined artist similarity, the types of acoustic features the recommender "knows" about, and how it extracts this information from audio files. |
| Manipulation of Explanation Goodness | Scaffolding versus No Scaffolding conditions. |
| Evaluation of Mental Models | The research focus was on the effects of mental model soundness. A questionnaire was used to evaluate participant mental models. It includes specific scenarios and required participants to say which action they would take (e.g., "How would you align the recommender to your preference for the Beatles?"). The questionnaire was administered after the initial instructions and after a five day period during which participants used the recommender system. The group receiving the scaffolding performed significantly better on both administrations of the comprehension test. On Day 1, the group not receiving the scaffolding was less certain that they understood how the system recommended songs. Neither group showed much change in mental models comparing Day 1 and Day 5. Therefore, a good initial instruction during the instructional phase of the study can be sufficient to engender good mental models. Comparing mental model soundness to debugging performance, participants whose mental models improved the most reported that the effort of debugging was significantly more worthwhile than participants whose mental models improved less, or not at all. |
| Performance (Task or Evaluation Measure) | Performance on the test scenarios presented in the comprehension questionnaire (number of correct actions chosen). Counts of the number of debugging interactions, interaction time, and cost/benefit (judgments of the effort required to debug the recommender). |
| Trust/Reliance Measures | Participants were asked about their satisfaction with the system, on Day 5. Participants whose mental model soundness increased comparing Day 1 to Day 5 showed greater satisfaction with the recommender system. Conversely, Participants whose mental model soundness decreased comparing Day 1 to Day 5 showed the least satisfaction. *"The idea is great to be able to 'set my preferences', but if the computer continues to play what I would call BAD musical choices— I'd prefer the predictability of using Pandora."* |

| Authors | Kulesza et al., 2013 |
|---|---|
| AI System | Intelligent agent was developed to make personalized song recommendations, complete with explanations describing why each song is recommended. |
| Research Participants | The participants were 17 individuals (ages 19-34) sourced from the local community, none of whom had a background in computer science. |
| Explanation Interface | Explanations which seek to aid a user in debugging a machine learning recommendation system. Emphasis was on soundness vs. completeness of explanations |
| Explanation Form | Explanations were presented to users via paper handouts bearing textual statements of system logic. Explanations contained graphical, textual, and tabular data. Heavy use of graphs and charts. |
| Study Design or Comparison Made | Four different treatment conditions were presented to participants in a between-subjects design: 1) high-soundness high-completeness, 2) high-soundness 3) low-completeness, medium-soundness medium-completeness, and 4) low-soundness high-completeness. |
| Manipulation of Explanation Goodness | Explanation goodness was evaluated on the criterion that, if an explanation is superior, then it would positively impact a user's mental model. Goodness was described in terms of soundness and completeness. |
| Evaluation of Mental Models | Mental models were evaluated for soundness through a scoring method, such that each participant's mental model score was the number of correct minus the number of incorrect statements they made about the system during the study and on a post-study questionnaire. |
| Performance (Task or Evaluation Measure) | Measured time-to-learn and overall accuracy based on the aforementioned mental model scores. |
| Trust/Reliance Measures | Trust in the AI was evaluated by questioning participants on their opinions of the soundness and completeness of the system's explanations, and using the results to create trust scores that could be associated with each treatment. HH (high-soundness, high-completeness) participants trusted their explanations more than participants in other treatment conditions. |

| Authors | Kulesza et al., 2015; see also Kulesza et al., 2010; Kulesza, Stumpf et al., 2011. |
|---|---|
| AI System | Test classification system using Bayesian methods. |
| Research Participants | Seventy-seven participants, university students and individuals from the general population. All reported little or no prior experience in software debugging or machine learning. |
| Explanation Interface | Histobar graph interface shows the user which text features were most important to the classification and a pie chart showing the probability that the classifier's classification is correct. |
| Explanation Form | In explanatory debugging the participants could correct the learning system and observe the results. This is suggestive of the notion of explanation as exploration. "*To enable an iterative cycle of explanations between the system and the user, in Explanatory Debugging the machine-to-user explanation should also serve as the user-to-machine explanation*" (p. 128). An aspect of explanatory debugging is reversibility, through which the user can engage in "self-directed tinkering." |
| Study Design or Comparison Made | Explanation versus No Explanation conditions. the initial instructions did not explain how the classifier made predictions. |
| Manipulation of Explanation Goodness | Good explanations are assumed to be iterative (that is, learned), truthful, complete (but not overwhelming), and actionable. |
| Evaluation of Mental Models | The researchers emphasize the relation of explanation goodness to the formation of valid, rich and useful mental models. After the debugging task, participants were presented a set of messages and were asked to predict how the classifier would classify each of them, and explain why. The test messages were designed such that only one feature was important to the classification, such as an obvious feature or a subtle feature. Participants shown the explanations (during the debugging task) identified more of the subtle features that the classifier would use to make its classifications (in the mental models task). |
| Performance (Task or Evaluation Measure) | Participants could add features to the classification scheme, remove features or adjust features' importance. Participants in the Explanation condition understood how the classifier operated about 50% better than control participants. Participants in the Explanation condition made fewer changes to the feature system but the changes they did make were more effective (greater improvements in the classifier's performance; 10% improvement overall). "*... many control participants' requested explanations similar to those the treatment participants saw suggests the need for machine learning systems to be able to explain their reasoning*" (p. 135). Explanatory debugging is more effective than simply being shown the features that a |
| | classifier uses to classify instances. |
| Trust/Reliance Measures | Nor directly evaluated. |

| Authors | Lamberti and Wallace, 1990 |
|---|---|
| AI System | Expert system designed to assist programmers in performing their job-related diagnostic duties in resolving technical issues in hardware or software. |
| Research Participants | Ninety novice to expert (practicing professional) diagnostic programmers. |
| Explanation Interface | Declarative vs. procedural textual explanations. Participants could view problems at a high/abstract level and then subsequently refine to see finer detail. |
| Explanation Form | Declarative explanation/summary of the problem and procedural/rule-based explanation of how to fix it. |
| Study Design or Comparison Made | Independent variables were task uncertainty, user expertise, knowledge presentation format (declarative vs. procedural), knowledge organization (do problems require abstract vs. concrete knowledge?), and decision-making procedures. Goals of the study were to (1) examine employee satisfaction with the system interface, functions, and capabilities; (2) assess the knowledge presentation formats (e.g., procedural vs. declarative) in diagnostic expert systems used for diagnostic problem solving; (3) examine the effect of task uncertainty on problem-solving and decision-making (speed and accuracy). |
| Manipulation of Explanation Goodness | Performance differences in the two types of explanations (procedural, declarative) are evaluated based on question-answering time, query time, and accuracy. The relationship between explanation goodness and user skill level and task uncertainty was also evaluated. |
| Evaluation of Mental Models | Self-report questionnaire to assess knowledge required for answering abstract vs. concrete questions, knowledge presentation format, and task uncertainty (high vs. low). It is argued that the user performance and confidence dependent variables may be indicative of stronger internalization of the system's processes. |
| Performance Evaluation | Human performance was measured in terms of participant scores for time (subcategories: problem-solving, question-answering, and decision-making time) and accuracy. |
| Trust/Reliance Measures | Trust in the AI was evaluated using participant ratings of their confidence in the accuracy of the system's recommendations, satisfaction with the system recommendations, lines of reasoning, and action plan decisions. |

| Authors | Lerch et al., 1997 |
|---|---|
| AI System | Does not present an AI system. |
| Research Participants | Experiment 1 had 92; Experiment 2 had 44; Experiment 3 67; Experiment 4 had 63; all participants were Carnegie Mellon undergraduates, and no participant took part in more than one of these experiments |
| Explanation Interface | PC program which displayed problems, advice, and sometimes advice rationale, and also allowed the user to provide ratings |
| Explanation Form | Experiment 3: Explanations were textual in either prose or rule-based format. No other experiment involved explanations. |
| Study Design or Comparison Made | Experiment 1: Each participant (randomly) assigned to one of three treatments (recommendations attributed to an expert system, human expert, or human novice); Experiment 2: Same as Experiment 1, except no human novice treatment group; Experiment 3: Each participant assigned to one of three treatments (ES-attributed advice with no explanations, prose explanations, or rule explanations); Experiment 4: Each participant assigned to one of two treatments (knowledge-linked vs. performance-linked descriptions of the ES-attributed advice). |
| Manipulation of Explanation Goodness | Experiment 3 was the only experiment touching on this. It evaluated explanations based on 1) user agreement with the ES-attributed advice given the different explanatory modes, and 2) user confidence in ES-attributed advice. |
| Evaluation of Mental Models | Not evaluated. |
| Performance (Task or Evaluation Measure) | Not evaluated. |
| Trust/Reliance Measures | Trust was measured subjectively and quantitatively with 1) user agreement (source "predictability"); 2) user confidence (source "dependability"); and 3) by examining the causal effects users attributed to the advice given by the source (such as ability, effort, task difficulty, and luck). |

| Authors | Lim & Dey, 2009; Lim et al., 2009 |
|---|---|
| AI System | The AI is a decision aid used for the task of Intensive Care Unit patient diagnosis. It was primarily implemented to provide empirical support for the novel framework presented in this paper, i.e., Context-aware intelligent systems. Makes use of four context-aware applications which function via environmental triggers: 1) an instant messenger plugin which predicts user response time; 2) a program capable of inferring the current status of remote elderly family |

| | members; 3) a program which gives reminders; and 4) a program which gives users (tourism) attraction recommendations. |
|---|---|
| Research Participants | Different age groups with education levels ranged from high school to post-graduate degree. Experiment 1 used 250 people from Mechanical Turk; Experiment 2 used 610 people from Mechanical Turk. The participants included the "co-designers"-- 14 clinicians employed at a local hospital. |
| Explanation Interface | The interface was a GUI that allows users to view a myriad of statistics about patients as well as the system's diagnosis and the explanations thereof. The applications do not give explanations. In Experiment 2, researchers produced questions and corresponding explanations for each application. It is likely that the interface for this was web-based GUI. Experiment 1 provides no explanations. |
| Explanation Form | Explanations came in textual, graphical, and tabular forms. |
| Study Design or Comparison Made | Experiment 1 divided participants into four groups based on which AI system they would be working with. Experiment 2 divided participants into 12 groups: 1 control (no question/explanation) and 11 conditions which each presented questions/explanations of a different type. |
| Manipulation of Explanation Goodness | Evaluated in Experiment 2 through measuring the effect of explanations on user-reported attitudes about satisfaction with the program and usefulness of explanation. Explanations were not quantitatively evaluated, but qualitative reports from participants were used to evaluate explanation strategies to limit user biases and moderate user trust. |
| Evaluation of Mental Models | Mental models were not evaluated. |
| Performance (Task or Evaluation Measure) | Human performance was not directly assessed. |
| Trust/Reliance Measures | Discussed by not evaluated. |

| Authors | Murphy, 1990 |
|---|---|
| AI System | Production copy of an expert system based on financial accounting standards, designed to help accounting firms evaluate the compliance of loan providers. <br> The expert system was likely rule-based. |
| Research Participants | Sixty-four students in a level accounting course (juniors and seniors). |
| Explanation Interface | Not described. |
| Explanation Form | Presumably text but how it was generated and displayed are not described in the report. |
| Study Design or Comparison Made | Three treatment groups: expert system with explanations, expert system without explanations, and a non-automated practice aid as a control |

| | group.<br>The non-automated practice aid was a text-based and abbreviated procedural guide, also based on the compliance standards.<br>Study looked at the effect of expert system use on the ability of novices to classify a client-proposed accounting treatment, and transfer their knowledge to a new situation. |
|---|---|
| Manipulation of Explanation Goodness | Not evaluated in this study. |
| Evaluation of Mental Models | Measure of the development of knowledge ("semantic memory") of the accounting standards, based on a post-test and performance on a test case after completing the experimental (learning) trials, in which participants had to either agree or disagree with a decision made about the loan, which itself could be correct or incorrect. |
| Performance (Task or Evaluation Measure) | Participants had to evaluate the compliance of loans with accounting standards.<br>The expert system led the participant through the evidence-gathering process and presented its determination for each loan transaction.<br>There was also an assessment of learning transfer, involving the analysis of a lease transaction.<br>Performance on the post test (knowledge test) was actually greater for the participants in the non-automated practice aid than participants in the expert system groups, including the group that saw the explanations.<br>Performance on the transfer test as greater for the expert systems groups.<br>The researchers conclude that expert systems created for commercial use may not be viable pedagogical tools in hat they may not impart knowledge or reasoning strategies. |
| Trust/Reliance Measures | Not evaluated in this study. |

| Authors | Narayana et al., 2018 |
|---|---|
| AI System | Artificial task and a simple decision model. |
| Research Participants | 600 Mechanical Turkers. |
| Explanation Interface | Explanations were human-generated but "could have been machine generated" given their form. |
| | Decision sets:<br>In one experiment, rules for when the "alien" is happy or sad given various combinations of setting and meal ingredients.<br>In another experiment rules for when one or another drug is appropriate for an "alien" given the alien's symptoms. |
| | Example<br>1. Alien is checking the news or is coughing implies the weather is windy.<br>2. Snowing or humid and weekend implies alien prefers spices or vegetables and grains. |
| Study Design or Comparison Made | Six experiments in total.<br>The independent variables all involved the complexity of a set of decision rules:<br>Explanation size - the number of terms and rules in the decision sets,<br>Presence/absence of new decision chunks - concepts that added to a decision set (e.g., "windy weather," alien is "coughing"),<br>Repeated or not repeated terms in a decision set. |
| Manipulation of Explanation Goodness | Explanation goodness was not evaluated but participants gave ratings of their satisfaction with each decision set. The ratings task is in some places called "explanation quality" and in some places called "satisfaction." |
| Evaluation of Mental Models | Mental models were not directly evaluated. |
| Performance (Task or Evaluation Measure) | In one experiment, participants were presented with case information consisting of setting (weather), the alien's meal preferences (list of ingredients) as provided by a machine learning system.  Participants' task was to predict whether an alien would be happy or sad with a meal, given the list of ingredients and the setting conditions.<br>In a second experiment, user task was to judge whether a given drug was appropriate for the alien given the symptoms, i.e., "is the alien happy with the prescription."<br>Response time and accuracy were the primary measures. "Accuracy" was whether the response was consistent with the input and the explanation.<br>Overall results were increased explanation length resulted in lower accuracy, longer response time, and lower ratings of explanation quality. |
| Trust/Reliance Measures | Trust/reliance were not directly evaluated. The rating of explanation quality is also sometimes referred to as "subjective satisfaction." |

| Authors | Pacer et al. 2013 |
|---|---|
| AI System | The paper describes four Bayesian network methods for computationally modelling explanation: Most Probable Explanation (MPE), Most Relevant Explanation (MRE), Explanatory Tree (ET), and CET (Causal Explanatory Tree). These methods were evaluated by human users. |
| Research Participants | Experiment 1 recruited 188 participants from Mechanical Turk (though only 109 participants were used in analysis), and most of these participants were male. Experiment 2 recruited 245 participants from Mechanical Turk (but only 165 were retained), and most of these participants were male. |
| Explanation Interface | In Experiment 1, participants provided explanations for an observed effect. These explanations were then coded. In Experiment 2, participants received explanations via an interface that allowed them to manipulate a slider whose value corresponded to their rating of the value of the explanation. |
| Explanation Form | In both experiments, explanations were textual in nature, and their content focused on providing the "single best explanation" (which may include multiple causal factors) for an observed effect. |
| Study Design or Comparison Made | Both experiments had two crossed conditions to which participants were randomly assigned. Conditions were Pearl versus Circuit types of Bayesian networks, and two different topical scenarios (novel diseases vs. lake ecology). |
| Manipulation of Explanation Goodness | Though explanations were evaluated in this experiment, the primary purpose of the research was to evaluate the effectiveness of the four Bayesian models enumerated above. See Performance for more details. |
| Evaluation of Mental Models | Not evaluated. |
| Performance (Task or Evaluation Measure) | Experiment 1 evaluated the explanation quality of participant-generated explanations. These explanations were evaluated for each of the four Bayesian models. Those models that ranked highest were said to be more effective models. <br> Experiment 2 allowed participants to evaluate explanations on a continuous scale (from "very bad" to "very good"). The four models were then evaluated based on a metric of how many of each model's top-k explanations match the top-k explanation rankings of the average participant. |
| Trust/Reliance Measures | Not evaluated. |

| Authors | Poursabzi-Sangdeh et al. (2018) |
|---|---|
| AI System | Weighted model for predicting the prices of rental apartments. |
| Research Participants | 1250 Mechanical Turkers. |
| Explanation Interface | Explanations per se were not provided. Rather , the models were made more or less "transparent" by providing data on fewer or more apartment features, and indications of which features were most important in the models' analyses. |
| Explanation Form | Data tables showing such things as number of rooms days on the market, square footage.<br>Presenting more data (data "transparency") hampered participants' ability to notice apartment anomalous features and correct a model's inaccurate predictions. |
| Study Design or Comparison Made | Four models shown in a within-subjects design: Number of features (2 versus 8 features) and feature visibility (feature weights shown or not shown), all versus No Model baseline condition. |
| Manipulation of Explanation Goodness | Not evaluated. |
| Evaluation of Mental Models | Evaluation confidence judgments. |
| Performance (Task or Evaluation Measure) | Guess of the model's predictions and subject's prediction of the prices of apartments.<br>Drawing attention to the apartment's anomalous features resulted in predictions better enabled the participants to correct the models' mistakes by adjusting their own predictions. |
| Trust/Reliance Measures | Some trials portrayed apartments with anomalously high numbers of rooms; visibility of the feature weights had no impact. Result was counterintuitive.<br>In a separate experiment, participants were told that the predictions came from a human expert. There were no significant differences. |

| Authors | Ribiero et al. 2016 |
|---|---|
| AI System | The system, LIME, works alongside existing classifiers to provide explanations for the classifiers' predictions in a way that is: 1) model-agnostic (the type of classifier does not matter); 2) locally faithful (explains local—but not necessarily global—classifier behaviors); and 3) interpretable (relates inputs and outputs in a way humans can understand). Two versions of LIME were explored: random pick (RP) and submodular pick (SP), each of which alter the method used to select instances to show to the user. |
| Research Participants | The paper includes several simulated-user studies that of course involve no human participants. In the three human-subjects experimental |
| | settings, there are at least 100, at least 15, and exactly 27 participants respectively from Mechanical Turk. In the last setting, subjects were graduate students who have taken at least one graduate course in machine learning; in the other two settings, participants are inexperienced with machine learning. |
| Explanation Interface | For the first two human-subjects experimental settings, explanations are presented alongside the classifier's prediction and the supporting raw data from the classified document. In the first setting, the user may interact with this interface by selecting the best algorithm out of several. In the second, the user is shown only one algorithm, and interacts with the system by selecting words to delete from the classifier. In the final setting, the exact interface is not clear but it allows for the display of classifier predictions—with or without side-by-side explanations—and modules for the user to answer questions. |
| Explanation Form | For the first two human-subjects experimental settings, explanations were bar charts that indicated the relative importance of words (present in the classified text) in terms of their contribution to the classifier's prediction. These explanations occurred alongside the source text in which these words are present to give context. In the second setting, explanations are graphical, and indicate which pixels of an image the classifier used as evidence for its decision. |
| Study Design or Comparison Made | The human-subjects experiment involves three settings, each designed to answer a specific research question. These questions are: 1) Can LIME's explanations help users decide which classifier generalizes better? 2) Can users improve untrustworthy classifiers through feature engineering using LIME? and 3) Can LIME's explanations assist in the identification of the presence of undesirable correlations in a classifier (and thus measure the degree to which the model should actually be trusted)? |
| Manipulation of Explanation Goodness | With regard to whether LIME's explanations could help users decide which classifier generalizes better, the evaluation was based on the number of times users correctly chose the more generalizable classifier (more correct means better explanations). With regard to whether users could improve untrustworthy classifiers through feature engineering, participants took turns removing words from the classifier, and after each such interaction the classifier was retrained and applied to data to detrmine its accuracy. The more the classifier improved over the course of the interactions, the better the explanations. With regard to whether LIME's explanations could asssist in the identification of the presence of undesirable correlations in a classifier  the researchers evaluated the ability of users to correctly identify features in the classifier that were biased/untrustworthy before and after receiving explanations. |
| Evaluation of Mental | Not evaluated. |

| Models | |
|---|---|
| Performance (Task or Evaluation Measure) | Not evaluated. |
| Trust/Reliance Measures | Measured in the third human-subjects setting (before and after exposure to explanations, but always after viewing classifier predictions) by asking each participant if they trusted the model (yes/no implied). Participants had to decide whether to trust a prediction, choose between models, improve an untrustworthy classifier, identify a trustworthy classifier. |

| Authors | Rook & Donnell, 1993 |
|---|---|
| AI System | An expert system designed specifically as a testbed to: 1) instill a certain mental model in its users; 2) independently solve diagnostic problems afflicting a hypothetical, simulated space station; and 3) provide its user with explanations for how it arrived to its solutions. Task context was reactor process control and the evaluation of the likelihood of subsystem failures (e.g., in the emergency core cooling system of the control drive hydraulic system). |
| Research Participants | The participants were 30 undergraduate students with similar math/computer science backgrounds. |
| Explanation Interface | Explanations of the system's underlying rule-based process were encapsulated in an overarching inference network containing nodes that could be expanded by the user into specific rules and decisions. Why and why-not explanations vs. how-to or what-if explanations were shown on a PC, in the task context of explaining a system that employed a decision tree algorithm. |
| Explanation Form | Explanations could be either textual (tracing ES "if…then" logic) or graphical (illustrating the relationship between nodes in the ES network).Text described the internal logic of the decision tree to explain the system in four different ways (Why? Why not? How? What if?) |
| Study Design or Comparison Made | There were six different experimental conditions, observing the interaction between the independent variables: user mental models (good graphical model, good textual model, poor mental model [the control]) and type of ES explanations (textual or graphical, described above). |
| Manipulation of Explanation Goodness | The effectiveness of textual vs. graphical explanations was evaluated for understandability and usefulness based on user performance (the greater the performance, the more effective the explanations). Researchers sought to explore the intelligibility of context-aware intelligent systems: what, why, why not, what if, and how to |
| Evaluation of Mental Models | Mental models were not evaluated after explanations were given. |
| Performance (Task or | Human performance was assessed using: 1) a measure of the number of |
| Evaluation Measure) | key problem-solving nodes identified successfully; 2) a subjective measure indicating subject attitudes towards the ES; and 3) a measure of participants' trajectories through the ES in solving each problem. The hypothesis was that explanations would lead to improved user understanding, trust, perception, and performance more than having no explanations. |
| Trust/Reliance Measures | A survey asked users to report their perceptions of the explanations and system in terms of understandability, trust and usefulness. |

| Authors | Shafto & Goodman, 2008 |
|---|---|
| AI System | The system was a computational model of pedagogical reasoning that assumed that teachers and learners are rational, and that therefore teachers would choose examples that should promote belief in the correct hypothesis (i.e., examples are pedagogically sampled). The experimental scenario used to test this model is called the rectangle game, in which teachers teach learners the shape of an unseen rectangle. |
| Research Participants | Experiment 1 used 18 participants, and Experiment 2 used 29 participants. All were University of Louisville undergraduates. |
| Explanation Interface | The interface for both experiments was run using MATLAB. Teachers used the interface to provide explanations by marking them on a grid, and learners use it to view explanations and draw inferences about the true shape of the rectangle the teacher is trying to describe to them. |
| Explanation Form | Explanations were participant-generated in Experiment 1. In both experiments, explanations consist of marks on a grid. These marks could either be labelled as inside the rectangle ("positive"), outside the rectangle ("negative"), or (in Experiment 2) unspecified. |
| Study Design or Comparison Made | The procedure for Experiment 1 had two steps: The teaching step (in which participants trief to explain the shape of a rectangle to a learner by indicating only 1-3 points on the screen) and the learning step (in which participants attempted to draw the rectangle using teacher explanations). Experiment 2 also had two steps. The participants first viewed all possible rectangles, then a learning step in which they were shown a set of examples and were then to select a subset of the examples to be labelled as either inside or outside the rectangle. The participants would then draw the rectangle on the screen to indicate its predicted position. |
| Manipulation of Explanation Goodness | Explanation quality was observed but not evaluated. Teachers tended to try to label rectangle points in such a way that would maximize the amount of knowledge they could transfer to the learner about the nature of the rectangle. Learners behaved differently when they knew the information they received is pedagogically oriented. |
| Evaluation of Mental Models | Mental models were indirectly evaluated, in the sense that this is a pedagogical model which reveals insight into how people learn, teach, and reason. The rectangle game was used to compare participants' |
| | inferences to statistical, qualitative model predictions and demonstrate the important effects of pedagogical situations on learning and reasoning |
| Performance (Task or Evaluation Measure) | In Experiment 1, the work is trying to measure how the empirical data conforms to the pedagogical model it postulates. For teachers, the distribution of positive and negative labels were compared to model predictions using Pearson's r. Learners were evaluated by using chi-squared to show that inferred rectangles were not randomly positioned, and using maximum a posteriori model predictions to measure consistency with the pedagogical model. In Experiment 2, the study aims to ensure that learners are able to make stronger inferences when they know they are receiving pedagogical information; performance here is measured using chi-squared and Pearson's r to show that the learners in Experiment 2 chose examples at random and had little correlation with model predictions. |
| Trust/Reliance Measures | Not evaluated. |

| Authors | Shafto et al. 2014 |
|---|---|
| AI System | Expands on Shafto and Goodman (2008), with additional experiments. The system is a computational model of rational pedagogical reasoning. Perhaps the most important concept covered is the effects of pedagogical vs. non-pedagogical sampling of explanations. Each experiment involved a different pedagogical scenario. |
| Research Participants | Experiment 1 involved 73 undergraduates from the University of Louisville; Experiment 2 involved 28 undergraduates from UC Berkeley and 84 members of the community around UC Berkeley. Experiment 3 involved 86 undergraduates from the University of Louisville. |
| Explanation Interface | The interface for the rectangle game used in Experiment 1 was run using MATLAB. Teachers used the interface to provide explanations by marking them on a grid, and learners use it to view explanations and draw inferences about the true shape of the rectangle the teacher is trying to describe to them. Experiment 2 employed 27 physical cards to present geometrical stimuli, and the participants would assume a teaching role and produce explanations. therefore the explanations devised by participants in a teaching role. In Experiment 3, "teacher" participants produced explanations using a GUI that graphically illustrated causal relationships between (nonsensical) variables. Participants could interact with this system by turning variables on or off and observing the effects the manipulation might have on other variables. |
| Explanation Form | In Experiment 1 explanations were generated by teachers and consisted |

| | of marks on a grid. These marks could either be labelled as inside the rectangle ("positive"), outside the rectangle ("negative"), or (in Experiment 2) unspecified.

In Experiment 2 the explanations generated by the teachers would ideally convey the distribution of lines on the full set of 27 cards to a learner who could only see three cards.

In Experiment 3 explanations were generated by teachers. Teachers selected two variable configurations ("interventions") to teach learners who would only be able to see the causal relationships the interventions include. |
|---|---|
| Study Design or Comparison Made | Experiment 1 had participants play the rectangle game, in one of three conditions: 1) Teaching-Pedagogical Learning (teaching and learning from a teacher), Pedagogical Learning (observing model hypotheses and learning from a teacher), and Non-Pedagogical Learning (observing model hypotheses and learning without a teacher).

Experiment 2 divided participants into teachers and learners. It then assigned learners into one of four conditions: 1) explanations sampled randomly and learners told as such; 2) explanations sampled pedagogically and learners told as such; 3) explanations sampled randomly but learners told they were sampled pedagogically; and 4) explanations sampled pedagogically but learners told they were sampled randomly.

Experiment 3 had two stages (a teaching or explanation stage, followed by a learning stage), and divided participants into three conditions: 1) Teaching-Pedagogical Learning (select interventions for teaching; learn pedagogically); 2) Pedagogical Learning (select interventions to explore the system; learn pedagogically); and 3) Non-Pedagogical Learning (select interventions to explore the system; learn non-pedagogically). Of interest, Experiment 3 explored pedagogy with regard to causal reasoning and probabilistic events. |
| Manipulation of Explanation Goodness | Explanation quality was observed but not evaluated. |
| Evaluation of Mental Models | Mental models were indirectly evaluated in the sense that the experiment would reveal into how people learn, teach, and reason. |
| Performance (Task or Evaluation Measure) | Teaching and learning performance were measured based on the extent to which the human data correlated with model predictions.

Teaching performance was quantified by comparing how the human data fit to pedagogical and non-pedagogical models using a likelihood ratio test.

Teaching performance was also quantified by observing the frequencies |

| | of intervention pairs chosen by teachers, and the correlating them to the model predictions. Teachers were also evaluated for how their teaching experience affected their ability to learn. |
|---|---|
| Trust/Reliance Measures | Not evaluated. |

| Authors | Shinsel et al.,2011. See also Kulesza et al., 2011. |
|---|---|
| AI System | Intelligent assistant based on machine learning, for classifying messages from public newsgroup postings into one of four categories: computers, religion, cars, or motorcycles. User's classifications are used as training data for the ML system. For each message that is to be classified, the system infers the user's judgments on similar messages. |
| Research Participants | Forty-eight university students. |
| Explanation Interface | Display components showed the message to be classified, the text pieces within the message that supported the classification, the decision of the classifier, the classifier's confidence it its classification, the decisions already made by the mini-crowd, and the Ml systems inference concerning the user's classification for similar messages. |
| Explanation Form | The text pieces within the message that supported the classification, Graphical elements such as icons, colors, widgets, and charts to communicate explanations; e.g., pie charts to explain confidence and uniqueness of system predictions. |
| Study Design or Comparison Made | Sizes of the "mini-crowds" that evaluated the systems: 1, 2, 6, or 11 people.  This was a within variable. |
| Manipulation of Explanation Goodness | Not evaluated. |
| Evaluation of Mental Models | NASA-TLX was applied to measure mental workload. Working in a larger crowd resulted in lowered frustration ratings. |
| Performance (Task or Evaluation Measure) | Participant judgments of whether the classifier's classification was "right, maybe right, maybe wrong or wrong." Ability of users to identify system errors (misclassifications) (The system was calibrated to 85%-88% accuracy.) Cost/benefit of smaller versus larger mini-crowds. |
| Trust/Reliance Measures | The development of trust is a primary focus of this study. There was a cost-benefit trade-off in that larger crowds found more errors, but more of the same errors. (Larger crowds were able to cover more messages in the time-limited task.)  Participants paid attention to the crowd's error-finding successes and its failures. |

| Authors | Swinney, 1995 |
|---|---|
| AI System | The AI system is an externally developed expert system that is used in the auditing industry to evaluate the amount of reserve required for loans and to explain the reasoning behind its evaluation. |
| Research Participants | 41 professional auditors having various levels of experience. |
| Explanation Interface | The interface of the expert system was not described, as participants did not interact with it. Participants were given information about a hypothetical audit, the expert system's conclusions and explanations. |
| Explanation Form | Participants received printouts expressing the conclusions and explanations of the expert system. |
| Study Design or Comparison Made | It should be noted that it was the experimenters—not the participants—who were the ones who actually used the expert system. The purpose was to study the effect of machine-generated versus participant-generated explanations of the expert system's output (positive or negative with regard to adequacy of the reserve amount of a loan) on the user's agreement (or disagreement) with the expert system's outputs. |
| Manipulation of Explanation Goodness | Explanation goodness was not evaluated. Focus was on the notion that a positive evaluation would limit the effectiveness of an audit by reducing the auditor's skepticism. |
| Evaluation of Mental Models | Participant mental models were not evaluated. |
| Performance (Task or Evaluation Measure) | Comparison of the expert system's explanations and the participant-written explanations in a hypothetical audit case; to what degree were participants' judgments swayed by the inclusion of an explanatory element? |
| Trust/Reliance Measures | Trust and reliance were not evaluated. |

| Authors | Teach and Shortliffe, 1981 |
|---|---|
| AI System | Clinical consultation systems |
| Research Participants | Eighty-five physicians who were attending a course on medical information systems. Sixty-one medical school faculty members, who were not taking the course on medical information systems. |
| Explanation Interface | Indicates a desire for non-dogmatic explanations which function similarly to a human consultant. |
| Explanation Form | Questionnaire on attitudes regarding clinical consultation systems, especially acceptability, expectations on how the systems were likely to affect medical practice, and physician's desires regarding the performance capabilities of the systems. |
| Study Design or Comparison Made | The questionnaire was administered prior to the course. Primary comparison was of the tutorial and non-tutorial groups of |
| | participants. Comparison was also made across the participant's specializations (e.g., surgeons, researchers, etc.) and across those with and those without computing experience. |
| Manipulation of Explanation Goodness | Not applicable. |
| Evaluation of Mental Models | Indirectly reflected in the ratings of acceptability across applications. Systems for medical records, hospital information systems, patient monitoring, and consultation ere rated as acceptable (97% - 81%) whereas systems for physical and licensing were rated as far less acceptable (52%-36%.  Systems intended as physician substituted were rated the least acceptable (32%). |
| | Indirectly reflected in the ratings on expectations of how the systems would affect medical practice.  Physicians expressed concern that the systems would lead to increased government control, shift the blame for poor decisions.  On the other hand, physicians did not think that the systems would lead to decreased efficiency, force the physicians to adapt their reasoning ("think like a computer"), or lead to job insecurity or loss of prestige. |
| Performance (Task or Evaluation Measure) | Physicians expressed a strong desire for systems to explain their advice, and also tailor their explanations to the needs and characteristics of the physician-user (e.g., level of background knowledge). |
| | It was preferred for systems to engage in "reasonable problem solving" and parallel the physician's reasoning as much as possible. |
| | It was preferred for systems to be easily learned. |
| | It was preferred that the interface be based on natural language, in a dialog form. |
| Trust/Reliance Measures | It was not seen necessary for the systems to be perfectly accurate or lead to perfect treatment plans.  Physicians would not accept a consultation system as a standard for medical practice. Nor would they recommend reducing the knowledge requirements of physicians because a medical information system (knowledge base) was available. |

| Authors | Tullio, et al. 2007 |
|---|---|
| AI System | Intelligent system to estimate the interruptibility of managers in an office setting. Based on a statistical model using data on manager hourly self-reports of interruptibility over a three month period. |
| Research Participants | Four managers and eight non-technical workers in a human resources department, having no previous knowledge of programming, or machine learning. |
| Explanation Interface | Color gradient display outside managers' offices, showing degrees of |

| | interruptibility based on the statistical model. |
|---|---|
| Explanation Form | Explanations designed to shift a user's mental model to be in line with how the AI system actually works, so as to increase user trust and efficacy. The AI system was itself not actually explained. Rather, participants had to try and figure that out. |
| Study Design or Comparison Made | Six-week field deployment. Two worker groups shown different gradient displays. One was the gradient display alone, the other showed data from "sensors," such as whether the manager's keyboard was in use, whether there was audio activity, etc. |
| Manipulation of Explanation Goodness | Not specifically evaluated, though explanation complexity was (see above). |
| Evaluation of Mental Models | The primary focus of the study was the valuation of mental models of an intelligent system and their development. Of particular concern was whether user's misconceptions of how the AI system works were corrected across experience, as the flaws or contradictory evidence were pointed out. <br><br> Primary data were from a series of structured interviews and a series of surveys consisting of a few questions that workers answered at the managers' doors. Workers reported on which sensors (data types) they used and which they did not. <br><br> Workers were asked to describe the reasoning they used in deciding whether a manager was interruptible (open door, sound of talking, door tags signaling availability, etc.). Workers were asked to describe the intelligent constructs that they ascribed to the AI. <br><br> Simple feedback was not sufficient to correct user's mental models. Misconceptions persisted (e.g., that the manager actually controlled the display, that the system relined on simple rules, something like a decision tree, etc.). Only three participants came to believe that the AI system was based on a statistical model. |
| Performance (Task or Evaluation Measure) | Workers have to estimate their own interruptibility and that of their supervisors. |
| Trust/Reliance Measures | Trust was also a focus of the study, based on the premise that recommender systems are trusted more if their recommendations are explained. Results showed that users relied more on the system if they trusted it, but nonetheless relied more on their own estimates. |

| Authors | van Lent, Fisher & Mancuso, 2004; see also Core, et al. 2006 |
|---|---|
| AI System | US Army system for command training. Considered a form of serious game. Includes software agents that control the actions of non-player characters, including simulated subordinate units (platoons) and the opposing forces.<br>Coined the term "Explainable AI" (XAI) |
| Research Participants | Soldiers in training at the Infantry Captain's Career Course. |
| Explanation Interface | One display presents an aerial perspective on theater of operations with side menus populated with tasks for each of the platoons. Another display shows a terrestrial perspective, for event playback and after-action review. The player can move through the landscape as the event is replayed, can pause, and can fast-forward. |
| Explanation Form | Key events, based on the mission plan, are depicted as flags on a mission timeline (presented at the bottom of the event playback window). Player can scroll the timeline and jump to the time of subevents or command decisions. Explanations are based upon the task decomposition, and for each subtask the statistics on weapon used, unit involved, task status, etc. The explanations are essentially a record of the log files, presented as a explanation of the platoon behavior. The XAI would provide records in response to these questions: What is the platoon's mission? What is the mission status? How is the task organized, and What is the platoon's ammunition status. Thus, the explanations are about the task, not about how the AI works. |
| Study Design or Comparison Made | No manipulated Independent variables. |
| Manipulation of Explanation Goodness | Explanation goodness was evaluated indirectly by a review of post-course comments and feedback. Sixty percent of trainees rated the training system as having high fidelity and training value. |
| Evaluation of Mental Models | Mental models were not evaluated. |
| Performance (Task or Evaluation Measure) | Not applicable. The authors note that "The fundamental limitation of the XAI is the depth of the explanations it can provide. The current XAI system is limited to explaining how the pre-existing task knowledge was applied... it doesn't include any deeper knowledge about why each task should be approached" (p. 907). |
| Trust/Reliance Measures | Trust/reliance were not specifically evaluated. |

| Authors | Williams, et al. 2016 |
|---|---|
| AI System | The AI (Adaptive eXplanation Improvement System) was designed to crowdsource explanations and ratings of explanations (of math problems) from learners (users) and to use machine learning to sift through the crowdsourced information to identify which explanations are most useful to learners such that they may be integrated into the system's explanation pool and presented to future users. |
| Research Participants | The case study, participants were 150 US residents sourced from Mechanical Turk. For the randomized experiment, the participants were 524 people who were also recruited with Mechanical Turk. |
| Explanation Interface | The interface was a GUI that integrates modules for question-answering, explanation-reading, explanation-rating, and explanation-writing user tasks. Through the GUI, the AI interacts with a community of learners to provide problem explanations which are effective to other learners, even in the complete absence of instructor resources. |
| Explanation Form | Textual explanations given by previous learners who used the system. |
| Study Design or Comparison Made | The purpose of the case study was mainly to seed the AI with user-generated explanations. The evaluation occurred in the randomized experiment, which distinguished four conditions (explanations from the pool after 75 learners, explanations from the pool after 150 learners, explanations rejected by the AI, and explanations written by instructors) and a control (no explanation provided) with respect to the nature of the explanation received by users for each question. |
| Manipulation of Explanation Goodness | Goodness was not evaluated per se. |
| Evaluation of Mental Models | Mental models were not evaluated per se. Though not argued explicitly, it is possible that measurement of learning, is a surrogate, since it takes into account a participant's ability to generalize what they have learned from explanations to similar—but novel—problems. |
| Performance (Task or Evaluation Measure) | Human performance was measured by self-reported ratings of perceived skill and by quantifying the degree by which participants' accuracy improved over the course of the experiment (i.e., a measure of learning), comparing the AI-generated explanations to the default practice without explanations. Results showed that the AI system elicits and identifies explanations that are helpful to users. |
| Trust/Reliance Measures | Trust or reliance were not evaluated. |

| Authors | Ye and Johnson, 1995 |
|---|---|
| AI System | Simulation of an an expert system's auditing diagnosis of a hypothetical medical products company. |
| Research Participants | Professional auditors and accountants, all having more than six years of experience. |
| Explanation Interface | Text output of the expert's system's determination (written by the experimenters), accompanied by text versions of the three types of explanation. |
| Explanation Form | Based on the early research on expert systems with explanation capabilities (i.e., Clancey, 1983; Swartout, 1983), the researchers used three types of explanations: Trace (the inferential steps taken by the expert system, Justification (the causal argument or rationale for each of the inferential steps), and Strategy (high level goal structure that determined how the expert system used its knowledge base). |
| Study Design or Comparison Made | Fully within, single group, repeat measure design. For each of a series of auditing decisions made by the expert system, participants could opt to see one of the three types of explanations. (The initial instructions included a practice case, for which the participants could see all three of the types of explanations.) |
| Manipulation of Explanation Goodness | Not explicitly evaluated, except for the manipulation of Type of Explanation. |
| Evaluation of Mental Models | Attempted to show whether explanations influence user confidence in an expert system (a medical diagnosis recommender system). Previous research had focused primarily on the problem of generating explanations. |
| Performance (Task or Evaluation Measure) | Participants rated their belief in the six conclusions of the expert system (Likert scale), both before and after seeing one of the types of explanation. |
| Trust/Reliance Measures | By implication, explanations have a positive impact on on user acceptance of an expert system; explanations increased participant's belief in the system recommendations. Participants preferred, and spent more time reviewing, the Trace-type and Justification-type explanations. |

| Authors | Zhang et al. 2016 |
|---|---|
| AI System | Robotic agent for robot task planning, using a "blocks world" type domain. |
| Research Participants | Thirteen university students evaluated 23 tower-building plans generated by the robot. |
| Explanation Interface | Robot plans are portrayed in photographs showing the series of robot actions. |
| Explanation Form | Explanations were manifest implicitly in the actions of robotic |

| | machines, as they employed more human-like methods to achieving goals (non-linguistic, non-graphical). |
|---|---|
| Study Design or Comparison Made | Participants evaluated 23 tower-building plans generated by the robot. |
| Manipulation of Explanation Goodness | Not evaluated. |
| Evaluation of Mental Models | Not evaluated. |
| Performance (Task or Evaluation Measure) | Explicability (Ability to associate labels for tasks in a plan). Predictability (predict next action) After labeling each plan, participants rated the plan comprehensibility |
| Trust/Reliance Measures | Indirectly evaluated: The researchers conclude that their measure of "explicability" captures human human interpretation of plans. |